\documentclass{article}
\usepackage{iclr2027_conference,times}

\usepackage{amsmath,amsfonts,bm}

\def\eqref#1{equation~\ref{#1}}

\def\1{\bm{1}}

\DeclareMathAlphabet{\mathsfit}{\encodingdefault}{\sfdefault}{m}{sl}
\SetMathAlphabet{\mathsfit}{bold}{\encodingdefault}{\sfdefault}{bx}{n}

\usepackage{hyperref}
\usepackage{url}
\usepackage{enumitem}
\usepackage{xcolor}

\newcommand{\parabf}[1]{\noindent\textbf{#1}\xspace}

\usepackage{graphicx}
\usepackage{subcaption}
\usepackage{wrapfig}
\usepackage{needspace}
\usepackage{listings}
\usepackage{booktabs}
\usepackage{array}
\usepackage{multirow}
\usepackage{makecell}
\newcolumntype{C}[1]{>{\centering\arraybackslash}p{#1}}

\usepackage{amssymb}
\usepackage{pifont}

\definecolor{lightcyan}{RGB}{10,110,150}
\hypersetup{
  colorlinks=true,
  linkcolor=lightcyan,
  citecolor=lightcyan,
  filecolor=lightcyan,
  urlcolor=lightcyan
}

\usepackage{xspace}
\newcommand{\harness}{{OpenRUA}\xspace}

\usepackage[capitalize,noabbrev,nameinlink]{cleveref}
\crefname{appendix}{Appendix}{Appendices}

\usepackage{tikz}

\newcommand{\graycircle}[1]{%
  \tikz[baseline=(char.base)]{
    \node[
      shape=circle,
      fill=gray!30,
      text=black,
      inner sep=0.2pt,
      minimum size=1em,
      font=\sffamily\small
    ] (char) {#1};}}

\title{\harness: Robot-Use Agents Are Zero-Shot Visuomotor Policies}

\author{\textbf{Zhaoyang Chu$^{1}$ \quad Earl T. Barr$^{1}$ \quad Claire Le Goues$^{2}$ \quad Peter O'Hearn$^{1}$} \\
\textbf{Mark Harman$^{1}$ \quad Federica Sarro$^{1}$ \quad He Ye$^{1}$\thanks{Corresponding author: \texttt{he.ye@ucl.ac.uk.}}} \\
$^{1}$University College London \quad
$^{2}$Carnegie Mellon University
}

\iclrfinalcopy

\begin{document}

\maketitle
\lhead{Preprint}

\begin{abstract}
Coding agents are extending their reach into the physical world by writing and executing robot control programs.
One might expect the agents to use the existing mature software stack that engineers have developed over decades to access sensors and control motion.
Yet prior work primarily engineers complex custom harnesses to orchestrate agents for robot use, particularly by prescribing specialized workflows and providing bespoke interfaces.
This raises the question: \textit{``Is such additional harness engineering necessary?''}
We introduce \harness, a zero-abstraction harness that bypasses bespoke abstraction layers by providing off-the-shelf coding agents with only terminal access to the robot's native software interface ROS~2.
\harness employs a minimalist \textit{workspace-as-harness} design, only offering ROS~2 documentation and basic tools while leaving the coding agent to organize its own work without orchestrating any agentic workflow.
Within this workspace, \harness recasts perception as file I/O and manipulation as coding.
With Claude Code powered by Claude Opus 5, \harness achieves success rates of 99.0\% on CaP-Bench and 87.0\% on LIBERO-PRO, demonstrating that an off-the-shelf coding agent can serve as a zero-shot visuomotor policy through the robot's native interface, without bespoke primitives or task-specific training.
Under this minimalist design, further analysis reveals striking emergent behaviors of coding agents:
\graycircle{1}~For perception, the agent spontaneously writes programs that process raw sensory inputs (\textit{e.g.}, reconstructing point clouds from depth) and derive metric measurements (\textit{e.g.}, computing coordinates for Cartesian motion targets) in 96.80\% of episodes.
\graycircle{2}~For manipulation, the agent spontaneously builds motion-control clients (\textit{e.g.}, gripper control) in 95.87\% of episodes and closed-loop control programs (\textit{e.g.}, adjusting motion based on sensor feedback) in 50.13\% of episodes.
Our code is available at \url{https://github.com/terminalworld/OpenRUA}.

\end{abstract}
\section{Introduction}
\label{sec:intro}

Robot control has long been engineered by writing explicit programs that translate task specifications into physical motion~\citep{fikes1971strips, li1994mathematical, siciliano2009robotics}.
These programs execute on a multi-layered software stack, through which human engineers coordinate drivers, messaging, planning, and actuation to implement the required behavior~\citep{macenski2022ros2, garcia2020rse}.
However, this process requires substantial effort and human expertise, with further engineering often needed for new tasks or environments.

One path is to learn end-to-end \textit{visuomotor policies}, such as \textit{Vision-Language-Action} (VLA) models~\citep{google2025gemini_robotics, kim2024openvla, intelligence2026pi07, bjorck2025gr00t} and \textit{World Action Models} (WAMs)~\citep{agarwal2026cosmos3, ye2026world, li2026causal}, that map visual observations and language instructions directly to low-level actions.
However, they require large-scale demonstration data, and their generalization remains limited beyond the training distribution~\citep{zhou2026libero_pro, fei2025libero_plus, kim2026libero_para}. 
For example, LIBERO-PRO~\citep{zhou2026libero_pro} reports that OpenVLA~\citep{kim2024openvla} and $\pi_0$~\citep{black2024pi_0}, despite average success rates above 90\% on standard LIBERO, achieve 0.0\% success under position and task perturbations.

Another path is to \textit{revisit} explicit programming for robot control~\citep{fu2026capx}, while replacing the human engineer with modern coding agents such as Claude Code~\citep{anthropic2025claudecode} or Codex~\citep{openai2025codexcli}.
Such an approach is known as \textit{Code as Policies} (CaP)~\citep{liang2023cap}, where the policy itself can be inspected, edited, and debugged by the coding agent without retraining~\citep{elmaaroufi2026rho, lu2026aspire}.
As a result, the coding agent acts as a \textit{Robot-Use Agent} (RUA)~\citep{isola2026rua}, operating the robot the way it operates a computer~\citep{google2025gemini_robotics, fu2026capx, robocurve2026astra, berman2026claudeplaysrobotics}.

\begin{figure}[t]
\centering
\includegraphics[width=0.96\linewidth]{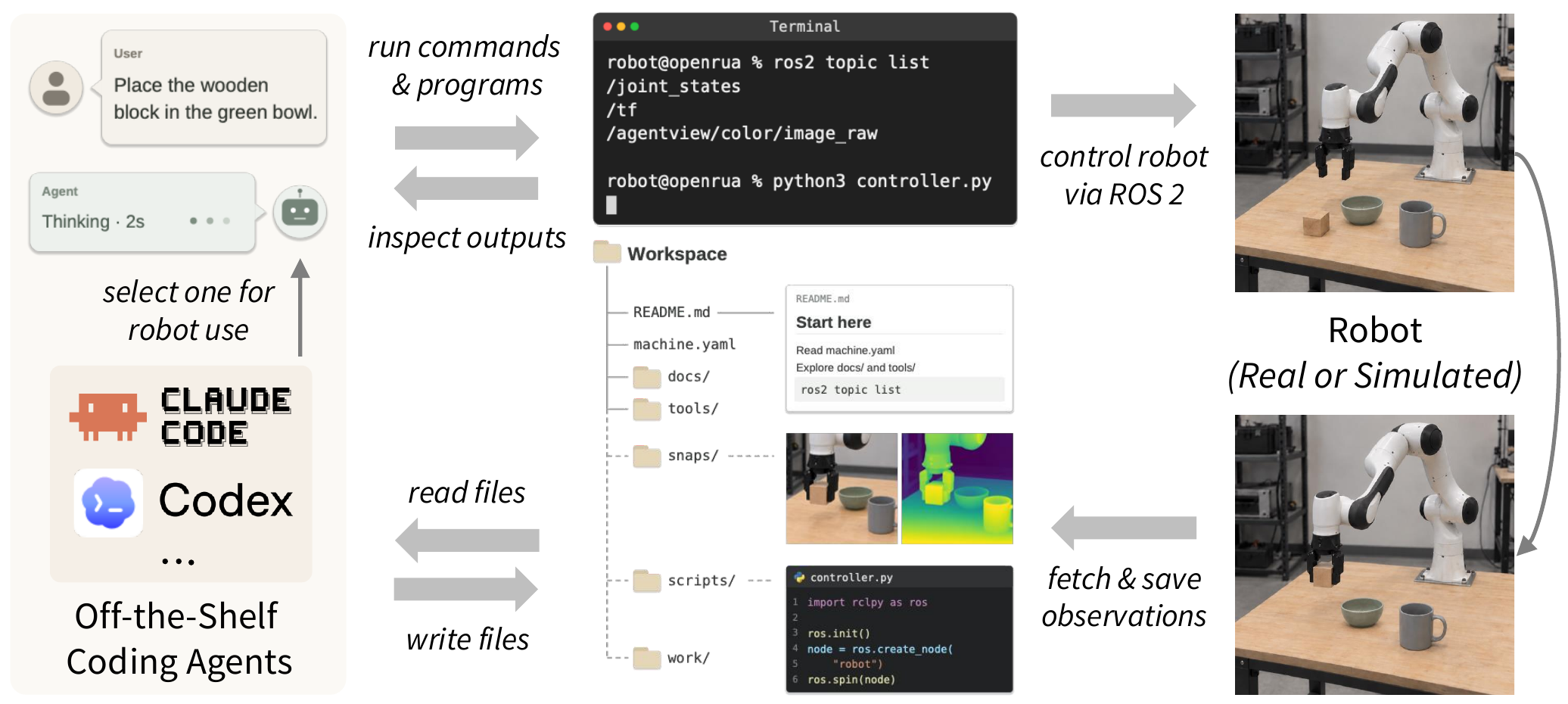}
\caption{\textbf{\harness: a minimalist workspace-as-harness for robot-use agents.}
Given a task such as placing the block in the bowl, \harness enables any off-the-shelf coding agent to operate a real or simulated robot through its native ROS~2 interface, bypassing bespoke abstraction layers.
The workspace serves as the harness by only providing ROS~2 documentation and basic tools, making robot perception accessible through file I/O and manipulation through coding, without orchestrating any agentic workflow.
Within this workspace, the agent exhibits emergent perception and control capabilities as it works to complete the task.
It writes and executes \texttt{ros2cli} commands or \texttt{rclpy} programs that send control requests to the robot's controllers, here grasping and lifting the wooden block.
The agent saves new RGB-D observations as workspace files and uses images, computed measurements, and terminal feedback to autonomously verify progress and refine its actions until the block is placed in the bowl.
}
\label{fig_overview}
\vspace{-1em}
\end{figure}

Despite this progress, existing RUA systems devote significant effort to engineering custom harnesses for robot use, especially designing the tool interfaces and primitive libraries through which agents compose policies to operate robots in prescribed workflows.
These systems either wrap learned visuomotor policies as callable tools~\citep{zhang2026harnessvla, galanti2026pigey, ding2026zetta, wang2026shaper} or hand-craft custom libraries of perception and control primitives based on human expertise~\citep{fu2026capx, lu2026aspire, elmaaroufi2026rho, zhang2026rats}.
However, it remains unclear whether such additional harness engineering is necessary for coding agents to operate robots effectively, given that robots have a mature ecosystem of software interfaces developed over decades for human engineers to access sensors and control motion~\citep{macenski2022ros2, garcia2020rse}.
This raises the question: \textit{``Can an off-the-shelf coding agent serve as a zero-shot visuomotor policy through the robot's native software interface, without relying on bespoke primitives or learned policies?''}

In this work, we introduce \harness, a zero-abstraction harness for robot-use agents that builds on the robot's native interface, \textit{Robot Operating System 2} (ROS~2)~\citep{macenski2022ros2}, the de facto standard of the robotics community.
\harness bypasses bespoke abstraction layers by offering off-the-shelf coding agents only terminal access to ROS~2, allowing them to operate the robot as a human engineer would.

As illustrated in~\cref{fig_overview}, \harness frames robot use as a live, interactive software project and employs a minimalist \textit{workspace-as-harness} design.
This design rests on the observation that frontier coding agents already operate autonomously on unfamiliar repositories to complete tasks~\citep{openai2024swebenchverified, merrill2026terminalbench, chu2026terminalworld}.
We hypothesize that the orchestration with bespoke interfaces in existing RUA systems compensates for limited agent capability rather than reflecting a requirement of robot control, and that a capable coding agent needs only a workspace directory.
Specifically, the workspace serves as the harness by only offering ROS~2 documentation and basic tools, leaving the coding agent to organize its own work without a prescribed agentic workflow.
Within this workspace, \harness recasts perception as file I/O and manipulation as coding: task-specific perception and control strategies emerge as the agent acquires observations, interprets images, computes metric measurements, and writes control programs.
As a result, the coding agent itself serves as a zero-shot visuomotor policy: it goes from raw observations to motor commands, reasoning explicitly about geometry and iteratively refining its actions from execution feedback.

We conduct comprehensive experiments to evaluate \harness on three widely used benchmarks, \textit{i.e.}, CaP-Bench~\citep{fu2026capx}, LIBERO-PRO~\citep{zhou2026libero_pro}, and RoboCasa365~\citep{nasiriany2026robocasa365}, in a zero-shot setting without access to privileged simulator state and the benchmark's success predicates.
With Claude Code powered by Claude Opus 5, \harness achieves success rates of 99.0\% on CaP-Bench and 87.0\% on LIBERO-PRO, exceeding leading systems such as ASPIRE~\citep{lu2026aspire} and Harness VLA~\citep{zhang2026harnessvla} by 18.0 and 4.6 percentage points, respectively.
On unseen composite tasks in RoboCasa365, \harness achieves 28.1\% success, exceeding Harness VLA by 13.1 percentage points.
The results show that an off-the-shelf coding agent can serve as a zero-shot visuomotor policy through the robot's native software interface, without bespoke primitives or learned policies.

Further analysis reveals several findings about emergent agent capabilities:
\graycircle{1}~For perception, the agent spontaneously writes programs that process raw sensory inputs (\textit{e.g.}, cropping images, adding grids, and reconstructing point clouds from depth) and derive metric measurements (\textit{e.g.}, computing coordinates for Cartesian motion targets) in 96.80\% of episodes.
\graycircle{2}~For manipulation, the agent spontaneously builds its own motion-control clients (\textit{e.g.}, trajectory execution and gripper control) in 95.87\% of episodes and closed-loop control programs (\textit{e.g.}, adjusting motion or stopping based on sensor feedback) in 50.13\% of episodes.
\graycircle{3}~On RoboCasa365, \harness maintains comparable success rates on seen and unseen composite tasks (26.3\% and 28.1\%, respectively), whereas RLDX-1 drops to 5.0\% and Harness VLA, which orchestrates RLDX-1, drops to 13.8\% on unseen tasks.
This contrast highlights the generalization potential of \harness without reliance on a learned robot policy.
\graycircle{4}~Increasing test-time computation through more interaction or greater reasoning effort improves task success without additional robot training or bespoke primitives, while performance eventually saturates at levels that vary across models.

\section{Background and Related Work}
\label{sec:background}

We review visuomotor policies, robot-use agents, and CLI coding agents, and introduce the Robot Operating System; an extended version is provided in~\cref{app_extended_related_work}.

\parabf{Visuomotor Policies.}
A visuomotor policy maps raw sensory observations directly to low-level robot actions~\citep{levine2016visuomotor}.
Such policies are predominantly learned end to end from robot demonstrations, as in VLA models~\citep{brohan2023rt2, kim2024openvla, google2025gemini_robotics, bjorck2025gr00t, intelligence2026pi07} and WAMs~\citep{agarwal2026cosmos3, ye2026world, li2026causal}.
However, their generalization remains bounded by the training distribution~\citep{zhou2026libero_pro, fei2025libero_plus, kim2026libero_para}.
\harness instead casts an off-the-shelf coding agent as a zero-shot visuomotor policy, requiring no robot demonstration data or fine-tuning.

\parabf{Robot-Use Agents.}
RUA~\citep{isola2026rua} refers to LLM agents that operate a robot the way they operate a computer, through tool calls or executable programs, a paradigm that traces back to CaP~\citep{liang2023cap, singh2022progprompt}.
One line of work wraps learned visuomotor policies as callable tools, with the agent decomposing tasks and selecting tools while the underlying policies handle physical execution~\citep{ichter2023saycan, zhang2026harnessvla, galanti2026pigey}.
Another exposes low-level actions directly to the agent, which specifies end-effector poses step by step or writes control programs using predefined, hand-crafted primitives~\citep{hu2026via, zhou2026minimal_interface, fu2026capx, elmaaroufi2026rho}.
\harness instead provides the coding agent with only terminal access to the robot's native ROS~2 interfaces and a workspace, allowing it to build task-specific perception and control tools during execution rather than requiring them to be engineered in advance.

\parabf{CLI Coding Agents.}
CLI coding agents such as Claude Code~\citep{anthropic2025claudecode} and Codex~\citep{openai2025codexcli} resolve real-world software engineering tasks through autonomous iteration~\citep{openai2024swebenchverified, merrill2026terminalbench}.
These agents exhibit closed-loop agency over long horizons: they navigate codebases, edit and run code, issue CLI commands, interpret terminal output, and recover from errors.
\harness turns this competence toward robot control by treating the robot as a live software project, allowing the agent to apply these capabilities to physical tasks without adaptation.

\parabf{Robot Operating System.}
ROS~2 connects robot components and exposes their sensing, state, and control capabilities via topics, services, and actions~\citep{macenski2022ros2}.
The native \texttt{ros2cli} command-line tools and \texttt{rclpy} Python client library allow developers to inspect these interfaces, acquire data, and send control requests.
Prior work connects LLM agents to ROS through an additional abstraction layer of custom tools or predefined primitives, with evaluations limited to in-house tasks rather than standard robot control benchmarks~\citep{royce2026rosa, mower2024ros_llm}.

\section{\harness: Workspace as Harness for Robot-Use Agents}
\label{sec:method}

Given a natural-language task, \harness provides an off-the-shelf coding agent with only terminal access to the robot's native ROS~2 interface and a workspace on the robot's computer (\cref{fig_overview}).
The zero-abstraction design exposes the robot's existing ROS~2 interfaces without introducing an additional agent-specific interface layer.
The workspace provides ROS~2 documentation and 
\begin{wrapfigure}{r}{0.5\linewidth}
\centering
\begin{subfigure}[t]{0.96\linewidth}
\setlength{\fboxsep}{5pt}\fbox{\begin{minipage}{\dimexpr\linewidth-2\fboxsep-2\fboxrule}
\ttfamily\scriptsize\setlength{\parskip}{2pt}\setlength{\parindent}{0pt}\raggedright
You are working on a robot's onboard computer.

Your task: \{task\}

Survey the machine yourself to find out what robot this is and what it can do; work until the task is physically done, verify it your own way, then finish.

The workspace contains starter docs and tools you can use.
\end{minipage}}
\caption{The opening prompt.}
\label{fig_workspace_prompt}
\end{subfigure}

\begin{subfigure}[t]{\linewidth}
\includegraphics[width=\linewidth]{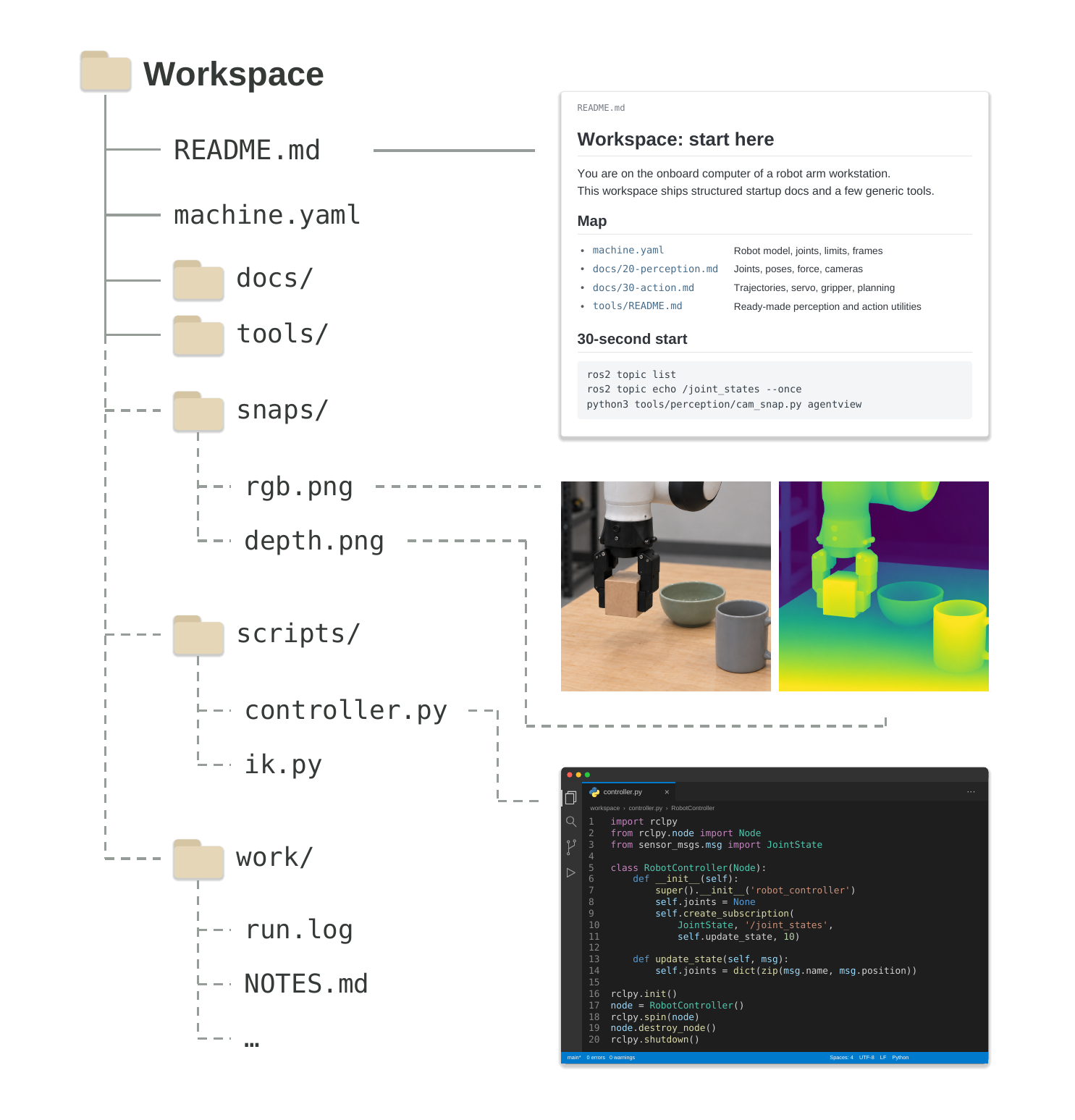}
\caption{The workspace tree, with solid branches indicating starter documentation and tools and dashed branches indicating artifacts created by the agent during execution.}
\label{fig_workspace_tree}
\end{subfigure}
\caption{\textbf{A minimalist workspace-as-harness for robot use.}
(a) The opening prompt specifies the task and asks the agent to explore the machine, complete the task, and verify the result.
(b) The workspace offers documentation and tools for getting started and stores the observations, programs, and records the agent produces during execution.}
\label{fig_workspace}
\vspace{-25pt}
\end{wrapfigure}
basic starter tools for the agent to explore the robot, and stores the observations, programs, and notes the agent produces during execution.
Within this workspace, \harness recasts perception as file I/O and manipulation as coding.
We expect task-specific perception and control strategies to emerge as the agent explores this workspace, writes and tests programs, and uses execution feedback to refine its solutions, much as it would when working on an unfamiliar software project.
Without orchestrating any agentic workflow, \harness allows 
the agent itself to decide when to observe, which commands or programs to write and execute, and how to verify task completion.

\parabf{Workspace as Harness.}
\harness only provides guidance for robot use through a simple workspace containing ROS~2 documentation and basic tools, in which the coding agent can approach the robot as an ordinary software project.
As illustrated in~\cref{fig_workspace}, the opening prompt specifies the task and
asks the agent to survey the machine, complete the task, and verify the result.
The documentation describes the robot's capabilities and how to access them, while the starter tools provide executable examples that the agent can inspect, use, or adapt.
These materials provide a starting point rather than a prescribed workflow: the agent itself decides which resources to consult and how to organize its work using its built-in file and terminal capabilities (see~\cref{app_workspace} for details).
In this workspace, the agent itself can decide which resources to consult and how to organize its work (see~\cref{app_workspace} for details).

The initial workspace provides \texttt{README.md}, which is the entry point, directing the agent to \texttt{machine.yaml} for the robot's configuration, including sensors, control interfaces, and joint limits, and to the documentation for instructions on using these functionalities.
A small set of starter tools offers ready-to-run implementations of basic operations (\textit{e.g.}, saving camera images and sending joint trajectories).
The tools are provided as source files; the agent can inspect and adapt them, use them as references, or write its own programs using the native ROS~2 interfaces.

During task execution, the workspace itself also serves as an external, persistent memory, preserving observations and intermediate results beyond the agent's context window.
We expect the agent to autonomously organize images, programs, notes, and logs into folders such as \texttt{snaps/}, \texttt{scripts/}, and \texttt{work/}.
By revisiting them, the agent can recover prior findings, reuse or revise earlier programs, and build on prior work as the task progresses.

\parabf{Perception as File I/O.}
\harness reformulates robot perception as a standard file I/O task, allowing the agent to acquire data from the robot's native ROS~2 sensor topics and save it as workspace files.
Depending on the robot's available sensors and interfaces, these files can contain \graycircle{1}~visual data (\textit{e.g.}, RGB and depth images); \graycircle{2}~proprioceptive and kinematic states (\textit{e.g.}, joint states and frame transforms); and \graycircle{3}~force and effort feedback (\textit{e.g.}, force/torque readings and joint efforts); see~\cref{app_observation_acquisition} for details.
The agent can run the supplied starter tools, adapt their source code, or write its own \texttt{rclpy} programs to subscribe to the relevant topics and save the received data.
The saved files are accessible through the agent's built-in file tools: images can be opened for visual inspection, while numerical arrays and structured text can be read by programs for computation.
The agent is left to decide when to acquire new observations, which files to inspect, and what processing code to write based on the task and execution feedback.
We examine the agent's perception strategies that emerge under this design in~\cref{sec_perception}.

\parabf{Manipulation as CLI Coding.}
\harness reformulates robot manipulation as a CLI coding task, allowing the agent to write and execute \texttt{ros2cli} commands or \texttt{rclpy} programs in the terminal.
These commands and programs send trajectory, velocity, or gripper control requests through native ROS~2 interfaces to the robot's controllers, which execute the requests to produce physical motion.
The agent is free to issue commands step by step and inspect feedback between actions, or write programs that run closed-loop control using sensor feedback.
The agent can inspect terminal output and acquire new observations to assess execution outcomes and revise its commands or programs.
We examine the agent's control strategies that emerge under this design in~\cref{sec_behavior}.

\section{Experiments}

\subsection{Experimental Setup}
\label{sec_setup}
\parabf{Benchmarks.}
We evaluate \harness on three simulation benchmarks:
\graycircle{1}~CaP-Bench~\citep{fu2026capx} covers seven single-arm and bimanual manipulation tasks; we follow its protocol of 100 trials per task.
\graycircle{2}~LIBERO-PRO~\citep{zhou2026libero_pro} evaluates robustness to position and task perturbations across four LIBERO suites; we run ten trials per task, following prior work~\citep{zhang2026harnessvla, zhang2026rats, galanti2026pigey}.
\graycircle{3}~RoboCasa365~\citep{nasiriany2026robocasa365} evaluates mobile manipulation on atomic and composite kitchen tasks; we use its 50 target tasks with ten trials per task, following prior work~\citep{zhang2026harnessvla}.
We report success rates under each benchmark's official success predicate; task coverage, sampling, and scoring details are provided in~\cref{app_protocol}.

\begin{table}[t]
\centering
\footnotesize
\setlength{\tabcolsep}{2pt}
\renewcommand{\arraystretch}{1.15}
\caption{
\textbf{\harness achieves 99.0\% success on CaP-Bench and 87.0\% on LIBERO-PRO in a zero-shot setting.}
\textit{Train.}, \textit{Exp.}, and \textit{Oracle} indicate training on robot demonstrations, experience accumulated before evaluation, and access to the benchmark's success predicate during execution, respectively.
Reasoning effort is shown in parentheses; CC denotes Claude Code.
Baseline sources and evaluation protocols are detailed in~\cref{app_compared,app_protocol}.
}
\vspace{-0.5em}
\label{tab_main}

\begin{subtable}{\textwidth}
\centering
\caption{CaP-Bench}
\label{tab_main_capbench}
\resizebox{\textwidth}{!}{%
\begin{tabular}{llllccccccccccc}
\toprule
\multicolumn{2}{c}{\multirow{2}{*}{Method}} & \multicolumn{5}{c}{Setting} & \multicolumn{8}{c}{Success Rate (\%)} \\
\cmidrule(lr){3-7} \cmidrule(lr){8-15}
& & Agent & Model (Effort) & Train. & Exp. & Oracle & Lift & Stack & Wipe & Nut & Restack & 2A-Lift & Handover & Avg. \\
\midrule
\multicolumn{2}{l}{Human Expert} & -- & -- & -- & -- & -- & 93.0 & 73.0 & 100.0 & 80.0 & 47.0 & 94.0 & 91.0 & 82.6 \\
\midrule
\multirow{3}{*}{Agent} & CaP-X & \makecell[l]{CaP-\\Agent0} & \makecell[l]{GPT-5.2 (none)\\Opus 4.5 (off)\\Gemini 3 Pro (high)} & \ding{55} & \ding{51} & \ding{55} & 97.0 & 98.0 & 100.0 & 0.0 & 89.0 & 74.0 & 20.0 & 68.3 \\
 & RHO & Codex & GPT-5.5 (xhigh) & \ding{55} & \ding{51} & \ding{55} & 98.0 & 98.0 & 100.0 & 1.0 & 54.0 & 53.0 & 86.0 & 70.0 \\
 & ASPIRE & CC & Opus 4.6 (high) & \ding{55} & \ding{51} & \ding{55} & 97.0 & 99.0 & 99.0 & 9.0 & 100.0 & 71.0 & 92.0 & 81.0 \\
\midrule
Agent & \harness & CC & Opus 5 & \ding{55} & \ding{55} & \ding{55} & 100.0 & 100.0 & 100.0 & 100.0 & 100.0 & 99.0 & 94.0 & 99.0 \\
\bottomrule
\end{tabular}}
\end{subtable}

\vspace{6pt}
\begin{subtable}{\textwidth}
\centering
\caption{LIBERO-PRO}
\label{tab_main_liberopro}
\resizebox{\textwidth}{!}{%
\begin{tabular}{llllcccC{0.72cm}C{0.72cm}C{0.72cm}C{0.72cm}C{0.72cm}C{0.72cm}C{0.72cm}C{0.72cm}C{0.95cm}C{0.95cm}}
\toprule
\multicolumn{2}{c}{\multirow{3}{*}{Method}} & \multicolumn{5}{c}{Setting} & \multicolumn{10}{c}{Success Rate (\%)} \\
\cmidrule(lr){3-7} \cmidrule(lr){8-17}
& & \multirow{2}{*}{Agent} & \multirow{2}{*}{Model (Effort)} & \multirow{2}{*}{Train.} & \multirow{2}{*}{Exp.} & \multirow{2}{*}{Oracle} & \multicolumn{2}{c}{Object} & \multicolumn{2}{c}{Spatial} & \multicolumn{2}{c}{Goal} & \multicolumn{2}{c}{LIBERO-10} & \multicolumn{2}{c}{Average} \\
\cmidrule(lr){8-9} \cmidrule(lr){10-11} \cmidrule(lr){12-13} \cmidrule(lr){14-15} \cmidrule(lr){16-17}
& & & & & & & P & T & P & T & P & T & P & T & 6 Cells & 8 Cells \\
\midrule
\multirow{3}{*}{\makecell[l]{Learned\\Policy}} & OpenVLA & -- & -- & \ding{51} & \ding{55} & \ding{55} & 0.0 & 0.0 & 0.0 & 0.0 & 0.0 & 0.0 & -- & -- & 0.0 & -- \\
 & $\pi_0$ & -- & -- & \ding{51} & \ding{55} & \ding{55} & 0.0 & 0.0 & 0.0 & 0.0 & 0.0 & 0.0 & -- & -- & 0.0 & -- \\
 & $\pi_{0.5}$ & -- & -- & \ding{51} & \ding{55} & \ding{55} & 17.0 & 1.0 & 20.0 & 1.0 & 38.0 & 0.0 & -- & -- & 12.8 & -- \\
\midrule
\multirow{3}{*}{\makecell[l]{Agent +\\Policy}} & Pigey & \makecell[l]{Pigey-\\Agent} & Opus 4.7 (off) & \ding{51} & \ding{55} & \ding{55} & 54.0 & 54.0 & 66.0 & 80.0 & 44.0 & 22.0 & -- & -- & 53.3 & -- \\
 & \multirow{2}{*}{Harness VLA} & Codex & GPT-5.5 (xhigh) & \ding{51} & \ding{51} & \ding{51} & 91.0 & 94.0 & 69.0 & 81.0 & 66.0 & 75.0 & 49.0 & 52.0 & 79.3 & 72.1 \\
 &  & CC & Opus 4.7 (max) & \ding{51} & \ding{51} & \ding{51} & 90.0 & 88.0 & 80.0 & 94.0 & 87.0 & 87.0 & 62.0 & 71.0 & 87.7 & 82.4 \\
\midrule
\multirow{4}{*}{Agent} & CaP-X & \makecell[l]{CaP-\\Agent0} & \makecell[l]{GPT-5.2 (none)\\Opus 4.5 (off)\\Gemini 3 Pro (high)} & \ding{55} & \ding{55} & \ding{55} & 22.0 & 18.0 & 12.0 & 14.0 & 26.0 & 17.0 & -- & -- & 18.2 & -- \\
 & RHO & Codex & GPT-5.5 (xhigh) & \ding{55} & \ding{51} & \ding{55} & 12.2 & 37.2 & 53.0 & 61.6 & 50.6 & 55.6 & -- & -- & 45.0 & -- \\
 & ASPIRE & CC & Opus 4.6 (high) & \ding{55} & \ding{51} & \ding{55} & 98.0 & 95.0 & 51.0 & 60.0 & 81.0 & 45.0 & -- & -- & 71.7 & -- \\
 & RATS & \makecell[l]{RATS-\\Agent} & \makecell[l]{Gemini 3.1 Pro\\(high)} & \ding{55} & \ding{51} & \ding{51} & 61.0 & 63.0 & 29.0 & 31.0 & 43.0 & 36.0 & -- & -- & 43.8 & -- \\
\midrule
Agent & \harness & CC & Opus 5 & \ding{55} & \ding{55} & \ding{55} & 98.0 & 97.0 & 98.0 & 100.0 & 79.0 & 79.0 & 80.0 & 65.0 & 91.8 & 87.0 \\
\bottomrule
\end{tabular}}
\end{subtable}
\vspace{-1.5em}

\end{table}

\parabf{Agents and Models.}
We evaluate \harness with Claude Code powered by Claude Opus~5, Sonnet~5, and Fable~5.1, all at high reasoning effort, and with Codex powered by GPT-6 Astra at medium and high reasoning effort.
Each episode starts with a fresh workspace, without robot demonstrations, task-specific rehearsal, or memory from earlier episodes.
The agent has no access to privileged simulator state or the benchmark's success predicate (see~\cref{app_execution_protocol} for the evaluation protocol). 

\subsection{Main Results}
\label{sec_main}

\cref{tab_main} compares \harness on CaP-Bench and LIBERO-PRO against four types of systems: human-written programs, learned visuomotor policies, agents that orchestrate learned policies, and agents that act as the policy through predefined primitives.
We report the success rates of \harness alongside the numbers that prior systems report in their papers.
For each system, we also mark whether its policy is trained on demonstrations, whether it acts zero-shot on the evaluation tasks or draws on experience accumulated beforehand, and whether it can query the benchmark's success predicate during execution.
More evaluation settings are detailed in~\cref{app_compared}.

\parabf{Robot-use agents are zero-shot visuomotor policies.}
With Claude Code powered by Claude Opus~5 (high), \harness achieves success rates of 99.0\% on CaP-Bench and 87.0\% on LIBERO-PRO.
It exceeds ASPIRE's 81.0\% and Harness VLA's 82.4\%, without robot-policy training, prior task experience, or access to the benchmark's success predicate during execution.
These results show that an off-the-shelf coding agent can serve as a zero-shot visuomotor policy under our zero-abstraction design, without a learned robot policy or task-specific primitive library.
On Nut, \harness achieves 100\% success, compared with 0--9\% for CaP-X, RHO, and ASPIRE despite their synthesized skills or prior optimization.
Accumulating skills over predefined primitives does not supply sensing capabilities absent from their interface, whereas native ROS~2 access allows the agent to incorporate force feedback directly into its insertion program.
Detailed comparisons and analysis are provided in~\cref{app_main_results,app_native_interface_analysis}.
We further evaluate the generalization of \harness on RoboCasa365 in~\cref{sec_generalization}.

\subsection{How the Agent Perceives}
\label{sec_perception}

\begin{wrapfigure}{r}{0.5\linewidth}
    \vspace{-1em}
    \centering
    \includegraphics[width=0.92\linewidth]{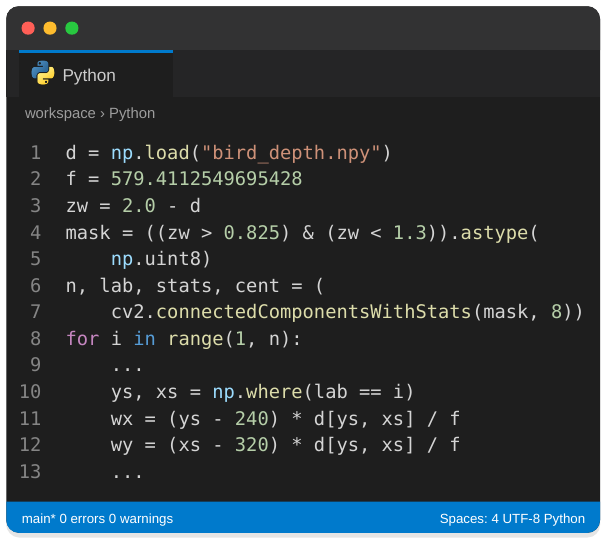}
    \vspace{-0.5em}
    \caption{\textbf{An agent-written program for locating the nut and pegs from depth}. See~\cref{app_programs} for further code details.}
    \label{fig_metric_code}
    \vspace{-12pt}
\end{wrapfigure}

\begin{figure}[t]
\centering
\begin{subfigure}[t]{0.24\linewidth}\includegraphics[width=\linewidth]{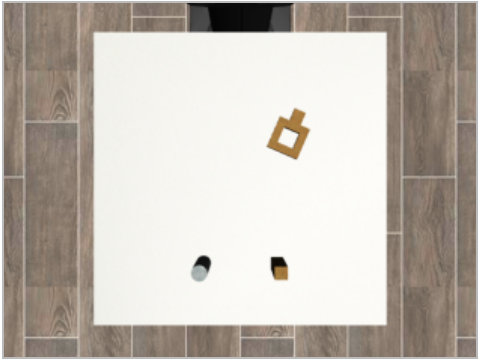}\caption{Overhead camera image.}
\label{fig_perception_case_a}
\end{subfigure}\hfill
\begin{subfigure}[t]{0.24\linewidth}\includegraphics[width=\linewidth]{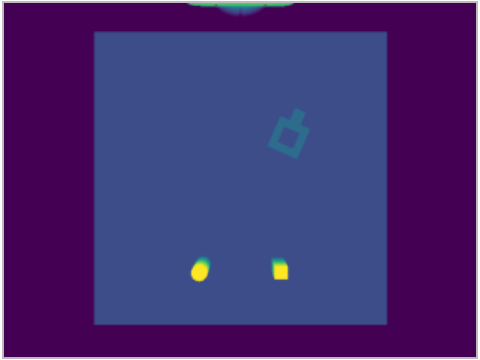}\caption{Depth array segmented by height.}
\label{fig_perception_case_b}
\end{subfigure}\hfill
\begin{subfigure}[t]{0.24\linewidth}\includegraphics[width=\linewidth]{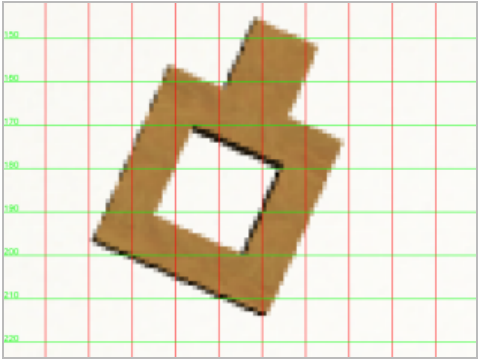}\caption{Enlarged crop with a pixel grid.}
\label{fig_perception_case_c}
\end{subfigure}\hfill
\begin{subfigure}[t]{0.24\linewidth}\includegraphics[width=\linewidth]{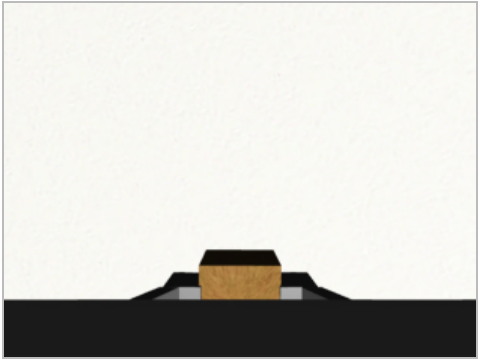}\caption{Wrist camera image after grasping.}
\label{fig_perception_case_d}
\end{subfigure}
\vspace{-0.5em}

\caption{\textbf{An example of agent-built perception in the CaP-Bench Nut task.}
The agent combines visual understanding with numerical processing:
(a) it reads the overhead camera image to identify the nut and two pegs;
(b) it segments the depth array by height and uses camera calibration to compute their positions in world coordinates;
(c) it enlarges a crop of the nut and adds a pixel grid to inspect the handle's orientation and width;
and (d) it checks the grasp using the wrist camera image together with the measured finger gap.}
\label{fig_perception_case}

\end{figure}

\begin{figure}[t]
    \centering
    \begin{subfigure}[t]{0.24\linewidth}
    \includegraphics[viewport=12.43 25.10 196.70 163.48,clip,width=\linewidth]{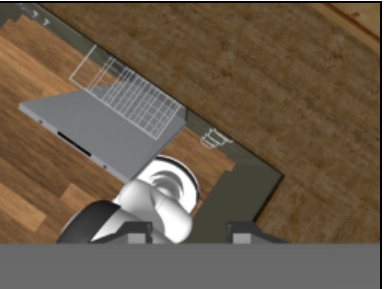}
    \caption{Wrist-camera image.}
    \label{fig_perception_case_cloud_a}
    \end{subfigure}\hfill
    \begin{subfigure}[t]{0.24\linewidth}
    \includegraphics[viewport=206.90 25.10 391.17 163.48,clip,width=\linewidth]{figures/perception_case_cloud.pdf}
    \caption{Depth points colored by height.}
    \label{fig_perception_case_cloud_b}
    \end{subfigure}\hfill
    \begin{subfigure}[t]{0.24\linewidth}
    \includegraphics[viewport=401.37 25.10 585.64 163.48,clip,width=\linewidth]{figures/perception_case_cloud.pdf}
    \caption{Point cloud from a chosen viewpoint.}
    \label{fig_perception_case_cloud_c}
    \end{subfigure}\hfill
    \begin{subfigure}[t]{0.24\linewidth}
    \includegraphics[viewport=595.84 25.10 780.11 163.48,clip,width=\linewidth]{figures/perception_case_cloud.pdf}
    \caption{Point-cloud cross-sections.}
    \label{fig_perception_case_cloud_d}
    \end{subfigure}
    \vspace{-0.5em}

    \caption{\textbf{The agent writes its own code to reconstruct scene geometry from depth and create views tailored to its own visual understanding.}
    In a RoboCasa365 dishwasher task, (a) it reads the wrist-camera image to identify the rack and its rail.
    (b) It projects depth points onto the counter plane and colors them by height.
    (c) It renders the reconstructed point cloud from a chosen viewpoint to inspect the rack's structure.
    (d) It slices the point cloud along one axis and plots cross-sections to locate the rail along which the rack moves.}
    \vspace{-1em}
    \label{fig_perception_case_cloud}
    \end{figure}

We analyze the agent's programs and execution traces across 1,500 episodes on CaP-Bench and LIBERO-PRO to examine how it develops perception strategies during task execution.
\cref{tab_perception} reports how often the agent processes sensor images with self-written code or runs self-written metric programs;
\cref{fig_perception_case,fig_metric_code} show a concrete example from the CaP-Bench Nut task, while \cref{fig_perception_case_cloud} illustrates how the agent reconstructs scene geometry from depth and prepares views for its own inspection in RoboCasa365.
Additional analysis is provided in~\cref{app_behavior}.

\begin{wraptable}{r}{0.56\linewidth}
\centering
\footnotesize
\setlength{\tabcolsep}{3pt}
\renewcommand{\arraystretch}{1.15}
\caption{\textbf{The agent writes its own programs to process sensor images and compute metric measurements.}
Percentage of episodes in which Claude Code with Opus 5 processes sensor images with self-written code or runs a self-written metric program. All successful and failed episodes are included. More analysis is provided in~\cref{app_behavior}.}
\vspace{-0.5em}

\label{tab_perception}
\begin{tabular*}{\linewidth}{@{\extracolsep{\fill}}lrcc@{}}
\toprule
Benchmark & Episodes & \makecell{Image\\processing (\%)} & \makecell{Metric\\programs (\%)} \\
\midrule
CaP-Bench & 700 & 47.43\% & 97.43\% \\
LIBERO-PRO & 800 & 94.75\% & 96.25\% \\
Overall & 1,500 & 72.67\% & 96.80\% \\
\bottomrule
\end{tabular*}

\end{wraptable}

\parabf{The agent builds its own image-processing tools for better visual understanding.}
The built-in visual capabilities of frontier models enable the agent to identify task-relevant objects, assess spatial relationships, and check action outcomes.
Yet interestingly, our analysis reveals that the agent develops its own image-processing tools to better understand the scene, spontaneously writing programs that transform raw sensory inputs into views tailored to the task.
We observe this emergent behavior in 72.67\% of episodes across CaP-Bench and LIBERO-PRO (\cref{tab_perception}), indicating that programmatic image processing is a common part of the agent's perception.
For example, in~\cref{fig_perception_case}, the agent (a) identifies the nut and pegs in the overhead image, then (c) writes a program to crop the nut region from the image, enlarge it, and overlay a pixel grid, making the handle's orientation and width easier to inspect.
Beyond processing RGB images, the agent also reconstructs the dishwasher rack's point clouds from depth and renders them from chosen viewpoints to inspect object geometry (\cref{fig_perception_case_cloud}).
These views let it inspect geometry that is difficult to judge from the original camera image alone.
\cref{app_visual_understanding} offers further examples and analysis.

\parabf{The agent builds its own tools for metric measurement to guide physical actions.}
Beyond visual understanding, manipulation requires numerical measurements  such as target positions, object dimensions, and contact forces.
Crucially, the agent constructs its own measurement programs during task execution, combining depth arrays with camera intrinsics and frame transforms to derive task-specific measurements.
We observe this emergent behavior in 96.80\% of episodes across CaP-Bench and LIBERO-PRO (\cref{tab_perception}), showing that self-written measurement programs are a common part of its approach to manipulation.
For example, in~\cref{fig_perception_case}(b), the agent writes a Python program to segment the nut and pegs by height and compute their positions in world coordinates from the depth array, as shown in~\cref{fig_metric_code}.
The resulting coordinates provide motion targets, while measurements such as the gripper's finger gap help the agent check execution outcomes.
\cref{app_metric_measurement} provides further details and examples.

\parabf{The agent autonomously pulls observations on demand.}
Rather than passively receiving sensory streams that continuously fill its context window, the agent actively decides which observations to fetch and when, based on the task and execution feedback.
For example, in~\cref{fig_perception_case}, the agent selectively captures (a) an overhead image to locate the nut and pegs and (d) an eye-in-hand image to verify the grasp, rather than reading every camera frame throughout execution.
In this way, the agent controls what sensory information enters its context, using selected observations to verify progress and guide subsequent actions.
Further details are provided in~\cref{app_observation_acquisition}.

\subsection{How the Agent Manipulates}
\label{sec_behavior}

\begin{wraptable}{R}{0.56\linewidth}
\centering
\footnotesize
\setlength{\tabcolsep}{3pt}
\renewcommand{\arraystretch}{1.15}

\caption{\textbf{The agent writes its own motion-control clients and feedback-control programs.}
Percentage of episodes in which Claude Code with Opus 5 writes a motion-control client or executes a self-written program that uses sensor feedback to adjust motion or determine when to stop. All successful and failed episodes are included. More analysis is provided in~\cref{app_behavior}.}
\label{tab_manipulation}
\vspace{-0.5em}

\begin{tabular*}{\linewidth}{@{\extracolsep{\fill}}lrcc@{}}
\toprule
Benchmark & Episodes & \makecell{Motion-control\\clients (\%)} & \makecell{Feedback-control\\programs (\%)} \\
\midrule
CaP-Bench & 700 & 97.00\% & 60.14\% \\
LIBERO-PRO & 800 & 94.88\% & 41.38\% \\
Overall & 1,500 & 95.87\% & 50.13\% \\
\bottomrule
\end{tabular*}

\end{wraptable}

\begin{figure}[t]
\centering
\begin{subfigure}[t]{0.24\linewidth}\includegraphics[width=\linewidth]{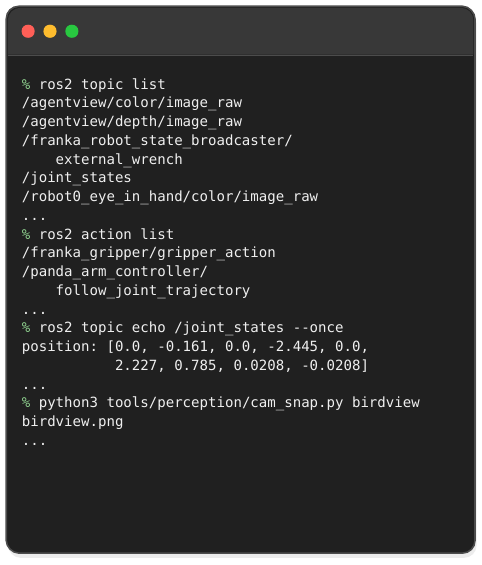}\caption{Inspect interfaces.}\end{subfigure}\hfill
\begin{subfigure}[t]{0.24\linewidth}\includegraphics[width=\linewidth]{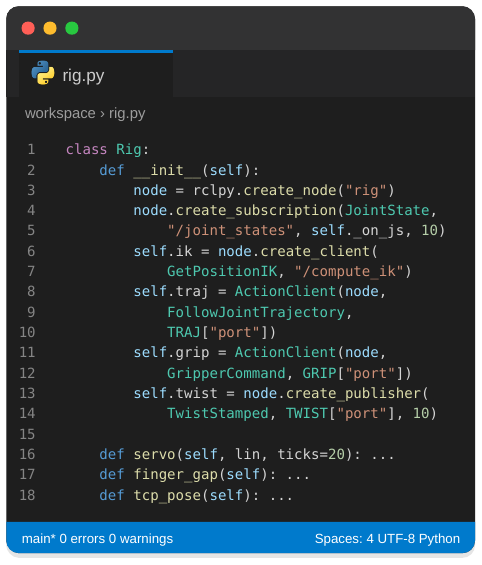}\caption{Write a client library.}\end{subfigure}\hfill
\begin{subfigure}[t]{0.24\linewidth}\includegraphics[width=\linewidth]{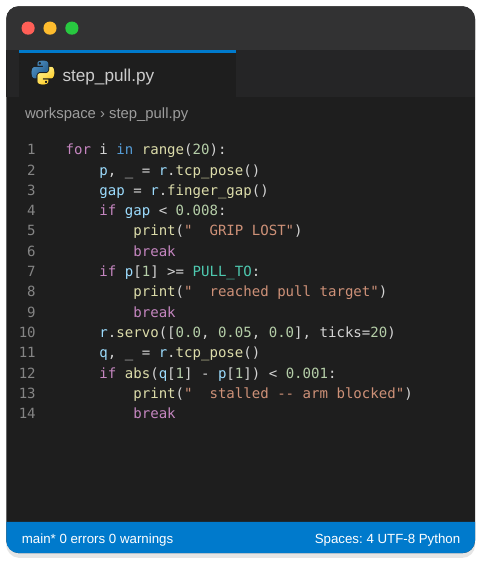}\caption{Pull with feedback.}\end{subfigure}\hfill
\begin{subfigure}[t]{0.24\linewidth}\includegraphics[width=\linewidth]{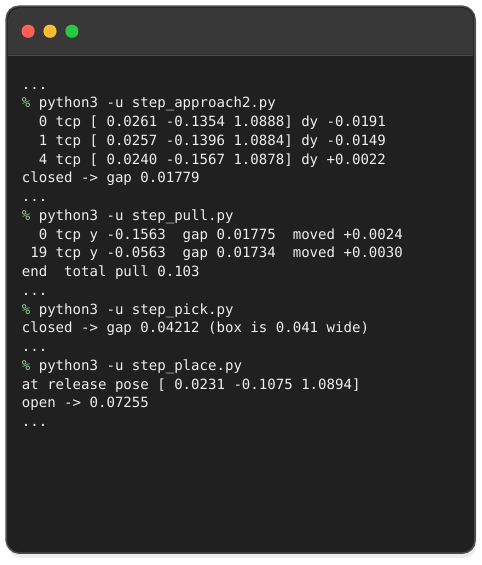}\caption{Inspect results.}\end{subfigure}
\vspace{-0.5em}

\caption{\textbf{The agent constructs motion-control clients and feedback logic during a LIBERO-PRO drawer task.}
(a) It inspects the robot's interfaces and acquires observations.
(b) It writes a reusable client library for inverse kinematics, trajectory and gripper actions, and velocity commands.
(c) It writes a pulling program that checks finger gap and displacement, stopping if the grip is lost or motion stalls.
(d) It executes its programs and reads the resulting measurements to assess progress.
Some lines are omitted to highlight the key operations.
See~\cref{app_programs} for the programs and~\cref{app_manipulation_execution} for execution details.}
\label{fig_drawer_strip}
\vspace{-0.5em}
\end{figure}

\begin{figure}[t]
\centering
\begin{subfigure}[t]{0.24\linewidth}\includegraphics[width=\linewidth]{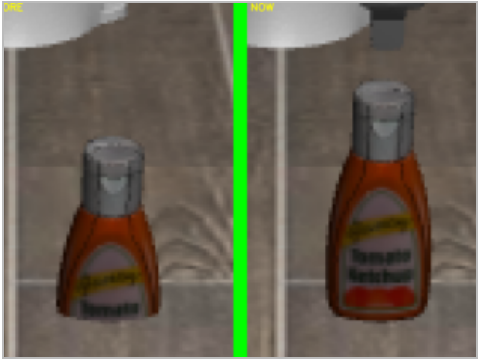}\caption{Check scene motion.}\end{subfigure}\hfill
\begin{subfigure}[t]{0.24\linewidth}\includegraphics[width=\linewidth]{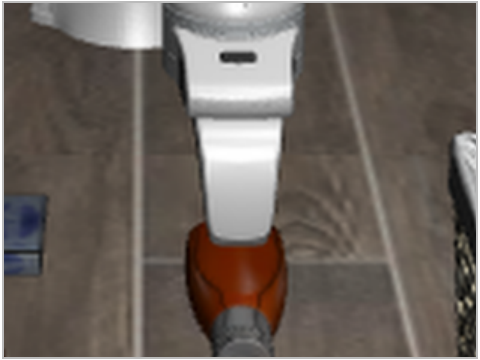}\caption{First grasp fails.}\end{subfigure}\hfill
\begin{subfigure}[t]{0.24\linewidth}\includegraphics[width=\linewidth]{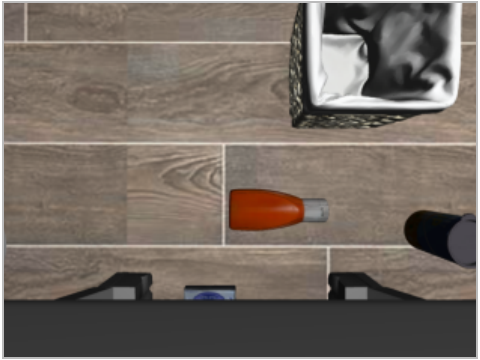}\caption{Remeasure the fallen bottle.}\end{subfigure}\hfill
\begin{subfigure}[t]{0.24\linewidth}\includegraphics[width=\linewidth]{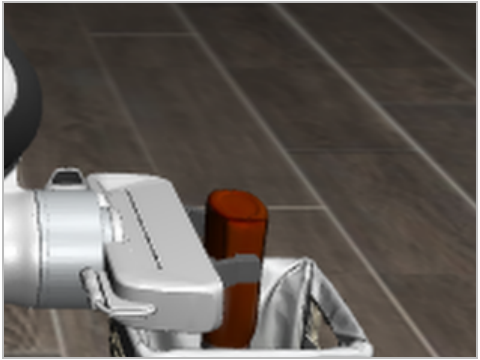}\caption{Place the bottle in the basket.}\end{subfigure}
\vspace{-0.5em}
\caption{\textbf{The agent revises its grasp after a failed attempt in a LIBERO-PRO task.}
(a) It compares camera frames and detects that the scene has not settled.
(b) Its first grasp misses and knocks the ketchup bottle over; finger-gap readings and a camera view reveal the failure.
(c) It uses a wrist-camera observation and a reconstructed point cloud to estimate the fallen bottle's orientation and plan a new grasp.
(d) It places the bottle upright in the basket and checks the result.
The panels show RGB-D observations processed by agent-written code.}
\label{fig_recovery_case}
\vspace{-1em}
\end{figure}

We analyze the agent's programs and execution traces across the same 1,500 CaP-Bench and LIBERO-PRO episodes to examine how it constructs control tools and responds to execution feedback.
\cref{tab_manipulation} reports how often the agent writes motion-control clients and executes self-written feedback-control programs; \cref{fig_drawer_strip,fig_recovery_case} show examples of tool construction and recovery from a failed action.
Additional analysis and examples are provided in~\cref{app_behavior}.

\parabf{The agent builds its own motion-control tools using native ROS~2 interfaces.}
The supplied starter tools provide ready-to-run implementations of basic robot motions (\cref{sec:method}).
Interestingly, the agent develops its own control clients during task execution, writing \texttt{rclpy} programs that access trajectory, gripper, velocity, and inverse-kinematics interfaces.
We find agent-written motion clients in 95.87\% of episodes across CaP-Bench and LIBERO-PRO (\cref{tab_manipulation}), indicating that constructing control tools is a common part of the agent's work.
For example, in~\cref{fig_drawer_strip}, the agent (a) inspects the robot's native interfaces and (b) writes a client library that it imports into subsequent action programs.
This library lets it compose robot operations in reusable code as it works through the task.
\cref{app_programs} provides the corresponding client and action programs.

\parabf{The agent writes its own feedback-control programs to adjust physical actions.}
Beyond issuing individual motion commands, the agent constructs programs that repeatedly read sensor data and use those readings to adjust motion or determine when to stop.
Crucially, the feedback conditions are encoded in the agent's own control logic and evaluated by the running program.
We observe executed programs with such feedback logic in 50.13\% of episodes across CaP-Bench and LIBERO-PRO (\cref{tab_manipulation}).
For example, in~\cref{fig_drawer_strip}(c), the agent writes a program that pulls the drawer in velocity bursts and checks the finger gap and displacement after each burst, stopping if the gap falls below 8\,mm or the drawer ceases to move.
In another episode, an agent-written descent program stops when the change in measured force exceeds its threshold (\cref{fig_app_loops}(a)).
These examples show task-specific feedback logic emerging in programs written during execution.
\cref{app_manipulation_execution} provides the control traces, feedback conditions, and further examples.

\parabf{The agent revises its actions when execution feedback reveals a failure.}
Alongside feedback within its programs, the agent inspects terminal output and new observations between actions to decide whether to continue or revise its approach.
Importantly, its checks extend beyond whether a command completed to whether the intended physical change occurred.
For example, in~\cref{fig_recovery_case}, the agent (b) misses the ketchup bottle during its initial grasp and knocks it over.
It detects the failure from the finger gap and camera observations, then (c) reconstructs the fallen bottle's geometry from depth and plans a new grasp based on its orientation.
Finally, it (d) grasps the bottle and places it upright in the basket.
The recovery is worked out during execution from the changed scene, rather than supplied as a task-specific recovery routine.
Further cases and an analysis of unsuccessful episodes are provided in~\cref{app_episodes,app_failure}.

\subsection{Generalization under Distribution Shift}
\label{sec_generalization}

\cref{tab_robocasa} reports success rates on the 50 target tasks of RoboCasa365, which the benchmark splits into 18 atomic tasks, 16 composite tasks whose templates appear in its pretraining set (Seen), and 16 composite tasks whose templates are held out of it (Unseen); protocol details are in \cref{app_protocol}.

\parabf{\harness maintains performance on unseen tasks, demonstrating generalization without relying on a learned robot policy.}
\harness achieves comparable success rates on Seen and Unseen composite tasks, at 26.3\% and 28.1\%, respectively.
In contrast, RLDX-1 drops from 21.3\% to 5.0\%, while Harness VLA, which orchestrates the RLDX-1 with a coding agent, drops from 61.0\% to 13.8\%.
These results suggest that reliance on a learned robot policy can limit generalization beyond its training distribution, whereas \harness handles both task groups by synthesizing and refining control programs during execution rather than relying on motor skills learned from robot demonstrations.
Detailed comparisons and analysis are provided in~\cref{app_generalization}.

\begin{table}[t]
\centering
\footnotesize
\setlength{\tabcolsep}{2pt}
\renewcommand{\arraystretch}{1.15}
\caption{\textbf{\harness generalizes to unseen composite tasks in RoboCasa365.}
Success rates (\%) are reported for atomic tasks and composite tasks whose templates are included in (Comp.-Seen) or held out from (Comp.-Unseen) the benchmark's pretraining data.
Detailed evaluation protocols are provided in~\cref{app_protocol}.
}
\label{tab_robocasa}
\resizebox{\textwidth}{!}{%
\begin{tabular}{llllccccccc}
\toprule
\multicolumn{2}{c}{\multirow{2}{*}{Method}} & \multicolumn{5}{c}{Setting} & \multicolumn{4}{c}{Success Rate (\%)} \\
\cmidrule(lr){3-7} \cmidrule(lr){8-11}
& & Agent & Model (Effort) & Train. & Exp. & Oracle & Atomic-Seen & Comp.-Seen & Comp.-Unseen & Overall \\
\midrule
\multirow{3}{*}{\makecell[l]{Learned\\Policy}} & \multirow{2}{*}{GR00T N1.5} & -- & -- & \ding{51} & \ding{55} & \ding{55} & 41.9 & 0.0 & 0.2 & 15.1 \\
 &  & -- & -- & \ding{51} & \ding{51} & \ding{55} & 68.5 & 40.6 & 42.1 & 51.1 \\
 & RLDX-1 & -- & -- & \ding{51} & \ding{55} & \ding{55} & 60.0 & 21.3 & 5.0 & 30.0 \\
\midrule
\multirow{2}{*}{\makecell[l]{Agent +\\Policy}} & \multirow{2}{*}{Harness VLA} & Codex & GPT-5.5 (xhigh) & \ding{51} & \ding{51} & \ding{51} & 92.0 & 61.0 & 13.8 & 57.1 \\
 &  & CC & Opus 4.7 (max) & \ding{51} & \ding{51} & \ding{51} & 79.4 & 47.5 & 15.0 & 48.6 \\
\midrule
Agent & \harness & CC & Opus 5 & \ding{55} & \ding{55} & \ding{55} & 61.1 & 26.3 & 28.1 & 39.4 \\
\bottomrule
\end{tabular}}
\end{table}

\begin{figure}[t]
\centering

\begin{subfigure}[t]{0.49\linewidth}\includegraphics[width=\linewidth]{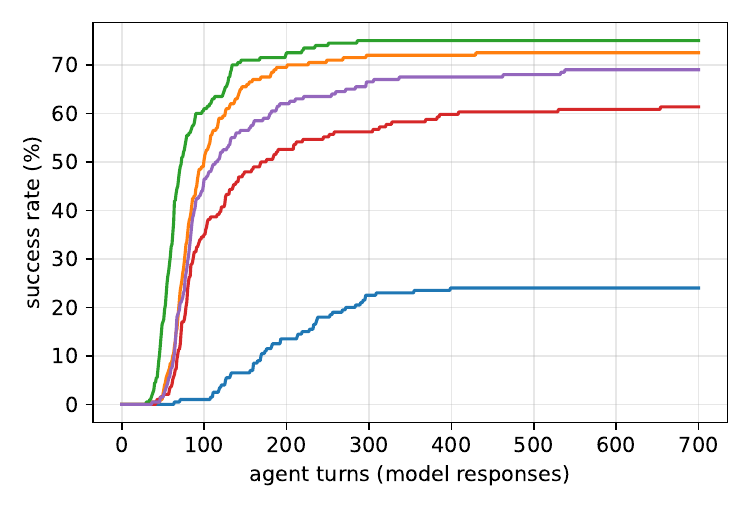}\vspace{-0.5em}\caption{agent-turn budget}\label{fig_scaling_turns}\end{subfigure}\hfill
\begin{subfigure}[t]{0.49\linewidth}\includegraphics[width=\linewidth]{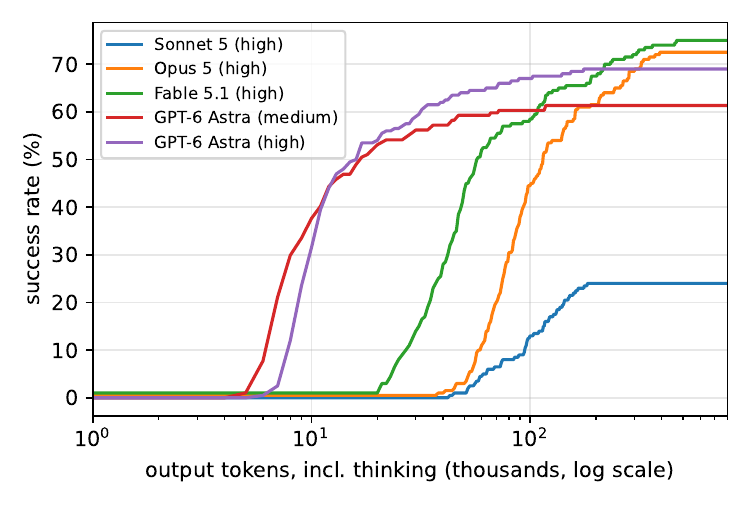}\vspace{-0.5em}\caption{output-token budget (log scale)}\label{fig_scaling_tokens}\end{subfigure}
\vspace{-0.5em}

\caption{\textbf{Success improves with test-time computation but plateaus at different levels across models.}
Success rates on the LIBERO-10 suite of LIBERO-PRO are plotted against agent-turn and output-token budgets, with 200 trials per configuration across both perturbation cells.
A trial counts as solved within a budget if the benchmark predicate first held before it; one turn is one model response, and tokens are those the model wrote, including thinking.
The right end of each curve is the success rate reported in \cref{tab_scaling}; the runs behind the curves, and how the token position of a Claude Code episode is estimated, are described in \cref{app_scaling}.}
\label{fig_scaling}
\vspace{-1.5em}

\end{figure}

\subsection{Scaling with Coding Agents and Models}
\label{sec_scaling}

\cref{fig_scaling} shows how success rates change with increasing test-time computation (\textit{e.g.}, agent turns, output tokens, reasoning effort) across coding agents and models on the LIBERO-10 suite of LIBERO-PRO; detailed interaction and cost profiles are provided in~\cref{app_scaling}.

\parabf{Test-time computation improves robot control through a zero-abstraction interface.}
Increasing the agent-turn budget from 100 to 200 raises the success rate of Claude Code with Opus~5 from 50\% to 70\% (\cref{fig_scaling_turns}).
Increasing reasoning effort also improves performance: Codex with GPT-6 Astra reaches 69.0\% at high effort, compared with 62.5\% at medium effort.
Both agents use the same workspace and native interface, so these gains require no additional tools or primitives to be engineered in advance.
The gains nevertheless diminish as the curves flatten, while differences between models persist: Sonnet~5 reaches 24.0\% despite continuing to gain successes up to about 400 turns, compared with 72.5\% for Opus~5 and 75.0\% for Fable~5.1.
Thus, more interaction and greater reasoning effort improve performance under our native, zero-abstraction interface, while the achievable performance remains dependent on the underlying model.
Detailed comparisons of interaction patterns, costs, and per-task results are provided in~\cref{app_scaling}.

\subsection{Failure Study}
\label{sec_failure}
We analyze the failed episodes of \harness with Claude Code and Opus~5 across all three benchmarks, assigning each to the stage at which the decisive error lies.
The study covers 413 failed episodes (\cref{tab_failure}).
The classifications are based on transcripts, workspace files, and benchmark predicates; the detailed procedure and classification limitations are provided in~\cref{app_failure}.

\parabf{Task criteria and undocumented machine behavior account for many failures.}
Classes A, B, and C account for 238 of the 413 failures (58\%).
The agent may complete its interpretation of an instruction while missing a numerical threshold or spatial region required by the benchmark predicate.
For example, in 14 drawer episodes, it pulls the drawer 62--136\,mm and stops, whereas the predicate requires more than 140\,mm.
Other episodes involve conflicting task criteria or budgets exhausted while handling undocumented machine behavior, such as a mobile base displaced by contact forces or a freezer door that closes while the agent fetches another tray.
These cases expose a gap between task instructions, machine behavior, and completion criteria in current benchmarks.

\begin{table}[t]
\centering
\footnotesize
\setlength{\tabcolsep}{4pt}
\renewcommand{\arraystretch}{1.1}
\caption{\textbf{Task criteria and undocumented machine behavior account for over half of the analyzed failures.}
Counts of failed episodes with Claude Code and Opus~5 across CaP-Bench, LIBERO-PRO, and RoboCasa365, assigned one category each by transcript and workspace analysis.
Classification procedures, limitations, and sub-causes are provided in~\cref{app_failure,tab_failure_detail}.}
\label{tab_failure}
\resizebox{\textwidth}{!}{%
\begin{tabular}{lrrrr}
\toprule
Cause of failure & LIBERO-PRO & CaP-Bench & RoboCasa365 & All \\
\midrule
A\; Success criterion not stated in the instruction & 42 & 1 & 36 & 79 \\
B\; Predicate contradicts the instruction & 17 & 0 & 2 & 19 \\
C\; Budget exhausted while handling undocumented machine behavior & 15 & 0 & 125 & 140 \\
D\; Object or force beyond the machine's limits & 12 & 0 & 63 & 75 \\
E\; Perception, control, or waiting error of the agent & 18 & 6 & 76 & 100 \\
\midrule
Analyzed failed episodes & 104 & 7 & 302 & 413 \\
\bottomrule
\end{tabular}}
\end{table}

\parabf{Physical constraints remain a challenge for agent-written programs.}
Class D accounts for 75 failures (18\%), involving gripper geometry, contact forces, and reachability.
Examples include object widths near or beyond the gripper opening, thin objects squeezed out during transport, and dropped objects that the arm cannot reach.
The same tasks can succeed under other initial conditions, so these cases identify limited operating margins rather than tasks that are universally impossible.

\parabf{The agent also makes control, perception, and execution-management errors.}
Class E accounts for 100 failures (24\%): 46 control errors, 27 perception errors, and 27 waiting errors.
Control errors include an incorrect gripper-width command despite a measured handle width, an inverse-kinematics branch change that sends the hand below the table, and releasing an object before the arm arrives.
Perception errors include failing to search for a target outside the current view, while waiting errors include spending the budget polling background jobs.
Thus, the ability to write measurement and feedback programs does not ensure that the agent uses them correctly or verifies every required condition.
Detailed sub-causes and case sequences are provided in~\cref{app_failure}.

\section{Limitations}
\label{sec:limitations}

Although ROS~2 is a de facto standard used in both simulation and on physical robots~\citep{macenski2022ros2}, a shared software interface alone does not establish that our results generalize to real-world deployment.
Our evaluation is limited to simulation; experiments on physical robots are needed to assess the practical viability of \harness, particularly in three respects.

\parabf{Timing.}
We follow common simulation evaluation protocols~\citep{fu2026capx, galanti2026pigey, kim2024openvla}, in which simulation time advances during action execution but pauses during agent deliberation (\cref{app_execution_protocol}).
Thus, our evaluation does not capture the effects of inference latency on a physical environment that continues to evolve during deliberation.
Agent-written feedback controllers can react to sensor readings within a running program, but do not eliminate delays between agent-level decisions.
Real-time responsiveness remains to be evaluated, especially for dynamic tasks.

\parabf{Sensing.}
Although the agent receives raw simulated sensor data rather than privileged object states, these observations do not capture the full variability of real-world sensing.
Sensor noise, missing depth values, and calibration errors may affect both visual interpretation and programmatic measurement.
The agent's demonstrated perception capabilities in simulation therefore do not establish robustness to real-world sensor data.
Physical-robot experiments are needed to test whether agent-written perception programs can detect and accommodate such errors.

\parabf{Safety.}
Task success does not establish safe execution: our failure analysis includes motions that would collide with the table.
Although agents sometimes write their own force thresholds and stopping conditions, these protections are not guaranteed to be present or correct.
Physical deployment would require independently enforced safety controls, such as motion limits, collision checks, and emergency stops, rather than relying solely on agent-written safeguards.
Validating these controls while retaining the agent's flexibility remains future work.

\section{Conclusion}

We presented \harness, a zero-abstraction harness that provides off-the-shelf coding agents with terminal access to the robot's native ROS~2 interface.
Its minimalist \textit{workspace-as-harness} design places machine documentation and basic tools in a workspace, leaving the agent to organize its own work without a prescribed agentic workflow.
It recasts perception as file I/O and manipulation as coding.
Experiments show that coding agents can serve as zero-shot visuomotor policies without learned robot policies or task-specific primitive libraries.
Crucially, image-processing tools, metric measurement programs, motion-control clients, and feedback controllers emerge as the agent writes code during execution.
The findings suggest that robot-use agents need not depend on an expanding set of human-engineered abstractions; the robot's existing software stack can support agents in constructing the tools they need.
Our work highlights the potential for continued advances in coding agents to unlock increasingly general capabilities, extending from software tasks to interaction with the physical world.

\clearpage
\bibliography{references}
\bibliographystyle{iclr2027_conference}

\clearpage
\appendix
\appendix

\section{Extended Related Work}
\label[appendix]{app_extended_related_work}

\parabf{Visuomotor Policies.}
A visuomotor policy maps raw sensory observations, such as camera images and proprioceptive states, directly to low-level robot actions~\citep{levine2016visuomotor}.
Today such policies are predominantly learned end to end: VLA models attach an action head to a pretrained vision-language backbone and fine-tune it on robot demonstrations~\citep{brohan2023rt2, kim2024openvla, google2025gemini_robotics, bjorck2025gr00t, intelligence2026pi07}, and WAM variants extend this line by jointly predicting future observations and actions~\citep{agarwal2026cosmos3, ye2026world, li2026causal}.
However, they require large-scale demonstration data, and their generalization remains bounded by the training distribution: performance drops sharply under perturbations of objects, lighting, and poses~\citep{zhou2026libero_pro, fei2025libero_plus, kim2026libero_para}.
\harness instead casts an off-the-shelf coding agent as a zero-shot visuomotor policy, requiring no robot demonstration data or fine-tuning.

\parabf{Robot-Use Agents.}
RUA~\citep{isola2026rua} refers to LLM agents that operate a robot the way they operate a computer, a paradigm that traces back to CaP, in which an LLM writes executable programs as robot policies~\citep{liang2023cap, singh2022progprompt}.
By composing predefined perception and control primitives, CaP brings the LLM's world knowledge about geometry, physics, and programming to robot control without any robot demonstrations~\citep{kumar2026aor}.
Driven by frontier coding agents, today's RUA systems have extended CaP from one-shot code generation to an agentic loop: the agent acts through tool calls or generated programs, observes the outcome, and revises its next action~\citep{google2025gemini_robotics, fu2026capx, robocurve2026astra, berman2026claudeplaysrobotics}.

These systems broadly fall into two categories based on the interface through which the agent acts.
One line of work wraps learned visuomotor policies into callable tools, yielding a hierarchical architecture in which the high-level LLM agent decomposes tasks and selects tools while the low-level policy handles physical execution~\citep{ichter2023saycan, zhang2026harnessvla, galanti2026pigey, fang2026emerge, chen2026volo}.
However, delegating low-level execution to a learned policy can limit the agent's ability to diagnose and correct failures in out-of-distribution scenes~\citep{chen2026show_harness, zhao2026hymes, ding2026zetta, huang2026roboharness}.
Another line of work exposes low-level actions directly to the agent, which itself serves as the robot policy~\citep{jia2026agp, berman2026claudeplaysrobotics, robocurve2026astra, su2026astra}, either by issuing end-effector poses step by step~\citep{hu2026via, chen2026show_harness, liu2026guava, zhou2026minimal_interface} or by writing the control code with predefined primitives~\citep{fu2026capx, lu2026aspire, elmaaroufi2026rho}.
Nevertheless, these systems either read privileged simulator state (\textit{e.g.}, ground-truth object poses) that is unavailable on real hardware~\citep{singh2025malmm, tsui2026faea, xie2026pusht}, or rely on a custom, hand-crafted library of perception and control primitives (\textit{e.g.}, pose estimators and grasp samplers), which requires re-engineering for each new embodiment~\citep{fu2026capx, seker2026gta2, zhang2026rats}.
\harness instead provides the coding agent with terminal access to the robot's native ROS~2 interfaces and a workspace, allowing it to construct perception and control programs during execution.
ROS~2 is a de facto standard used in both simulation and on physical robots~\citep{macenski2022ros2}; our evaluation is limited to simulation, as discussed in~\cref{sec:limitations}.

\parabf{CLI Coding Agents.}
CLI coding agents, exemplified by Claude Code~\citep{anthropic2025claudecode} and Codex~\citep{openai2025codexcli}, have emerged as general-purpose agents that resolve real-world software engineering tasks, as measured by benchmarks such as SWE-bench~\citep{openai2024swebenchverified} and Terminal-Bench~\citep{merrill2026terminalbench}.
These agents exhibit strong closed-loop agency over long horizons: they navigate large, complex codebases, form and test hypotheses by editing and running code, issue CLI commands and interpret terminal output, and recover from their own errors through hours of autonomous iteration.
Recently, this competence has extended beyond software engineering to operating desktop and web applications~\citep{yang2026clianything, sun2026agentslastexam} and driving scientific discovery workflows~\citep{wang2026naturebench, tbscience2026}.
\harness turns this competence toward robot control by treating the robot as a live software project, in which the coding agent can transfer its strong capabilities to physical tasks without any adaptation.

\parabf{Robot Operating System.}
ROS~2 provides a software framework that connects robot components and exposes their sensing, state, and control capabilities through common communication interfaces~\citep{macenski2022ros2}.
Topics carry streams such as camera images and joint states, services answer requests such as inverse kinematics queries, and actions support longer-running operations such as trajectory execution, with feedback and results.
The native \texttt{ros2cli} command-line tools and \texttt{rclpy} Python client library allow developers to inspect these interfaces, acquire data, and send control requests through terminal commands and programs, respectively.
Robot controllers then execute the motion requests, while sensor and state interfaces provide feedback about the resulting physical changes.
Prior work connects LLM agents to ROS through an additional abstraction layer of custom tools or predefined action libraries, with evaluations limited to in-house tasks rather than standard robot control benchmarks~\citep{royce2026rosa, mower2024ros_llm}.

\section{Harness and Workspace}
\label[appendix]{app_harness}

\subsection{Native ROS Interfaces}
\label[appendix]{app_ros_interfaces}
ROS~2 organizes robot software into nodes that exchange typed messages through topics, services, and actions~\citep{macenski2022ros2}.
These communication mechanisms connect sensor drivers, state estimation, planning, and control components.
The available names, message types, coordinate frames, and units depend on the robot and are described in the workspace documentation.

\paragraph{Communication and execution.}
Topics support publish--subscribe communication for sensor and state streams and for commands such as Cartesian velocities.
Services return a response to a request, for example an inverse-kinematics solution for a target pose.
Actions accept execution goals and provide feedback and a final result, with support for cancellation; joint-trajectory execution is one example.
Controllers handle motion requests through the corresponding interfaces, while sensor and state messages report the robot's evolving physical state.
The observation message types and file representations are detailed in~\cref{app_observation_acquisition}, and the trajectory, gripper, and velocity interfaces in~\cref{app_control_interfaces}.

\paragraph{Terminal and programmatic access.}
The \texttt{ros2cli} tools can list available interfaces, inspect their types, read or publish messages, call services, and send action goals.
For example, \texttt{ros2 topic list -t} lists topics and their message types, while \texttt{ros2 interface show <type>} displays the fields of an interface definition.
The \texttt{rclpy} library provides Python access to these same communication mechanisms through subscriptions, publishers, and service and action clients.
A program can keep these connections active and process incoming messages through callbacks, allowing sensor readings and motion updates to be combined in a running feedback loop.
Both entry points belong to the existing ROS~2 development tools; \harness does not require an additional agent-specific API to invoke them.

\subsection{Workspace}
\label[appendix]{app_workspace}

The initial workspace contains \texttt{README.md}, \texttt{machine.yaml}, documentation, and ordinary source files for basic operations such as saving camera observations and sending motion requests.
The machine description specifies the robot's sensors, control interfaces, joint limits, and coordinate frames.
The README directs the agent to these resources; the agent can inspect or modify starter tools and write additional programs through its existing terminal and file capabilities.
The workspace does not contain task solutions or findings from previous episodes.

The opening prompt is reproduced in~\cref{fig_workspace_prompt}; only its task placeholder changes between episodes.
It asks the agent to survey the machine, carry out the task, and verify the result without prescribing a sequence of perception or manipulation steps.
During an episode, the agent can retain observations, numerical measurements, programs, notes, and logs in the workspace.
These files persist within the episode, including across a resumed session, but are not transferred to the next episode.
The evaluator remains separate as described in~\cref{app_execution_protocol}.

\paragraph{Provided guidance.}
The documentation includes recommendations for acquiring and checking observations, computing coordinates, and using sensor feedback, together with generic working patterns.
The starter tools include pixel back-projection and an odometry-based mobile-base controller.
These resources provide initial guidance and executable examples; the agent chooses which to use, modify, or replace when developing its task-specific solution.

The complete initial documents and starter-tool source are reproduced in~\cref{app_workspace_materials}.

\subsection{Observation Acquisition and File Representations}
\label[appendix]{app_observation_acquisition}
The agent obtains observations through the interfaces described in \texttt{machine.yaml} and the workspace documentation.
The acquisition methods below are alternatives the agent can select or combine according to its needs, rather than a prescribed sequence.

\paragraph{Available data and interface discovery.}
RGB and depth topics use \texttt{sensor\_msgs/\allowbreak msg/\allowbreak Image} messages.
Camera intrinsics are provided in \texttt{sensor\_msgs/\allowbreak msg/\allowbreak CameraInfo} messages.
Joint-state messages use \texttt{sensor\_msgs/\allowbreak msg/\allowbreak JointState} and associate joint names with positions, velocities, and available efforts.
Some robots also expose wrist force and torque as \texttt{geometry\_msgs/\allowbreak msg/\allowbreak WrenchStamped} messages and base odometry as \texttt{nav\_msgs/\allowbreak msg/\allowbreak Odometry} messages.
The machine description identifies the available sensors and ports; these modalities are not assumed to be present on every robot.

The agent can inspect the running graph with \texttt{ros2 topic list}, examine a topic with \texttt{ros2 topic info}, and inspect its message definition with \texttt{ros2 interface show}.
Small messages, such as joint states or camera intrinsics, can be read through \texttt{ros2 topic echo <topic> --once}, with the output optionally redirected to a workspace file.

\paragraph{Image and numerical files.}
The starter tool \texttt{tools/perception/cam\_snap.py} subscribes to an image topic and saves one received frame.
It accepts either a camera name or a full topic path and an optional output filename.
For a color image, it converts the ROS message with \texttt{cv\_bridge} and writes a PNG.
For a depth image, it preserves the numerical array in a \texttt{.npy} file and writes a separate PNG visualization.
The visualization rescales depth values for display; metric calculations use the original array and its documented units.
The agent can inspect or modify this tool, or write an \texttt{rclpy} subscriber to save selected message fields in a format suited to subsequent processing.
For joint data, values must remain associated with their joint names rather than be interpreted using an assumed message order.

\paragraph{Calibration, coordinate frames, and timing.}
Geometric calculations require the camera intrinsics and the relation between the camera frame and the desired reference frame.
The agent can retrieve intrinsics from \texttt{CameraInfo} and query available frame transforms through \texttt{tf2}; the workspace documentation describes machine-specific alternatives when the required transform is unavailable.
Saving RGB and depth separately does not itself guarantee temporal alignment: the starter snapshot tool captures a single topic and does not synchronize multiple streams.
When a computation requires aligned observations, an agent-written acquisition program must account for message timestamps and the correspondence between image pixels, calibration, and coordinate frames.
This is particularly relevant when either the robot or the observed scene moves between captures.

\subsection{Native Control Requests}
\label[appendix]{app_control_interfaces}
The workspace documentation identifies the available action, service, and topic names, together with their message types and coordinate conventions.
For joint motion, a client sends a \texttt{control\_msgs/action/FollowJointTrajectory} goal containing joint names, target positions, and trajectory timing.
A Cartesian target can first be converted to joint positions through an available inverse-kinematics service.
Gripper requests use \texttt{control\_msgs/action/GripperCommand}, while a velocity client publishes commands such as \texttt{geometry\_msgs/msg/TwistStamped} in the documented reference frame.
The agent can invoke action and service clients through \texttt{ros2cli} or construct them in \texttt{rclpy}; the appropriate robot controller executes the requested motion.

\section{Evaluation Protocol and Compared Systems}
\label[appendix]{app_protocol}

\subsection{Benchmarks and Trial Sampling}
\label[appendix]{app_sampling}
The main evaluation contains 2,000 episodes: 700 on CaP-Bench, 800 on LIBERO-PRO, and 500 on RoboCasa365.
All three benchmarks are evaluated in simulation using their task definitions and success predicates.

\paragraph{CaP-Bench.}
We evaluate Lift, Stack, Wipe, Nut, Restack, Two-Arm Lift, and Handover, with 100 trials per task~\citep{fu2026capx}.
The last two tasks use two Panda arms; the others use one.
Trial identifiers range from 1 to 100, following the benchmark's evaluation grid.
The benchmark's reset implementation samples object placements randomly; these identifiers do not specify a fixed set of reproducible initial placements.
Our trial count therefore matches the published protocol, but the initial states are not paired episode by episode with those of prior systems.

\paragraph{LIBERO-PRO.}
We evaluate the Object, Spatial, Goal, and LIBERO-10 suites under two perturbations: initial-position swaps (P in~\cref{tab_main}) and task-instruction changes (T)~\citep{zhou2026libero_pro}.
Each of the eight suite--perturbation cells contains ten tasks, and we use initial-state indices 0--9 from the released initial-state files for each task.
This gives 100 trials per cell and 800 trials overall.
The six Object, Spatial, and Goal cells form the 600-episode subset shared with several prior coding-agent evaluations; the two LIBERO-10 cells add 200 episodes.
We retain the benchmark's perturbed scenes and task instructions and do not select initial states based on \harness performance.
The ten-trial count follows recent agent evaluations~\citep{zhang2026harnessvla, zhang2026rats, galanti2026pigey}; some earlier systems instead report 50 trials per task, as described in~\cref{app_compared}.

\paragraph{RoboCasa365.}
We use the target50 evaluation set in the benchmark's target kitchen split: 18 Atomic-Seen tasks, 16 Composite-Seen tasks, and 16 Composite-Unseen tasks~\citep{nasiriany2026robocasa365}.
Seen and Unseen refer to whether the task templates occur in the benchmark's pretraining demonstrations, rather than whether \harness has previously attempted them.
Every task is evaluated for ten trials, using seeds 0--9, for 180, 160, and 160 episodes in the three groups.
Each reset uses the benchmark's sampler to select an allowed kitchen layout and style, object instances, and placements; we do not force one trial per kitchen.
Harness VLA uses the same task groups but evaluates ten held-out seeds per atomic task and five per composite task after a separate reference-seed exploration~\citep{zhang2026harnessvla}.
Our ten-trial setting thus follows its atomic-task evaluation count and applies that count to all three groups, without reference-seed exploration.
The GR00T results reproduced from the RoboCasa365 paper use 30 trials per task, as specified in its Appendix G.2.

\subsection{Agent Configuration and Execution Protocol}
\label[appendix]{app_execution_protocol}
The main experiments use Claude Code with Claude Opus 5 and high reasoning effort.
Each episode uses the same workspace template for its robot configuration and a fresh agent session; files and task-specific findings are not carried between episodes.
The agent receives the task instruction and the opening prompt shown in~\cref{fig_workspace_prompt}, with machine documentation and starter tools as described in~\cref{app_workspace}.
No robot demonstrations or benchmark-specific rehearsal episodes are provided.
The alternative agent and model configurations used in the scaling analysis are described in~\cref{app_scaling}.

\paragraph{Simulation and control.}
Simulation advances when control commands trigger environment steps and remains paused between those steps, including while the agent reasons or reads observations.
A running control program can continue to issue commands and advance the simulation.
Thus, model inference latency does not itself advance physical time.
The robot exposes sensor, state, trajectory, gripper, and available velocity interfaces through ROS~2; the bridge executes these requests using the simulator's robot controllers.
The runs use joint-position control for arm trajectories and image observations rendered at $640\times480$ pixels.
These execution settings differ from evaluations that feed low-resolution images to a learned policy and use its operational-space action representation; the benchmark scenes and success predicates remain those of the original tasks.

\paragraph{Budgets and termination.}
An episode ends when the agent finishes, reaches 500 agent turns, or exhausts 240 minutes of active wall-clock time, whichever occurs first.
There is no additional simulation-step horizon in these runs.
Active wall-clock time includes reasoning and tool execution but excludes periods when an episode is suspended because model access is unavailable.
Resuming a suspended episode preserves its state and consumes the remaining budget rather than starting a new episode.
These budgets are not assumed to match those of every baseline; comparisons use the published settings of each system.

\paragraph{Evaluation isolation.}
The agent runs in a separate sandbox with access to the robot's ROS~2 interfaces and its own workspace.
The simulator implementation, privileged task-object poses, benchmark goal predicates, and evaluation records are not exposed in that workspace or as agent-callable tools.
The evaluator checks success independently, and its verdict is neither returned to the agent nor used to terminate an episode early.
The agent must decide when to finish using observations and execution feedback.

\subsection{Scoring and Aggregation}
\label[appendix]{app_scoring}
The evaluator checks the benchmark's original success predicate at every simulation step.
An episode counts as successful if the predicate is satisfied at any step before termination; once recorded, success is retained even if the final state later changes.
The agent's own declaration of completion is not used as the success label.
The first successful step and its timestamps are recorded for the budget analyses in~\cref{sec_scaling}.

Success rate is the fraction of evaluated episodes that satisfy this rule.
CaP-Bench averages over seven equally sized task groups.
For LIBERO-PRO, \cref{tab_main} reports both the six-cell average over Object, Spatial, and Goal and the eight-cell average that also includes LIBERO-10.
The latter corresponds to all 800 episodes; missing baseline cells are not imputed.
For RoboCasa365, the overall rate averages over all 50 tasks, weighting Atomic-Seen, Composite-Seen, and Composite-Unseen by their task counts of 18, 16, and 16.
Each main configuration is evaluated once over its stated episode set; we do not interpret small numerical differences as evidence of statistical significance.

\subsection{Compared Systems}
\label[appendix]{app_compared}

\subsubsection{Sources and Comparison Scope}
All baseline results in~\cref{tab_main,tab_robocasa} are reproduced from the cited papers or their accompanying leaderboards; we do not rerun these systems.
The comparisons therefore concern published system configurations, which can differ in model, reasoning budget, available tools, prior task experience, and evaluation protocol.
They are not controlled ablations that isolate the effect of the interface alone.
Models and reasoning-effort settings are listed alongside the results in the main tables.

\paragraph{Human-written programs.}
The CaP-Bench human reference row uses the S4 task-success values in Figure 17 of CaP-X~\citep{fu2026capx}: 93, 73, 100, 80, 47, 94, and 91 percent for Lift, Stack, Wipe, Nut, Restack, Two-Arm Lift, and Handover.
Their arithmetic mean is 82.6\%.
This row represents the benchmark's expert-written programs under its low-level primitive interface, rather than a new human evaluation conducted for this paper.

\paragraph{Learned visuomotor policies.}
The OpenVLA, $\pi_0$, and $\pi_{0.5}$ results on the six LIBERO-PRO cells are reproduced from the comparison in CaP-X Table 2~\citep{fu2026capx}.
We leave the two LIBERO-10 cells unreported for these rows instead of combining results from a different evaluation source.
The two GR00T N1.5 rows use the pretraining-only and pretraining-plus-full-target-data configurations in Table 2 of RoboCasa365~\citep{nasiriany2026robocasa365}.
The latter includes demonstrations of the target Composite-Unseen tasks; the split name describes their absence from pretraining, not their absence from this post-training stage.
We do not use the results from that paper's Table 1, which evaluates a different kitchen split.
The RLDX-1 row is the direct frozen-policy baseline reported in Harness VLA Table 4~\citep{zhang2026harnessvla}.

\paragraph{Agents that orchestrate learned policies.}
Pigey's six-cell LIBERO-PRO row comes from its Table 2, using a frozen $\pi_{0.5}$ backend and ten trials per task~\citep{galanti2026pigey}.
Its reported zero-shot protocol does not use task-specific memory or reference-seed exploration.
Harness VLA's LIBERO-PRO and RoboCasa365 results come from Tables 3 and 4, respectively~\citep{zhang2026harnessvla}.
It constructs task-specific memory through reference-seed exploration before evaluating on held-out seeds.
On LIBERO-PRO, it evaluates ten held-out seeds for each task in all eight cells; on RoboCasa365, it uses ten for atomic tasks and five for each composite group.
These counts and the reference-seed preparation differ from our fresh-session evaluation described in~\cref{app_sampling,app_execution_protocol}.

\paragraph{Agents that construct the policy.}
The CaP-X row denotes CaP-Agent0: its CaP-Bench values are the 3M configuration in Table 5, and its LIBERO-PRO values are from Table 2~\citep{fu2026capx}.
The RHO rows use its CaP-Bench S4 results in Table 1 and LIBERO-PRO results in Tables 6--8~\citep{elmaaroufi2026rho}.
RHO develops its repository on the first ten CaP-Bench trial identifiers per task and reports evaluation over all 100; for LIBERO-PRO, it develops on the unperturbed LIBERO tasks and evaluates on the perturbation suites.
ASPIRE's rows come from Tables 2 and 3~\citep{lu2026aspire}.
It develops a skill library before evaluation; on CaP-Bench it uses seeds 101--125 for development and evaluates generated programs on seeds 1--100.
The RATS row is the full execution system in its Table 1 after 50 play iterations, evaluated on 60 LIBERO-PRO task variants with ten initializations each~\citep{zhang2026rats}.
We use its reported 43.8\% full-system result rather than its separate plug-in configuration, and retain CaP-X's original baseline row rather than the RATS paper's rerun of that baseline.

\subsubsection{Training, Experience, and Oracle Access}
The table annotations distinguish three aspects of each reported configuration.
\textit{Training} indicates whether a robot policy is trained on demonstrations.
\textit{Experience} records additional preparation on the evaluation tasks or their underlying environments before the reported test episodes, such as rehearsed programs, accumulated skills, task-specific memory, or target-task adaptation.
\textit{Oracle} indicates whether the benchmark's success predicate is accessible during execution; access to rewards during an earlier development phase is a separate issue.
These annotations describe the reported configurations and are not mutually exclusive.
For \harness, task-specific programs and notes are produced only within the current episode and are not prior experience under this definition.

Trial counts, initial states, stopping rules, and models differ across sources.
For example, RHO reports 50 initial states per task on the six LIBERO-PRO cells, whereas our evaluation uses ten.
Pigey scores simulation trials using the LIBERO predicate and its own completion protocol, including failure when the agent times out without declaring completion~\citep{galanti2026pigey}; our scoring rule is specified in~\cref{app_scoring}.
Accordingly, we retain source-specific evaluation settings rather than claim that every baseline shares our execution budget or termination rule.

\section{Additional Comparative Analysis}
\label[appendix]{app_additional_results}

\subsection{Detailed Analysis of Main Results}
\label[appendix]{app_main_results}
The LIBERO-PRO percentages in this subsection refer to the six Object, Spatial, and Goal perturbation conditions.

\parabf{Robot-use agents are zero-shot visuomotor policies.}
\harness achieves 99.0\% on CaP-Bench and 91.8\% on the six LIBERO-PRO conditions considered here, without a learned robot policy, task-specific primitive library, or prior task experience.
For comparison, the expert-written programs achieve 82.6\% on CaP-Bench and provide a separate solution for each task.
The demonstration-trained policies OpenVLA, $\pi_0$, and $\pi_{0.5}$ remain at or below 12.8\% on the six LIBERO-PRO conditions after scene perturbations.

Among agents that construct the policy, CaP-X reaches 18.2\% on these LIBERO-PRO conditions using its provided primitives.
RHO, ASPIRE, and RATS reach 43.8--71.7\% after accumulating policy repositories or skills over the provided primitives before evaluation~\citep{elmaaroufi2026rho, lu2026aspire, zhang2026rats}.
These systems show how prior program development can improve robot use; \harness constructs its programs within the current episode through native ROS~2 interfaces.

Among agents that orchestrate learned policies, Pigey combines an LLM agent with a frozen $\pi_{0.5}$ and reaches 53.3\%.
Harness VLA combines a coding agent with a learned policy and reaches 87.7\%, with task-specific memory from reference-seed exploration and access to the success predicate during execution.
The corresponding \harness rate is 91.8\% without these resources.
Together, these comparisons support the main finding that the coding agent itself can serve as a zero-shot visuomotor policy.
The differences characterize the reported systems; their models and evaluation settings also differ, as described in~\cref{app_compared}.

\subsection{Native Interfaces and the Limits of Accumulated Skills}
\label[appendix]{app_native_interface_analysis}

\parabf{Experience improves performance but leaves difficult tasks unresolved.}
The low Nut success rates are not results from agents without prior skill development.
CaP-Agent0 uses nine synthesized utilities distilled from successful CaP-Bench rollouts~\citep[Section~4 and Appendix~H.1]{fu2026capx}.
RHO evolves multi-file policy repositories through repeated execution and reflective optimization~\citep[Section~3]{elmaaroufi2026rho}, while ASPIRE accumulates validated repairs in a skill library and searches over candidate programs~\citep[Section~2]{lu2026aspire}.
Despite this experience accumulation, their reported Nut success rates are 0\%, 1\%, and 9\%, respectively, compared with 100\% for \harness.
On the LIBERO-PRO Spatial suite, the corresponding position/task results are 12\%/14\% for CaP-X, 53.0\%/61.6\% for RHO, and 51\%/60\% for ASPIRE, versus 98\%/100\% for \harness.
Experience improves many tasks, but the reported optimization and skill accumulation do not close these gaps.

\parabf{The primitive interface constrains what experience can encode.}
ASPIRE explicitly identifies this limitation: programs remain bounded by the provided perception, planning, and control API; capabilities outside that API require inefficient approximation or a human extension of the interface~\citep[Section~5]{lu2026aspire}.
Accumulating strategies over an interface does not by itself expose an unavailable sensor or control operation.
CaP-X's documented S3/S4 simulation interface provides image observations, segmentation, grasp proposals, inverse kinematics, and joint and gripper moves, but does not list a force-reading or force-conditioned motion primitive~\citep[Appendix~G]{fu2026capx}.
This restriction concerns the evaluated simulation interface; CaP-X separately describes force-wrench access for its real-robot setup.
CaP-X also reports difficulties with insertion requiring continuous feedback, and attributes LIBERO-PRO failures to perception errors, angled grasp proposals, and the lack of collision-aware motion planning~\citep[Appendices~A and~J]{fu2026capx}.
Thus, the comparison concerns the capabilities available for programs to use as well as the experience accumulated in composing them.

\parabf{Native sensor interfaces support agent-written feedback programs.}
\harness exposes sensor and control interfaces through ROS~2, allowing the agent to write programs that monitor physical feedback during execution.
The Nut example in~\cref{fig_manipulation_case}(c) executes insertion in seven trajectory segments with a force reading after each; \cref{sec_behavior} reports the broader use of agent-written feedback programs and motion clients.
This provides a concrete mechanism for using sensing during manipulation without a pre-specified task primitive or a learned robot policy.
Harness VLA takes a different route, composing analytic primitives with a frozen VLA for contact-rich execution; its strongest Spatial result in~\cref{tab_main} is 94\%~\citep{zhang2026harnessvla}.
The consequences of relying on a learned policy under distribution shift are examined in~\cref{sec_generalization}.

\subsection{Generalization and Demonstration Coverage}
\label[appendix]{app_generalization}
This appendix expands the RoboCasa365 results in~\cref{sec_generalization}, examining demonstration coverage and composite-task difficulty.

\parabf{Target-task demonstrations improve the evaluated learned policy.}
The two GR00T N1.5 rows in~\cref{tab_robocasa} distinguish pretraining alone from pretraining followed by post-training on the target-task demonstrations~\citep[Table~2]{nasiriany2026robocasa365}.
On the same 16 Composite-Unseen tasks, success rises from 0.2\% to 42.1\% after post-training on their demonstrations.
This contrast shows the importance of demonstration coverage for the evaluated learned policy.
The term \textit{Unseen} refers to task templates absent from the benchmark's pretraining set; these templates are included in the target data used by the post-trained GR00T configuration.
Its 42.1\% therefore represents performance with target-task demonstration training, whereas the 28.1\% of \harness is obtained without such training.

\parabf{Harness VLA improves on RLDX-1 but retains a Seen--Unseen performance gap.}
RLDX-1 achieves 21.3\% on Composite-Seen tasks and 5.0\% on Composite-Unseen tasks in the frozen-policy evaluation reported by Harness VLA~\citep[Table~4]{zhang2026harnessvla}.
Harness VLA uses this same policy as its contact-rich execution primitive, together with analytic primitives, a coding agent, task-specific and global memory, and access to the benchmark's success predicate.
With Codex, it reaches 61.0\% on Seen tasks and 13.8\% on Unseen tasks; with Claude Code, the corresponding rates are 47.5\% and 15.0\%.
Both configurations improve on the underlying policy, including on Unseen tasks, but retain a pronounced Seen--Unseen performance gap.
The comparison therefore shows that adding agent-level planning, memory, and verification does not remove this gap in the evaluated configurations: contact-rich execution still relies on the frozen robot policy.
In \harness, the coding agent instead constructs the control program during the episode, using current observations and the robot's native sensing and control interfaces.
The same workflow yields 26.3\% on Seen and 28.1\% on Unseen composite tasks, without a learned robot policy or task-specific rehearsal.

\parabf{Composite tasks remain difficult despite stable Seen--Unseen performance.}
Stable performance across Seen and Unseen templates does not imply that composite tasks are easy.
The success rate of \harness falls from 61.1\% on Atomic-Seen tasks to 26.3\% and 28.1\% on the two composite groups; every compared system also performs worse on composite tasks than on atomic tasks.
This atomic-to-composite comparison concerns the additional demands of composing and executing multiple steps, whereas the Seen--Unseen comparison concerns task-template coverage within the composite category.
\cref{app_failure} examines the remaining difficulties, including incomplete long-horizon execution, physical constraints, mismatches between instructions and scoring criteria, and agent errors.
Thus, the generalization result is that the absence of a task template from robot pretraining data is not accompanied by lower success for \harness in this evaluation, even though composite manipulation remains challenging.

\section{Agent Perception and Control}
\label[appendix]{app_behavior}

This appendix expands the perception and control analyses in~\cref{sec_perception,sec_behavior} with statistics, measurement definitions, and concrete examples from all three benchmarks.
The examples include sensor images the agent saved directly or processed with its own code, control loops plotted from its printed logs, and its terminal and programs.
Text and code shown in the example panels are verbatim; for illustration, lines between the ones shown are omitted and marked \texttt{...}.

\subsection{Perception Statistics}
\label[appendix]{app_perception_statistics}

\paragraph{Episode-level prevalence of agent-built perception.}
The two columns in~\cref{tab_perception} count distinct behaviors, each at most once per episode.
For the image-processing column, an episode is counted when the agent reads an output image attributed to its own code for cropping, enlargement, annotation, masking, depth rendering, or combining sensor images.
A metric-program episode contains an invocation of agent-written code matching depth or array processing together with camera-intrinsic or coordinate-transform operations; invoking the supplied \texttt{px2world.py} alone does not qualify.
Across CaP-Bench and LIBERO-PRO, 1,090 of 1,500 episodes contain the former behavior and 1,452 contain the latter; both occur in 1,050 episodes (70.00\%).
These measures use source-pattern and execution-trace matching, which can miss indirect implementations and do not establish that every computed measurement is correct.

\paragraph{Perception activity and use of computed coordinates.}
\cref{tab_app_perception_activity} provides complementary statistics across all three benchmarks: image reads and metric-program invocations per episode, the fraction of image reads attributed to views computed from sensor data by agent-written programs, and the fraction of Cartesian motion targets traced to computed coordinates.
These measures describe activity and use within episodes, whereas~\cref{tab_perception} reports the prevalence of the two behaviors across episodes.

\begin{table}[t]
\centering
\footnotesize
\setlength{\tabcolsep}{5pt}
\renewcommand{\arraystretch}{1.15}
\caption{\textbf{Most Cartesian motion targets trace to programmatically computed coordinates.}
Statistics cover all mainline episodes with Claude Code and Opus~5 across the three benchmarks.
\textit{Images read} and \textit{Metric programs} are median counts per episode: image-file \texttt{Read} calls and executions of agent-written geometric measurement programs, respectively.
\textit{Processed images} is the share of image reads attributed to views processed from sensor data by agent-written code, such as crops, zooms, grids, and overlays.
\textit{Targets traced} is the share of Cartesian-target motion commands whose coordinates match an earlier metric-program output within 1.5\,mm or are computed inside the motion program itself.}
\label{tab_app_perception_activity}
\begin{tabular}{lcccc}
\toprule
Benchmark & Images read & \makecell{Processed\\images} & Metric programs & \makecell{Targets traced to\\computed coordinates} \\
\midrule
CaP-Bench & 14 & 16\% & 13 & 94\% \\
LIBERO-PRO & 21 & 36\% & 21 & 93\% \\
RoboCasa365 & 31 & 40\% & 35 & 90\% \\
\bottomrule
\end{tabular}
\end{table}

\parabf{The agent uses computed coordinates for motion and processed images for visual understanding.}
Of the motion commands with a Cartesian target, 90--94\% use coordinates traced to an earlier agent-written metric program or computed inside the motion program itself (\cref{tab_app_perception_activity}).
These programs use depth, camera intrinsics, or frame transforms; observation and parking poses account for the remaining targets.
Thus, the traces connect the measurement programs to the physical actions they support.

Interestingly, 16--40\% of image reads are views processed from sensor data by the agent's own programs.
Among reads of processed images, 68--94\% involve crops or enlargements and 3--11\% involve pixel grids or markers (\cref{tab_app_reads}).
These operations prepare views for recognition, inspection, and outcome checks.
For example, in~\cref{fig_perception_case}, the agent (a) recognizes the parts, (b) computes their positions from depth, (c) examines the handle on a gridded crop, and (d) checks the grasp using an eye-in-hand image and the finger gap.
It also cross-checks the coordinates from a second camera.
\Cref{fig_metric_code} shows the depth-processing program, and~\cref{app_programs} provides the corresponding code.

\paragraph{Image-read purposes and processing categories.}
\cref{tab_app_reads} splits the image reads of \cref{tab_app_perception_activity} by the event that precedes them: before the first motion command (understanding the scene), after a motion command with no metric computation in between (checking its outcome), or after a metric computation (inspecting an intermediate result such as a crop or a segmentation).
It also groups the processed images by the operations applied to their sensor inputs; a program that both crops and draws a grid counts under both, so the shares do not sum to one.

\begin{table}[h]
\centering
\footnotesize
\setlength{\tabcolsep}{4pt}
\caption{\textbf{Crops and enlarged views dominate the processed images the agent reads.}
For Claude Code with Opus~5 across all mainline episodes, read purposes are reported as shares of all image reads and are inferred from preceding events.
Processing categories are shares of reads of processed images and can overlap; a crop with a grid counts in both categories.}
\label{tab_app_reads}
\resizebox{\textwidth}{!}{%
\begin{tabular}{lccc|ccccc}
\toprule
& \multicolumn{3}{c|}{Purpose} & \multicolumn{5}{c}{Processed images} \\
Benchmark & Scene & Verify & Inspect & Crop / zoom & Grid / markers & Depth view & Mosaic & Mask overlay \\
\midrule
CaP-Bench & 32\% & 37\% & 31\% & 80\% & 3\% & 3\% & 1\% & 2\% \\
LIBERO-PRO & 38\% & 35\% & 27\% & 94\% & 11\% & 4\% & 2\% & 2\% \\
RoboCasa365 & 18\% & 47\% & 36\% & 68\% & 10\% & 5\% & 6\% & 2\% \\
\bottomrule
\end{tabular}}
\end{table}

\begin{figure}[h]
\centering
\includegraphics[width=\linewidth]{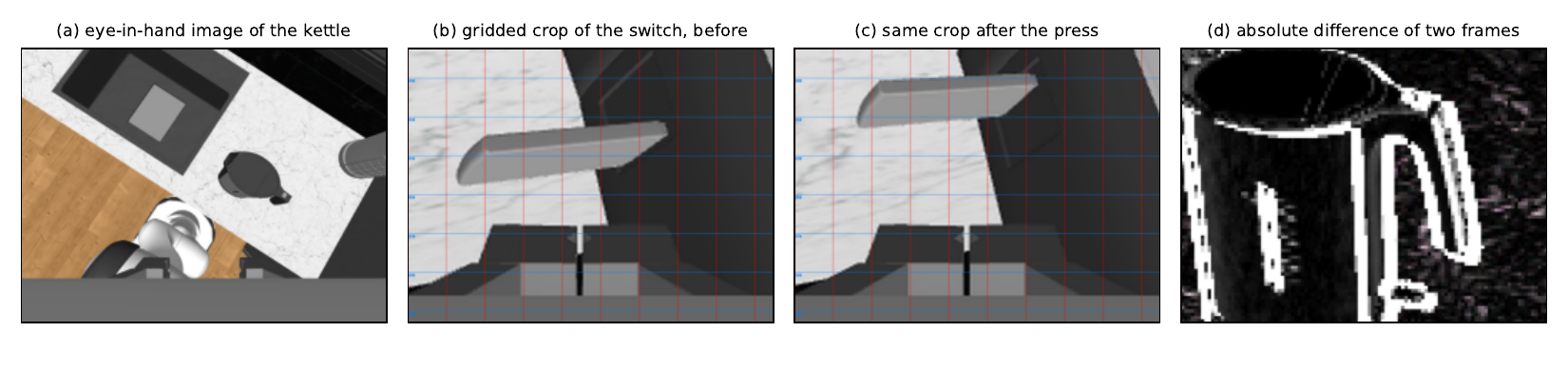}
\caption{\textbf{The agent writes image-processing code to check whether a switch moved.}
In a RoboCasa365 electric-kettle task, (a) it locates the kettle and switch in the eye-in-hand image, (b) crops the switch and adds a pixel grid, and (c) repeats the crop after pressing.
(d) Its difference image highlights changes in the lever and its shadow, providing evidence for checking whether the press took effect.}
\label{fig_perception_case_diff}
\end{figure}

\paragraph{Techniques in agent-written perception code.}
\cref{tab_app_techniques} lists the techniques found in agent-written perception code, as the share of episodes in which each appears.
\cref{fig_perception_case_diff,fig_perception_case_cloud} show two more pipelines the agent built for itself: frame differencing to check whether a switch moved, and a point cloud reconstructed from depth and rendered from a chosen viewpoint, with cross-sections to locate a rail.

\begin{table}[h]
\centering
\footnotesize
\setlength{\tabcolsep}{4pt}
\caption{\textbf{Agent-written perception code uses a range of geometric and image-processing techniques.}
Each percentage is the share of mainline episodes with Claude Code and Opus~5 in which the technique appears in agent-written code.
Techniques can co-occur; code presence does not establish successful execution.}
\label{tab_app_techniques}
\begin{tabular}{lccc}
\toprule
Technique & CaP-Bench & LIBERO-PRO & RoboCasa365 \\
\midrule
Depth threshold / height band & 97\% & 99\% & 98\% \\
Centroid / bounding box & 100\% & 96\% & 94\% \\
Point cloud / plane fit & 83\% & 69\% & 88\% \\
Connected components / contours & 68\% & 69\% & 56\% \\
Color threshold & 48\% & 24\% & 41\% \\
Crop / zoom for its own reading & 60\% & 99\% & 98\% \\
Frames stacked for comparison & 59\% & 34\% & 70\% \\
Cross-section / slice plots & 18\% & 35\% & 50\% \\
Frame differencing & 18\% & 14\% & 24\% \\
Multi-camera cross-check & 33\% & 16\% & 23\% \\
Point cloud rendered to an image & 6\% & 6\% & 15\% \\
\bottomrule
\end{tabular}
\end{table}

\subsection{Visual Inspection and Programmatic Image Processing}
\label[appendix]{app_visual_understanding}
Through its existing image-reading capability, the agent can inspect both camera images and views derived from sensor data by its own programs.
The operations below illustrate ways it can prepare observations for visual inspection; the harness does not prescribe a fixed image-processing sequence.

\paragraph{Preparing a view for inspection.}
The agent can use ordinary image-processing libraries to crop a region of interest, enlarge it, or overlay grids and markers, then save and read the resulting image.
In the Nut episode of \cref{fig_perception_case}, its program extracts rows 130--234 and columns 335--444 from the overhead image, enlarges the crop by a factor of six, and draws a grid at intervals of ten original pixels.
The grid labels retain the coordinates in the original image, allowing the agent to relate details in the enlarged view to the source pixels.
The program saves this view as \texttt{nut\_zoom.png}; \cref{app_programs} provides the code.
These image operations expose existing visual detail and add reference marks; physical dimensions still require a geometric relation between image coordinates and the scene.

\paragraph{Checking action outcomes.}
Visual inspection also supports checking changes after an action.
In \cref{fig_perception_case}, the agent inspects an eye-in-hand image after grasping and compares the finger gap with its earlier measurement of the handle width.
In the kettle episode of \cref{fig_perception_case_diff}, it crops the switch before and after pressing and produces a difference image to inspect the change.
Such views can be considered alongside numerical readings and terminal output when deciding whether to continue or revise an action.

\paragraph{Inspecting computed representations.}
Programs can also turn numerical results into images for the agent to inspect.
For example, \cref{fig_perception_case_cloud} shows an agent reconstructing a dishwasher rack from depth, rendering its point cloud from a chosen viewpoint, and plotting cross-sections to examine the rail.
Here, numerical processing constructs the representation and visual inspection helps the agent interpret it.
The prevalence of these operations across episodes is reported in~\cref{app_perception_statistics}.

\subsection{Metric Measurements from Sensor Data}
\label[appendix]{app_metric_measurement}
This subsection expands the numerical processing described in~\cref{sec_perception}.
The agent can combine measurements read from native interfaces with quantities computed from saved observations; the harness does not prescribe a task-specific measurement pipeline.

\paragraph{Direct readings and derived quantities.}
Joint positions, velocities, efforts, and available force/torque readings provide numerical information about the robot and its interaction with the scene.
Available frame transforms can be queried through \texttt{tf2}, for example with \texttt{ros2 run tf2\_ros tf2\_echo <reference> <source>}.
These transforms describe the frames exposed by the robot's software; locating a task object without a corresponding tracked frame requires additional estimation from observations.
The acquisition interfaces and file representations are described in~\cref{app_observation_acquisition}.

\paragraph{From depth pixels to physical coordinates.}
For a pixel $(u,v)$ in a pinhole camera image, let $z$ denote its depth along the camera's optical axis, in meters, and let $K$ be the intrinsic matrix for that image.
The corresponding camera-frame point and its reference-frame coordinates are
\begin{equation}
\mathbf{p}_{c}=zK^{-1}\begin{bmatrix}u & v & 1\end{bmatrix}^{\top},
\qquad
\mathbf{p}_{r}=R_{rc}\mathbf{p}_{c}+\mathbf{t}_{rc},
\label{eq_metric_backprojection}
\end{equation}
where $R_{rc}$ and $\mathbf{t}_{rc}$ transform coordinates from the camera optical frame to the chosen reference frame, such as the world or robot base frame.
Here, $u$ indexes columns and $v$ indexes rows; the formula assumes pixels consistent with the pinhole calibration, with distortion accounted for where necessary.
Depth values must use the documented scale and convention, and non-finite or non-positive values cannot be used as valid surface measurements.
If target pixels are selected in a color image, their correspondence with depth pixels must be established before applying the calculation.
Calibration, image timing, and frame transforms must refer to the observations being processed, as discussed in~\cref{app_observation_acquisition}.
The agent can apply this calculation to selected pixels or a segmented region to estimate positions and extents.

\paragraph{The Nut example.}
In the episode shown in~\cref{fig_perception_case}(b), the agent loads \texttt{bird\_depth.npy} and uses the overhead camera height of $2.0$\,m to compute world heights as $2.0-z$.
Its program selects points with world heights between $0.825$ and $1.3$\,m, forms connected components, and excludes components containing fewer than 50 pixels.
It then converts the remaining pixels into world coordinates and reports each component's coordinate ranges and centroid.
These thresholds and the simplified overhead-camera calculation belong to this episode and camera configuration, rather than to a fixed harness procedure.
The code, including a second-camera calculation using an explicit rotation and translation, is reproduced in~\cref{app_programs}.

\paragraph{Cross-checking and using measurements.}
Before moving, the agent repeats the geometric measurement from an oblique camera and compares the results in the same world frame.
After grasping, it compares the measured finger gap of $30.35$\,mm with the previously estimated handle width of $30.4$\,mm, alongside the eye-in-hand image in~\cref{fig_perception_case}(d).
This comparison supplies evidence about the grasp rather than a universal success criterion.
Likewise, force and effort readings can provide contact evidence, while accounting for their baseline and reference frame.
How programs incorporate such readings during motion belongs to the control mechanism described in~\cref{sec_behavior}.

\subsection{Additional Perception Examples}

\begin{figure}[h]
\centering
\begin{subfigure}[t]{0.24\linewidth}\includegraphics[width=\linewidth]{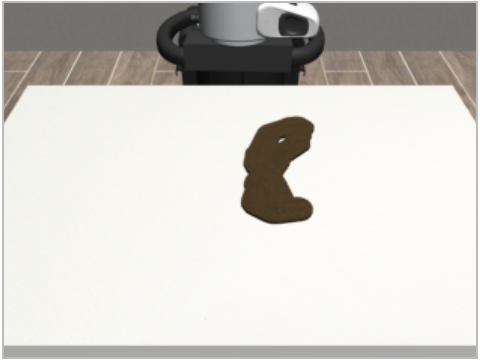}\caption{Spill, agent view.}\end{subfigure}\hfill
\begin{subfigure}[t]{0.24\linewidth}\includegraphics[width=\linewidth]{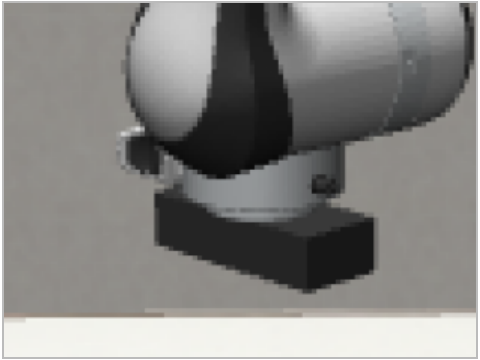}\caption{Crop of the pad at the probe.}\end{subfigure}\hfill
\begin{subfigure}[t]{0.24\linewidth}\includegraphics[width=\linewidth]{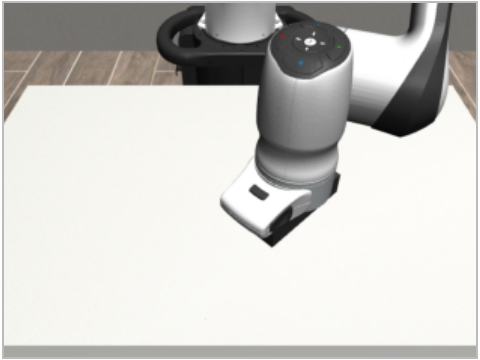}\caption{Mid-stroke.}\end{subfigure}\hfill
\begin{subfigure}[t]{0.24\linewidth}\includegraphics[width=\linewidth]{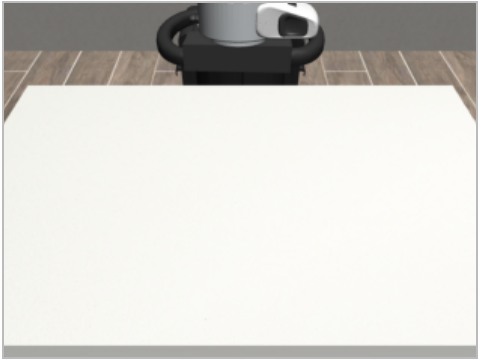}\caption{After, same pose as (a).}\end{subfigure}\\[2pt]
\begin{subfigure}[t]{0.24\linewidth}\includegraphics[width=\linewidth]{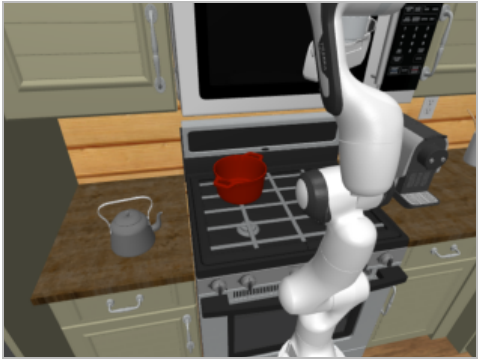}\caption{Kitchen, left camera.}\end{subfigure}\hfill
\begin{subfigure}[t]{0.24\linewidth}\includegraphics[width=\linewidth]{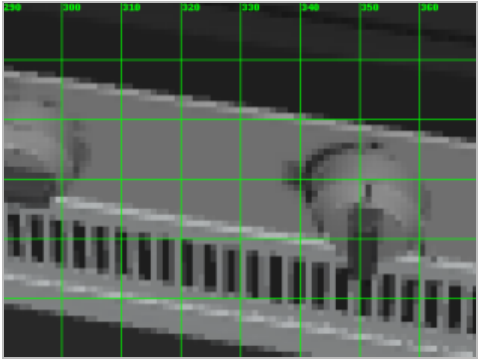}\caption{Gridded crop of the knobs.}\end{subfigure}\hfill
\begin{subfigure}[t]{0.24\linewidth}\includegraphics[width=\linewidth]{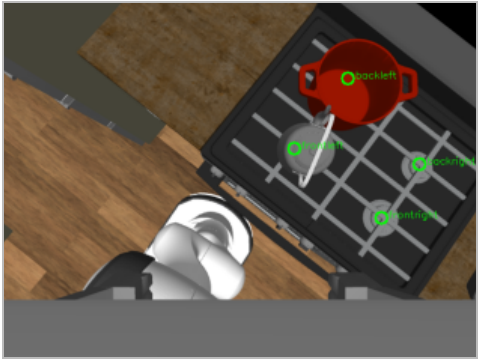}\caption{Burner centres projected back.}\end{subfigure}\hfill
\begin{subfigure}[t]{0.24\linewidth}\includegraphics[width=\linewidth]{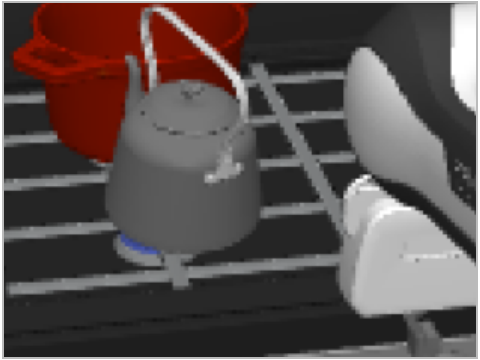}\caption{Flame found under the kettle.}\end{subfigure}
\caption{\textbf{The agent combines image processing and geometric measurements to locate targets and check action outcomes.}
Panels (a--d) show a CaP-Bench Wipe episode before, during, and after a stroke; (e--h) show a RoboCasa365 kettle episode, including a gridded knob crop, projected burner centers, and detected flame.
Panels (b), (f), (g), and (h) contain sensor observations processed by agent-written code.}
\label{fig_app_perc}
\end{figure}

\parabf{The agent combines image processing with geometry to locate targets and check outcomes.}
\Cref{fig_app_perc} provides two further examples of perception programs built during execution.
In CaP-Bench Wipe, the agent masks the spill by color in three cameras, projects each mask onto the table plane, and keeps connected regions with at least 90\% of their projected points on the table.
It rasterizes them into 5\,mm world grids and intersects the three grids to select the wipe region.
After the stroke, it repeats the measurement and compares the same camera view at the same arm pose (a--d).

In the RoboCasa365 kettle episode (e--h), the agent crops the stove knobs and adds a pixel grid to select knob pixels.
It then projects depth-derived burner centers back into the wrist image to check their locations and combines a color threshold with world-coordinate bounds to detect the flame.
Both episodes connect processed views and numerical measurements to a concrete action or outcome check.

\subsection{Control Statistics}

\paragraph{Episode-level prevalence of agent-written control tools.}
\cref{tab_manipulation} covers all 700 CaP-Bench and 800 LIBERO-PRO episodes, including successes and failures.
The motion-client column counts episodes whose agent-written workspace files contain native trajectory, gripper, velocity, or inverse-kinematics client code, excluding supplied tools and documentation.
These files contain \texttt{rclpy} and client or publisher construction; the count establishes the presence of client source, not its successful execution.
The feedback-program column counts episodes that invoke agent-written motion programs in which sensor-derived values affect a motion argument, loop condition, or branch; reading a sensor alone does not qualify.
The respective counts are 679 and 421 on CaP-Bench, and 759 and 331 on LIBERO-PRO, giving pooled rates of 1,438/1,500 (95.87\%) and 752/1,500 (50.13\%).
The classifications use source and execution-trace patterns; indirect implementations can be missed, and feedback programs need not run high-frequency velocity servos.

\paragraph{Manipulation activity and control modes.}
\cref{tab_app_manipulation_activity} retains the finer task-group breakdown and RoboCasa365 results, distinguishing motion-command frequency, the share of commands issued through agent-written programs, and episode-level use of feedback, force readings, and velocity control.

\begin{table}[t]
\centering
\footnotesize
\setlength{\tabcolsep}{4pt}
\renewcommand{\arraystretch}{1.15}
\caption{\textbf{The agent uses both individual motion commands and programs with sensor-dependent control logic.}
Statistics cover all mainline episodes with Claude Code and Opus~5, grouped by task set.
\textit{Motion commands} is the median number of arm, gripper, or base commands per episode; \textit{Through own programs} is the share issued by agent-written programs rather than starter tools.
The remaining columns are shares of episodes: \textit{Feedback program} uses a sensor reading to determine a motion, loop condition, or branch; \textit{Reads force} reads the wrench topic; \textit{Velocity servo} uses a feedback loop that streams Cartesian velocities.}
\label{tab_app_manipulation_activity}
\resizebox{\textwidth}{!}{%
\begin{tabular}{lccccc}
\toprule
Tasks & \makecell{Motion commands\\per episode} & \makecell{Through own\\programs} & \makecell{Feedback\\program} & \makecell{Reads\\force} & \makecell{Velocity\\servo} \\
\midrule
CaP-Bench, contact-rich (Nut, Wipe, Restack, 2A-Lift, Handover) & 14 & 99\% & 67\% & 86\% & 4\% \\
CaP-Bench, Lift and Stack & 9 & 86\% & 44\% & 97\% & 6\% \\
LIBERO-PRO, six perturbation cells & 20 & 68\% & 37\% & 64\% & 3\% \\
LIBERO-10 & 26 & 84\% & 54\% & 44\% & 3\% \\
RoboCasa365 & 50 & 86\% & 71\% & 63\% & 12\% \\
\bottomrule
\end{tabular}}
\end{table}

\parabf{The agent combines step-by-step execution with program-level feedback.}
On the contact-rich CaP-Bench tasks, 99\% of motion commands run agent-written programs, 67\% of episodes contain programs with sensor-dependent control logic, and 86\% read the force sensor (\cref{tab_app_manipulation_activity}).
On the six LIBERO-PRO perturbation cells, 68\% of commands run through agent-written programs, with the remainder using starter tools; feedback programs appear in 37\% of episodes.
These differences show that the agent uses both individual commands and programs that make adjustments from sensor readings.

Crucially, a feedback program need not be a high-frequency velocity servo.
Velocity servos appear in only 3--12\% of episodes across the task groups; feedback can instead operate between short trajectory segments.
For example, \cref{fig_manipulation_case} shows (a) a proportional approach to a drawer handle, (b) pulling guarded by finger-gap and displacement checks, (c) insertion with force readings between segments, and (d) force checks around a wiping trajectory.
The prevalence of agent-written motion clients is reported in~\cref{tab_manipulation}; the client library and programs from~\cref{fig_drawer_strip} are provided in~\cref{app_programs}.

\parabf{Starter tools support the agent without fixing its choice of control programs.}
The workspace supplies six scripts for basic operations (\cref{sec:method}), and~\cref{tab_app_tools} reports their use across the three benchmarks.
The camera snapshot tool is executed in every episode; use of the motion scripts varies by operation and benchmark.
For example, on CaP-Bench the inverse-kinematics tool is executed in 3\% of episodes and read without execution in 41\%, whereas the trajectory tool is executed in 63\% of LIBERO-PRO episodes and read without execution in 13\%.
These patterns show that starter source can serve as a reference as well as an executable tool.
Agent-written sensor subscriptions appear in 96--99\% of episodes and motion clients in 95--97\%; episodes contain a median of 6--14 program files, most organized as importable modules.

\begin{table}[h]
\centering
\footnotesize
\setlength{\tabcolsep}{4pt}
\caption{\textbf{Starter-tool use varies by operation and benchmark.}
For Claude Code with Opus~5 across all mainline episodes, each cell reports the percentage of episodes executing the tool, with mean calls per episode in parentheses.
Where shown, \textit{read} reports the percentage that only read the tool's source.}
\label{tab_app_tools}
\resizebox{\textwidth}{!}{%
\begin{tabular}{llccc}
\toprule
Tool & Purpose & CaP-Bench & LIBERO-PRO & RoboCasa365 \\
\midrule
cam\_snap & save an image topic to a file & 100\% (7.9) & 100\% (12.5) & 100\% (17.8) \\
px2world & back-project a pixel with depth & 29\% (0.6), read 24\% & 75\% (1.9), read 10\% & 57\% (1.6), read 24\% \\
fjt\_send & send one joint trajectory & 8\% (0.1), read 33\% & 63\% (2.8), read 13\% & 32\% (2.0), read 30\% \\
gripper\_cmd & open / close the gripper & 9\% (0.2), read 25\% & 56\% (1.8), read 17\% & 22\% (1.5), read 29\% \\
ik\_move & IK to a pose and move & 3\% (0.1), read 41\% & 41\% (1.2), read 36\% & 8\% (0.6), read 50\% \\
base\_goto & drive the base to a pose & -- & -- & 62\% (3.0), read 11\% \\
\bottomrule
\end{tabular}}
\end{table}

\subsection{Command Execution and Feedback Programs}
\label[appendix]{app_manipulation_execution}
This subsection expands the two forms of interaction described in~\cref{sec_behavior}.
Both can use the same native robot interfaces.
The distinction is whether the agent inspects an outcome and chooses another action, or a running program makes the next adjustment from sensor feedback.
A single task can combine both, and a program used for one step can itself contain a feedback loop.

Native control interfaces are described in~\cref{app_control_interfaces}; the following examples show how the agent uses them during execution.

In the drawer episode, the agent writes a reusable Python client for the IK service, trajectory and gripper actions, joint-state subscription, and velocity topic, then imports it into its action programs (\cref{app_programs}).

\begin{figure}[t]
\centering
\begin{subfigure}[t]{0.19\linewidth}\includegraphics[width=\linewidth]{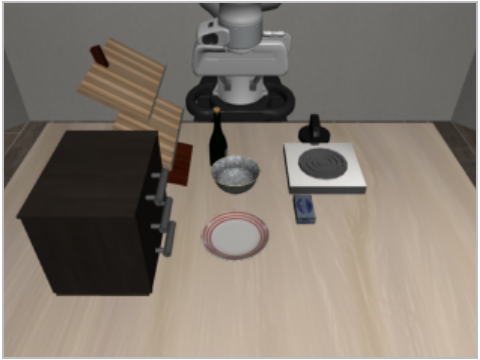}\caption{Survey the scene.}\end{subfigure}\hfill
\begin{subfigure}[t]{0.19\linewidth}\includegraphics[width=\linewidth]{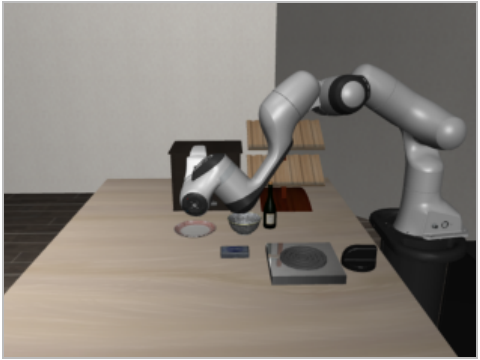}\caption{Before grasping the handle.}\end{subfigure}\hfill
\begin{subfigure}[t]{0.19\linewidth}\includegraphics[width=\linewidth]{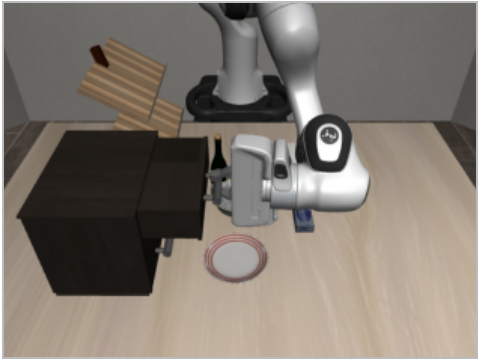}\caption{After pulling.}\end{subfigure}\hfill
\begin{subfigure}[t]{0.19\linewidth}\includegraphics[width=\linewidth]{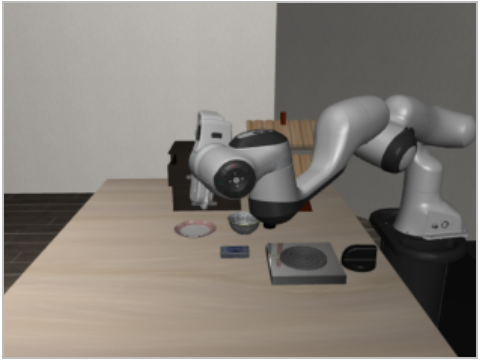}\caption{Drawer open, reaching the box.}\end{subfigure}\hfill
\begin{subfigure}[t]{0.19\linewidth}\includegraphics[width=\linewidth]{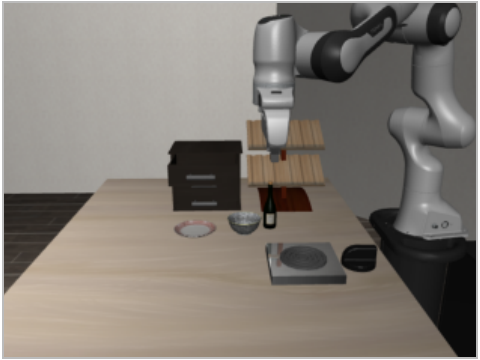}\caption{Box released in the drawer.}\end{subfigure}
\caption{\textbf{The agent uses visual checkpoints to assess progress during a drawer task.}
Camera observations from the LIBERO-PRO episode in~\cref{fig_drawer_strip} show (a) the initial scene, (b) the handle before grasping, (c) the drawer after pulling, (d) the reach toward the cream-cheese box, and (e) its release in the drawer.}
\label{fig_drawer_moments}
\end{figure}

\paragraph{Checking a step and revising the next one.}
Terminal output can report request acceptance, execution results, and measurements printed by a program.
The agent uses these together with physical observations when assessing progress.
In the drawer episode, the first grasp closes on air: the program reports a finger gap of approximately $1.9$\,mm.
The agent revises the approach target and runs a new program before attempting the pull; \cref{fig_drawer_strip,fig_drawer_moments} show its code and the corresponding visual checkpoints.
The numerical evidence and stopping conditions are task-specific, rather than fixed success rules imposed by the harness.

\begin{figure}[t]
\centering
\begin{subfigure}[t]{0.24\linewidth}\includegraphics[width=\linewidth]{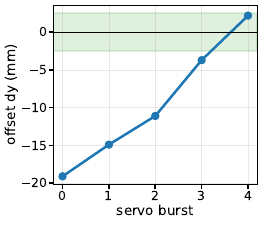}\caption{Approach the drawer handle.}\end{subfigure}\hfill
\begin{subfigure}[t]{0.24\linewidth}\includegraphics[width=\linewidth]{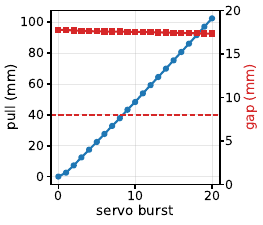}\caption{Pull the drawer.}\end{subfigure}\hfill
\begin{subfigure}[t]{0.24\linewidth}\includegraphics[width=\linewidth]{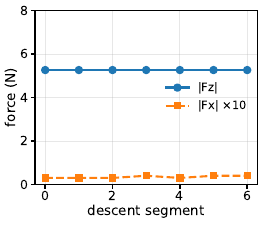}\caption{Insert the nut.}\end{subfigure}\hfill
\begin{subfigure}[t]{0.24\linewidth}\includegraphics[width=\linewidth]{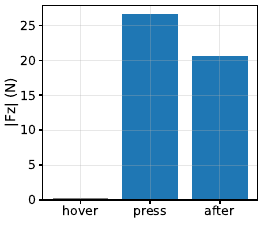}\caption{Wipe.}\end{subfigure}
\caption{\textbf{Agent-written programs use position, gripper, and force readings to monitor and adjust motion.}
The traces are plotted from the programs' printed logs.
In the LIBERO-PRO drawer episode, (a) proportional velocity corrections bring the hand within 2.5\,mm of the target (green band); (b) the pulling program checks displacement and finger gap after each burst, stopping if the gap falls below 8\,mm (dashed line) or motion stalls.
In CaP-Bench, (c) the Nut program reads the wrench after each of seven insertion segments, with the vertical load remaining at the tool's weight and no lateral force observed; (d) the Wipe program checks contact before executing 153 position-controlled waypoints and reads the load again afterward.}
\label{fig_manipulation_case}
\end{figure}

\paragraph{Position feedback and guarded pulling.}
The revised approach program reads the current tool position, computes its offset from the grasp target, and applies proportional velocity corrections along the relevant axes.
In this episode, it multiplies the errors by five, clips the two velocity components to $0.05$ and $0.02$\,m/s, and exits the approach loop when the absolute error along the approach axis falls below $2.5$\,mm or the iteration limit is reached.
Its velocity helper publishes bursts of 20 messages with a nominal rate parameter of 20\,Hz; the state checks occur between bursts, so this parameter does not imply that the complete feedback loop runs at 20\,Hz.
The pulling program checks the finger gap and displacement, stopping if the gap falls below $8$\,mm, the pull target is reached, displacement after a burst is below $1$\,mm, or the iteration limit is exhausted.
It prints the observed states and stop reason for the agent to inspect.
\cref{fig_manipulation_case} plots the recorded behavior, and \cref{app_programs} gives the corresponding code.

\paragraph{Force feedback between trajectory segments.}
Feedback can also determine whether to continue a sequence of short position-controlled motions.
In the CaP-Bench Restack example, the agent-written descent program records an initial force baseline, lowers the end effector in $2$\,mm increments, and reads the force after each segment.
It stops when the norm of the force change exceeds $1.5$\,N or the specified lower position limit is reached, returning the final height and a contact flag.
The threshold and step size belong to this program rather than to a harness-wide control policy.
\cref{fig_app_loops} illustrates the descent, and \cref{app_programs} reproduces the program.
Such feedback is processed directly by the running program; it need not be saved to an observation file or passed through the agent's context at each update.
The agent can inspect the resulting logs and observations before selecting another action or revising the program.

\subsection{Additional Control Examples}
\label[appendix]{app_episodes}

\paragraph{Recovery from a failed grasp.}
The ketchup-bottle recovery sequence is shown in~\cref{fig_recovery_case} and discussed in~\cref{sec_behavior}.

\begin{figure}[h]
\centering
\begin{subfigure}[t]{0.24\linewidth}\includegraphics[width=\linewidth]{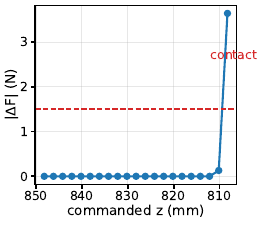}\caption{Guarded descent (Restack).}\end{subfigure}\hfill
\begin{subfigure}[t]{0.24\linewidth}\includegraphics[width=\linewidth]{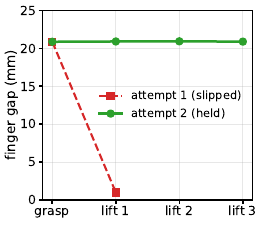}\caption{Staged lift (two-arm lift).}\end{subfigure}\hfill
\begin{subfigure}[t]{0.24\linewidth}\includegraphics[width=\linewidth]{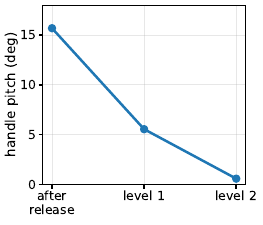}\caption{Levelling (handover).}\end{subfigure}\hfill
\begin{subfigure}[t]{0.24\linewidth}\includegraphics[width=\linewidth]{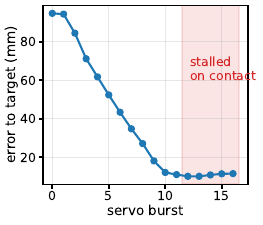}\caption{Servo to contact (RoboCasa365).}\end{subfigure}
\caption{\textbf{The agent uses measured force, finger gap, orientation, and position error to guide motion.}
Traces plotted from program logs show (a) a force-guarded descent in Restack, (b) a staged two-arm lift, (c) handle leveling during handover, and (d) a velocity approach to a jar in RoboCasa365.}
\label{fig_app_loops}
\end{figure}

\parabf{Sensor feedback supports stopping, recovery, and corrective motion.}
\Cref{fig_app_loops} plots four examples from the programs' printed logs.
(a) In CaP-Bench Restack, a descent proceeds in 2\,mm steps and stops when the wrist-wrench change exceeds 1.5\,N.
(b) In Two-Arm Lift, the first attempt's finger gap collapses from the bar thickness to 1\,mm; the agent aborts, measures the displaced pot again, drags it back with one arm, and grasps 7\,mm deeper.
The second attempt maintains its grip through all three lift stages.
(c) In Handover, the receiving arm's handle droops by 15\textdegree; the agent measures the pitch from the handle's top face and levels it in two iterations.
(d) In RoboCasa365, a velocity approach to a jar converges until contact leaves an 11\,mm position error, which the program reports as contact.
These cases illustrate both checks inside running programs and corrections selected by the agent between executions.

\begin{figure}[h]
\centering
\begin{subfigure}[t]{0.24\linewidth}\includegraphics[width=\linewidth]{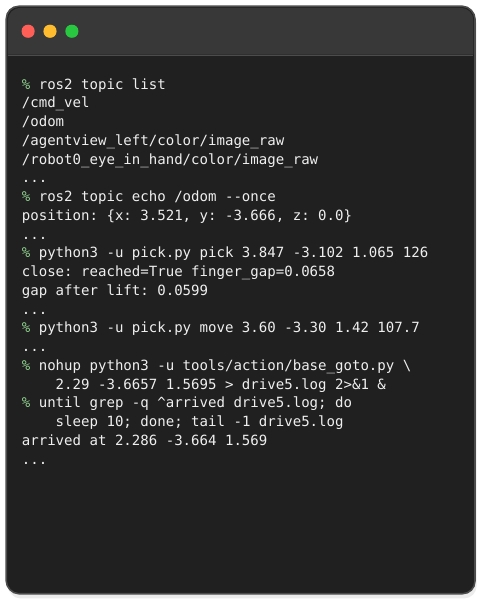}\caption{Drive, then reach.}\end{subfigure}\hfill
\begin{subfigure}[t]{0.24\linewidth}\includegraphics[width=\linewidth]{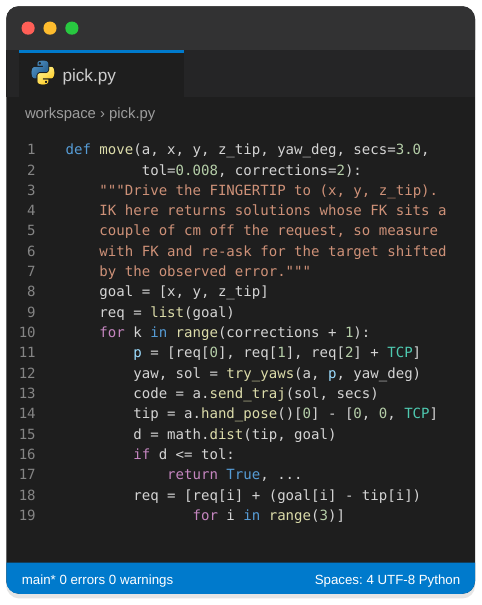}\caption{Move with an FK-measured correction.}\end{subfigure}\hfill
\begin{subfigure}[t]{0.24\linewidth}\includegraphics[width=\linewidth]{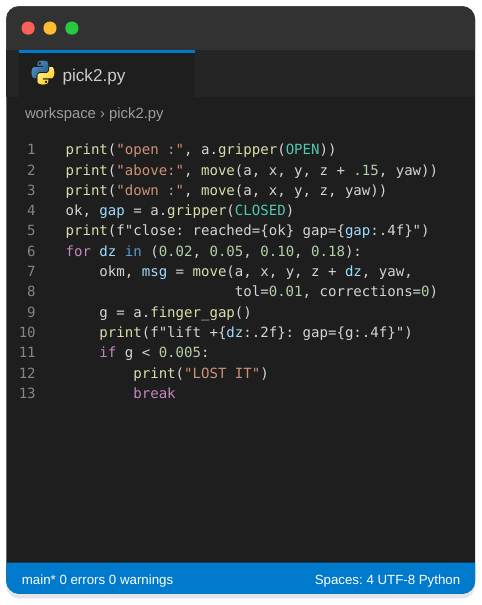}\caption{Lift in stages, watching the gap.}\end{subfigure}\hfill
\begin{subfigure}[t]{0.24\linewidth}\includegraphics[width=\linewidth]{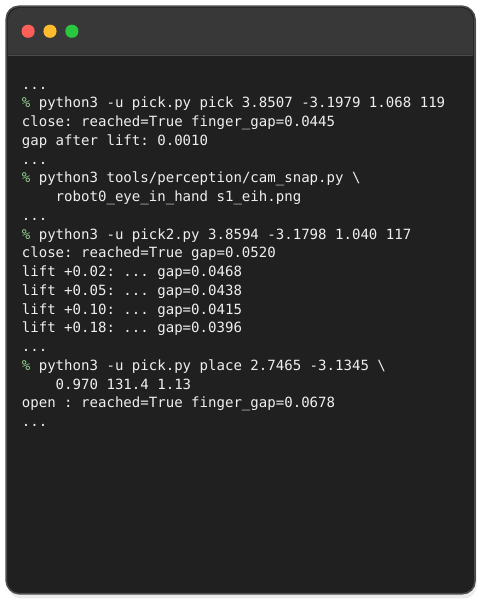}\caption{A failed grasp and the retry.}\end{subfigure}
\caption{\textbf{The agent writes programs that account for base motion and check grasps during transport.}
Code and terminal output from a RoboCasa365 hot-dog task show (a) driving before reaching, (b) correcting a move using forward kinematics, (c) checking finger gap during a staged lift, and (d) retrying a failed grasp.}
\label{fig_app_hotdog_strip}
\end{figure}

\begin{figure}[h]
\centering
\begin{subfigure}[t]{0.19\linewidth}\includegraphics[width=\linewidth]{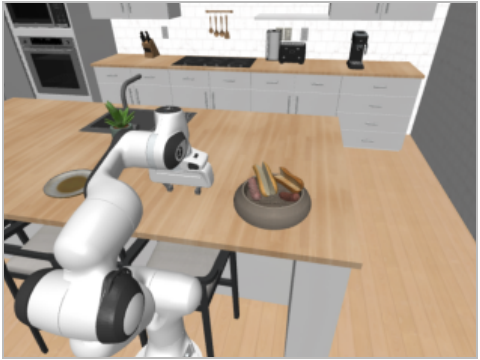}\caption{Bowl with the items.}\end{subfigure}\hfill
\begin{subfigure}[t]{0.19\linewidth}\includegraphics[width=\linewidth]{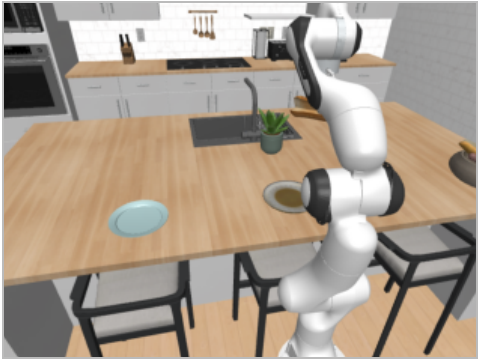}\caption{After the first drive: a second plate.}\end{subfigure}\hfill
\begin{subfigure}[t]{0.19\linewidth}\includegraphics[width=\linewidth]{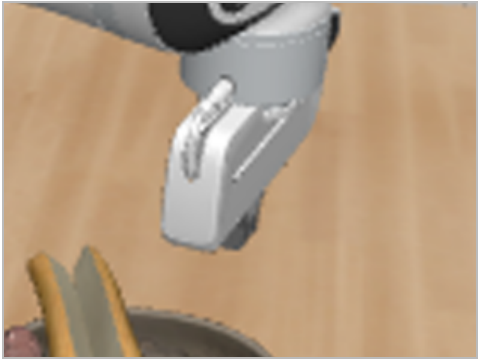}\caption{Empty gripper after a failed grasp.}\end{subfigure}\hfill
\begin{subfigure}[t]{0.19\linewidth}\includegraphics[width=\linewidth]{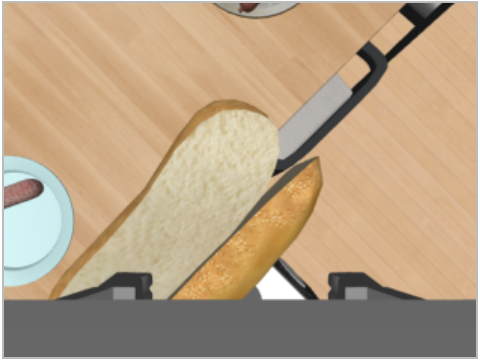}\caption{Bun wedged in the open fingers.}\end{subfigure}\hfill
\begin{subfigure}[t]{0.19\linewidth}\includegraphics[width=\linewidth]{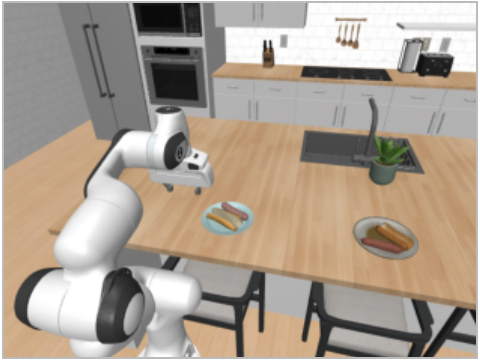}\caption{Both plates set.}\end{subfigure}
\caption{\textbf{Visual checkpoints reveal grasp failures during mobile manipulation.}
Agent-saved images from the RoboCasa365 episode in~\cref{fig_app_hotdog_strip} show the task's progress; (c) and (d) are crops used to diagnose grasp problems before the agent completes both plates.}
\label{fig_app_hotdog_moments}
\end{figure}

\parabf{The agent adapts its motion programs to base movement and grasp failures.}
In the RoboCasa365 hot-dog task, two buns and two sausages must be transferred from a bowl to two plates about 1.5\,m apart, requiring a base drive for each transfer (\cref{fig_app_hotdog_strip,fig_app_hotdog_moments}).
The agent uses the starter script to drive the base and writes an arm helper that reads the current base pose inside each inverse-kinematics call.
It measures the resulting motion with forward kinematics and corrects the next target by the observed error.
After a sausage slips, it revises the lift program to check the finger gap at four heights.
The second plate only becomes visible after the first drive, illustrating why the program must work with observations acquired during execution.


\begin{figure}[h]
\centering
\begin{subfigure}[t]{0.24\linewidth}\includegraphics[width=\linewidth]{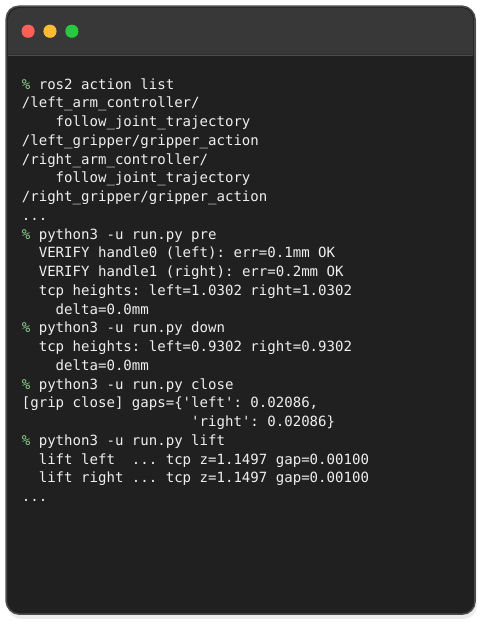}\caption{One command per phase, verified.}\end{subfigure}\hfill
\begin{subfigure}[t]{0.24\linewidth}\includegraphics[width=\linewidth]{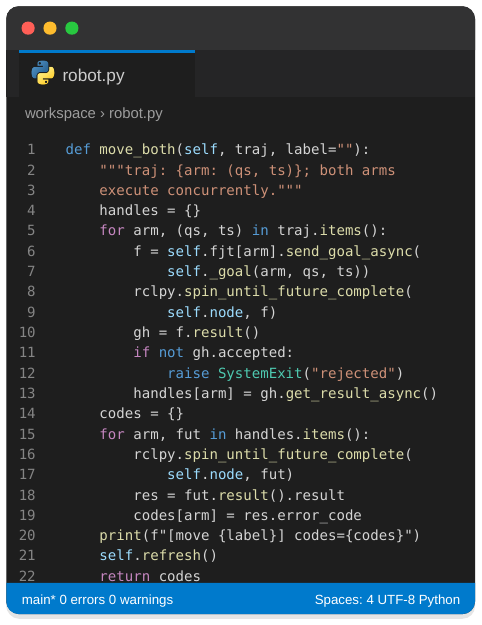}\caption{Send both goals, then wait for both.}\end{subfigure}\hfill
\begin{subfigure}[t]{0.24\linewidth}\includegraphics[width=\linewidth]{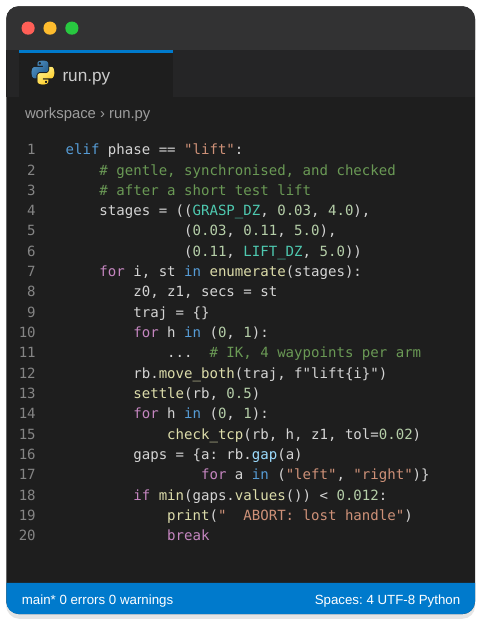}\caption{Staged lift with an abort guard.}\end{subfigure}\hfill
\begin{subfigure}[t]{0.24\linewidth}\includegraphics[width=\linewidth]{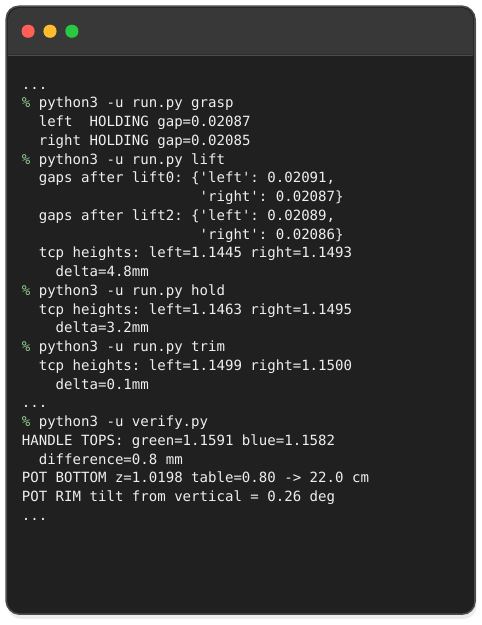}\caption{The second attempt, and the check.}\end{subfigure}
\caption{\textbf{The agent writes code to coordinate both arms and verify each lift stage.}
Code and terminal output from CaP-Bench Two-Arm Lift show (a) phase-by-phase execution, (b) dispatching both trajectory goals before waiting, (c) a staged lift with a gripper-gap guard, and (d) the second attempt and its verification.}
\label{fig_app_twoarm_strip}
\end{figure}

\begin{figure}[h]
\centering
\begin{subfigure}[t]{0.19\linewidth}\includegraphics[width=\linewidth]{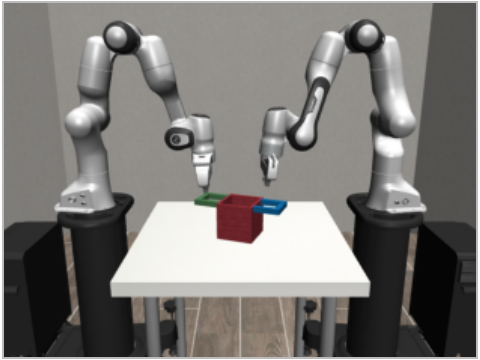}\caption{Pot between the arms.}\end{subfigure}\hfill
\begin{subfigure}[t]{0.19\linewidth}\includegraphics[width=\linewidth]{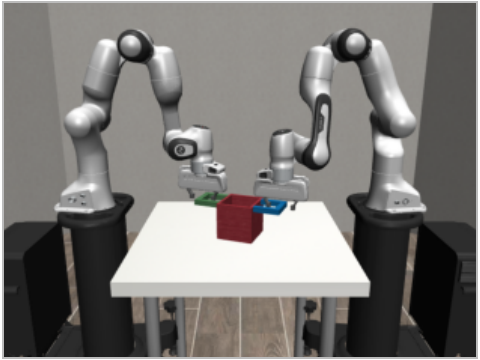}\caption{Both grippers on the handles.}\end{subfigure}\hfill
\begin{subfigure}[t]{0.19\linewidth}\includegraphics[width=\linewidth]{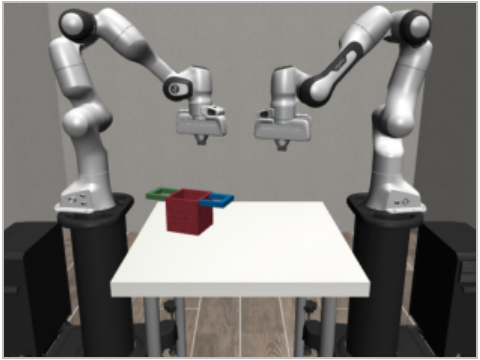}\caption{First lift: slipped.}\end{subfigure}\hfill
\begin{subfigure}[t]{0.19\linewidth}\includegraphics[width=\linewidth]{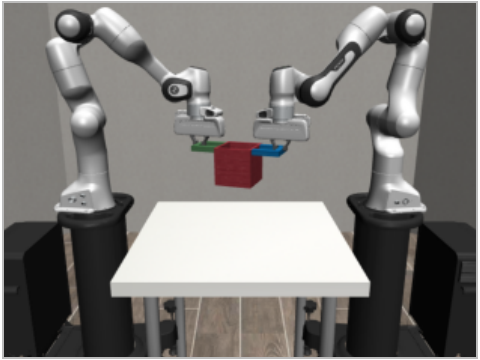}\caption{Second lift.}\end{subfigure}\hfill
\begin{subfigure}[t]{0.19\linewidth}\includegraphics[width=\linewidth]{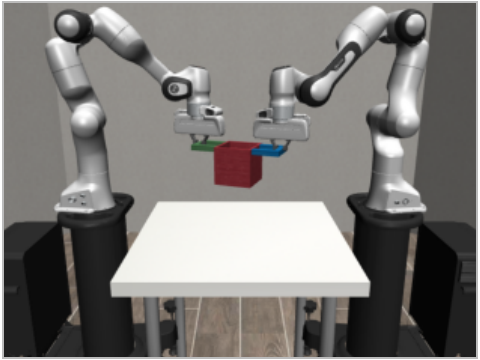}\caption{Held level.}\end{subfigure}
\caption{Front-camera checkpoints from the CaP-Bench Two-Arm Lift episode in~\cref{fig_app_twoarm_strip}, showing the failed first lift and successful second attempt.}
\label{fig_app_twoarm_moments}
\end{figure}

\parabf{The agent writes kinematics and coordinated control code for a two-arm lift.}
In the CaP-Bench episode shown in~\cref{fig_app_twoarm_strip,fig_app_twoarm_moments}, no motion planner is available on the machine.
The agent writes forward and inverse kinematics for the Panda and checks its forward-kinematics results against the transform tree to within 0.5\,mm.
It then constructs a robot class whose \texttt{move\_both} method sends both trajectory goals before waiting for either, allowing the arms to move together.
Each task phase runs as a command that prints verification measurements, and the lift proceeds in three stages with an abort if either finger gap collapses.
The first attempt slips; the second maintains its grip, and a final camera measurement places the two handle tops within 0.8\,mm of one another.
This example connects agent-written robot models, coordinated commands, and execution checks within one task.


\clearpage
\section{Scaling Analysis}
\label[appendix]{app_scaling}
\begin{table}[t]
\centering
\footnotesize
\setlength{\tabcolsep}{4pt}
\renewcommand{\arraystretch}{1.15}
\caption{\textbf{Higher success can coincide with fewer interactions and lower cost per success.}
Profiles of \harness on the LIBERO-10 suite of LIBERO-PRO cover both perturbation cells, with 200 trials per configuration.
Turns and output tokens per turn, including thinking, are medians over episodes; parenthesized turn counts separate successful and failed episodes.
A turn is one model response.
Cost is the mean per episode at API list prices, and cost per success is total cost divided by the number of successful episodes.
Logging coverage and cost accounting are described below; \cref{fig_scaling_profile} shows the distributions.}
\label{tab_scaling}
\resizebox{\textwidth}{!}{%
\begin{tabular}{llccccc}
\toprule
Agent & Model (Effort) & Success (\%) & \makecell{Turns, median\\(succ. / fail.)} & \makecell{Tokens / turn,\\median} & \makecell{Cost (\$),\\mean} & \makecell{Cost per\\success (\$)} \\
\midrule
CC & Sonnet 5 (high) & 24.0 & 301 (219 / 334) & 512 & 22.4 & 93.3 \\
CC & Opus 5 (high) & 72.5 & 103 (97 / 131) & 1,058 & 13.7 & 18.9 \\
CC & Fable 5.1 (high) & 75.0 & 73 (72 / 84) & 760 & 8.4 & 11.2 \\
Codex & GPT-6 Astra (medium) & 62.5 & 98 (103 / 89) & 90 & 11.8 & 19.1 \\
Codex & GPT-6 Astra (high) & 69.0 & 93 (97 / 86) & 122 & 10.0 & 14.4 \\
\bottomrule
\end{tabular}}
\end{table}

\paragraph{Runs behind \cref{tab_scaling}.}
All runs use the same harness, workspace, and prompt on the two LIBERO-10 perturbation cells (10 tasks $\times$ 10 initial states each).
Claude Code with Opus 5 and Fable 5.1 ran the full 200 episodes once at high reasoning effort, Claude Code's default.
Claude Code with Sonnet 5 ran the full 200 episodes at high effort.
Codex with GPT-6 Astra ran the full 200 episodes twice, once at medium reasoning effort and once at high; the two runs are reported as two rows.
The turn and cost profiles use the episodes with tool-call timestamps: all 200 for GPT-6 Astra at high effort, 192 for medium (the first eight episodes of that run were made before rollout logging was enabled).
\parabf{More interaction and greater reasoning effort improve success under the same native interface.}
The curves in~\cref{fig_scaling} are computed retrospectively from the first successful step of each recorded episode: an episode contributes to the success rate at a given budget if success occurred within that budget.
The agent-turn and output-token curves therefore describe when the recorded runs succeed, rather than separate runs in which the agent is instructed to work under each plotted budget.
For Opus~5, the share solved within 100, 150, and 200 turns is 50\%, 66\%, and 70\%, respectively; for Fable~5.1, it is 61\%, 71\%, and 72\%.
The reasoning-effort comparison uses separate 200-episode runs with the same Codex agent and GPT-6 Astra model: increasing effort from medium to high raises success from 62.5\% to 69.0\%.
These improvements occur without changing the initial workspace, adding task-specific primitives, or training a robot policy.
The output-token curves provide a complementary view of the computation consumed before success; the estimation of Claude Code token positions is described below.

\parabf{Performance gains diminish while differences between models persist.}
The Opus~5 and Fable~5.1 curves flatten within about 300 turns, whereas Sonnet~5 continues to gain successes up to about 400 turns and finishes at 24.0\%.
Within 100 turns, Fable~5.1 solves 61\% of episodes, Opus~5 solves 50\%, and Sonnet~5 solves 1\%.
These differences show that models vary both in final success and in the interaction needed to reach it.
Failed Claude Code episodes are longer than successful ones: median turns are 131 versus 97 for Opus~5, 84 versus 72 for Fable~5.1, and 334 versus 219 for Sonnet~5.
For Sonnet~5, 22\% of episodes reach the 500-turn limit, and none of those episodes succeeds.
Together, the curves and episode lengths show diminishing returns within the evaluated budgets, with substantial differences remaining between models.
They do not establish that extending a failed episode beyond the evaluated limit could never lead to success.

\parabf{Higher success can coincide with lower cost per success.}
As shown in~\cref{tab_scaling}, success rises from 24.0\% with Sonnet~5 to 72.5\% with Opus~5 and 75.0\% with Fable~5.1, while median turns fall from 301 to 103 and 73.
At API list prices, the corresponding cost per success falls from \$93.3 to \$18.9 and \$11.2, even though Sonnet~5 has the lowest per-token prices among these models.
Thus, lower token prices do not necessarily translate into lower cost per successful task: episode length and success rate also matter.
For GPT-6 Astra, increasing reasoning effort raises median output tokens per turn from 90 to 122, but reduces mean episode cost from \$11.8 to \$10.0 and cost per success from \$19.1 to \$14.4.
A larger output per turn therefore need not increase the cost of completing a task; total cost also includes input tokens and cache usage, as detailed below.

\parabf{Different coding agents use the same workspace and native interface.}
Claude Code and Codex use the same workspace, documentation, and native ROS~2 interface without agent-specific changes to the harness.
Their interaction profiles nevertheless differ: Codex with GPT-6 Astra produces a median of 90 or 122 output tokens per turn at medium or high effort, compared with 760 for Fable~5.1 and 1,058 for Opus~5.
At high effort, failed Codex episodes are shorter than successful episodes, with medians of 86 and 97 turns, respectively, unlike the Claude Code runs.
These profiles show that compatibility with the native interface does not require agents to follow the same interaction pattern.
The comparison varies both the coding agent and the model, so the observed differences characterize the evaluated combinations rather than isolating either factor.

\paragraph{Per-cell success.}
LIBERO-10 swap cells 0--9 / task cells 0--9, solved out of 10:
Sonnet 5: 8, 7, 0, 1, 0, 2, 1, 8, 0, 0 / 8, 3, 0, 0, 0, 3, 0, 4, 3, 0;
Opus 5: 9, 9, 10, 10, 9, 10, 8, 9, 6, 0 / 9, 8, 10, 5, 8, 0, 6, 10, 0, 9;
Fable 5.1: 9, 9, 10, 6, 9, 10, 3, 10, 9, 5 / 10, 8, 10, 5, 9, 7, 7, 7, 0, 7;
GPT-6 Astra, medium: 10, 9, 10, 10, 5, 3, 0, 9, 8, 2 / 10, 8, 10, 9, 2, 4, 0, 9, 0, 7;
GPT-6 Astra, high: 10, 9, 10, 9, 10, 1, 0, 9, 9, 8 / 10, 9, 10, 7, 5, 2, 1, 9, 0, 10.
Across the five evaluated configurations, five of the twenty cells have a spread of at most three successful episodes.
Comparing Opus 5 with GPT-6 Astra at high reasoning effort, the absolute difference exceeds three successful episodes in four cells (swap 5, 6, 9 and task 6).
\paragraph{Token budget (\cref{fig_scaling_tokens}).}
The Codex rollout log records the running output-token total after every model response, so the token at which the predicate first held is read directly.
Claude Code's transcript reports only the episode total; the split over turns is estimated from the thinking-token estimates the CLI streams between messages plus the visible text and tool inputs (at four characters per token), and the cumulative series is scaled so that the episode ends at the reported total.
The endpoint of every episode is therefore exact; the shape of the Claude curves within an episode is an estimate.
\paragraph{Cost accounting.}
Costs are computed from the token counts each CLI reports, at API list prices on 2026-09-20 (USD per million tokens): Opus 5 input 5, cache read 0.5, one-hour cache write 10, output 25; Fable 5.1 input 10, cache read 0.25, cache write 20, output 50; Sonnet 5 input 2, cache read 0.2, cache write 4, output 10; GPT-6 Astra input 10, cached input 1, output 50.
The runs themselves were billed on subscriptions; list prices are used only for comparability.
\begin{figure}[h]
\centering
\includegraphics[width=\linewidth]{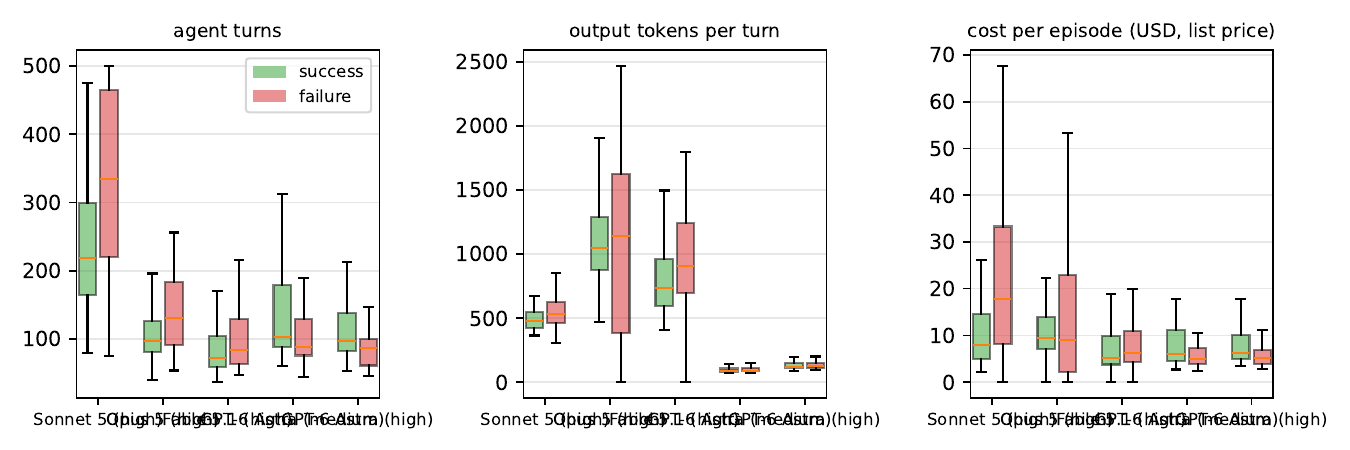}
\caption{\textbf{Interaction and cost distributions differ across agent--model configurations and episode outcomes.}
Agent turns, output tokens per turn, and episode costs are shown separately for successful and failed runs on the LIBERO-10 suite of LIBERO-PRO.
Costs use API list prices, as in~\cref{tab_scaling}.
Boxes span the interquartile range; whiskers extend to the most extreme values within 1.5 times that range.}
\label{fig_scaling_profile}
\end{figure}

\clearpage
\section{Failure Study}
\label[appendix]{app_failure}
This appendix expands the failure study in~\cref{sec_failure}, including the classification procedure, detailed examples, and sub-causes.

We analyzed 413 failed episodes of \harness with Opus 5 from the evaluations in \cref{tab_main,tab_robocasa}, and assigned each to the stage at which its decisive error lies (\cref{tab_failure}; the procedure and the class definitions are given below, and \cref{tab_failure_detail} lists the sub-causes).

\subsection{Classification Procedure and Scope}
\label[appendix]{app_failure_procedure}

\paragraph{Procedure.}
The study analyzes 413 failed episodes of \harness with Claude Code and Opus 5 in the main runs: 104 on LIBERO-PRO, 7 on CaP-Bench and 302 on RoboCasa365. For each episode we read the transcript and the agent's workspace, recovered the benchmark predicate (the BDDL goal and region definitions on LIBERO, the task class's success check on RoboCasa365, the stated condition on CaP-Bench), and compared the end state the agent measured with that predicate; for RoboCasa365 the scene was re-instantiated from the episode seed to recover quantities the predicate uses but the agent cannot observe, such as region boxes and target poses. To see what the demonstrations teach, we replayed the final state of each of the 50 LIBERO demonstrations for three of the tasks in the benchmark environment: the middle drawer ends at 127 to 160\,mm of travel (median 156, the joint's limit is 160), the plate ends within 4\,cm of the centre of the stove-front region in 49 of 50, and the bottle's base rests in the rack's 22\,mm band in the 25 demonstrations whose final state satisfies the predicate. Each episode receives one class, the stage at which the decisive error lies, and a record of whether its last message announces completion. %

\paragraph{Classes.}
A: the agent completed and verified its own reading of the instruction, and the predicate needs a quantity or a region the instruction does not state. B: the predicate contradicts the instruction (it scores another object, or a region the named object cannot occupy). C: the budget is exhausted while the agent handles undocumented machine behavior, with the task still in progress and no other decisive error identified (a base without a brake, doors that close themselves, fixtures whose scored joint is not the one that moves). D: the object or the required force lies at or beyond the limits of the gripper, the base or the arm's reach. E: an error of the agent itself, in perception (a target never searched for, a misread joint), in control (a wrong gripper width, a release before arrival, a trajectory through an obstacle), or in waiting (budget spent polling background jobs). \cref{tab_failure_detail} lists the sub-causes.

\paragraph{Limits.}
The study reads one model's episodes; the other models in \cref{tab_scaling} were not classified. Classes were assigned by reading and are recorded with a confidence level; five LIBERO-PRO episodes (a bottle-in-drawer task) could not be resolved between class A and a residual drawer opening below the point-cloud resolution and are counted under A.

\subsection{Detailed Findings}

\begin{figure}[h]
\centering
\captionsetup[sub]{font=footnotesize,labelfont=footnotesize,skip=2pt}
\begin{subfigure}[t]{0.32\linewidth}\includegraphics[width=\linewidth]{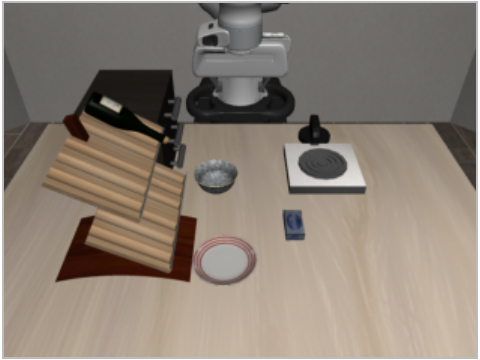}\caption{Bottle placement (class A).}\end{subfigure}\hfill
\begin{subfigure}[t]{0.32\linewidth}\includegraphics[width=\linewidth]{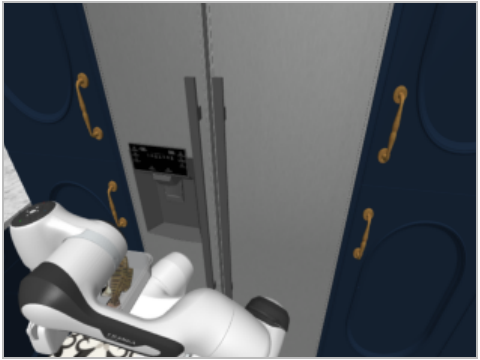}\caption{Self-closing door (class C).}\end{subfigure}\hfill
\begin{subfigure}[t]{0.32\linewidth}\includegraphics[width=\linewidth]{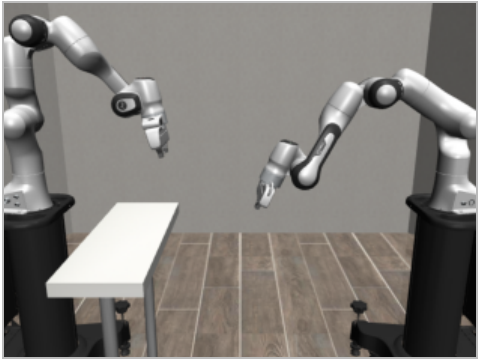}\caption{Handover error (class E).}\end{subfigure}
\caption{\textbf{Failures can involve an unstated completion criterion, undocumented machine behavior, or an agent control error.}
Agent-saved images show one episode from each benchmark: (a) LIBERO wine-rack placement (class A), (b) RoboCasa365 freezer loading (class C), and (c) CaP-Bench handover (class E).
\Cref{fig_fail_rack,fig_fail_freezer,fig_fail_handover} show the corresponding checkpoint sequences; \cref{fig_fail_moka} illustrates a class-B mismatch between instruction and predicate.}
\label{fig_failure_cases}
\end{figure}

\parabf{Task criteria and undocumented machine behavior account for many failures.}
Classes A, B, and C account for 238 of the 413 analyzed failures (58\%), as in~\cref{sec_failure}.
In class A, the agent completes and verifies its interpretation of the instruction, but the predicate requires an unstated threshold or spatial region.
For example, in 14 drawer episodes it pulls 62--136\,mm and stops before a bottle in the drawer's path, while the predicate requires more than 140\,mm.
In seven wine-rack episodes, it places the bottle along the slope to prevent rolling, but the predicate checks its base against a 22\,mm band at the foot of the rack (\cref{fig_failure_cases}(a)).
In nine board-scrubbing episodes, measured contact spans 3--10\,cm, while the predicate requires a 10\,cm sweep.

Class B covers 19 episodes in which the predicate conflicts with the instruction, including a goal that scores the pot initialized on the opposite side and a 55\,mm target slot for a 100\,mm mug.
Class C covers budgets exhausted while handling undocumented machine behavior.
Examples include an unbraked base displaced by 0.15--0.39\,m during contact, freezer doors that close while the robot fetches another tray (\cref{fig_failure_cases}(b)), and a faucet with two hinges of which only one is scored.
The agent discovers these behaviors during execution and spends its budget writing position-holding code or repeating earlier actions.

Demonstrations can convey completion conditions that the instruction leaves implicit.
In our replay of LIBERO demonstrations, the drawer ends at a median 156\,mm of its 160\,mm travel, the plate rests inside the goal region, and the bottle's base lies at the foot of the rack; \cref{app_failure_procedure} gives the sample details.
A zero-shot agent must instead infer these conditions from its instruction and observations.
These cases expose a gap between task instructions, machine behavior, and completion criteria in current benchmarks.

\parabf{Physical constraints remain a challenge for agent-written programs.}
Class D accounts for 75 failures (18\%), involving gripper geometry, contact forces, and reachability.
The twelve LIBERO-PRO cases include a 106\,mm bowl and 103\,mm mug rims against a 72--80\,mm gripper opening, as well as plate rims squeezed out during lifting.
The same tasks succeed under other initial conditions, indicating limited operating margins rather than tasks that are universally impossible.

On RoboCasa365, 63 failures involve physical constraints of the gripper, base, or arm.
For example, the binary gripper can continue closing on a soft or thin object during transport; the finger gap on a straw decreases from 14 to 1\,mm over one base drive.
Contact forces above the base's 250\,N friction displace the robot and can raise its torso by up to 0.34\,m.
An object dropped onto the floor cannot be retrieved when it lies below the arm's reachable height of approximately $z=0.3$\,m.

\parabf{The agent also makes control, perception, and execution-management errors.}
Class E accounts for 100 failures (24\%): 46 control errors, 27 perception errors, and 27 waiting errors.
Control errors include commanding 0.008\,m per finger on a measured 0.030\,m handle, despite the documentation's warning, and ejecting the hammer when the other arm releases it (\cref{fig_failure_cases}(c)).
Other cases involve an inverse-kinematics branch change that drives the hand below the table, releasing before arrival, or knocking over an already placed object during recovery.

The 27 perception errors comprise three LIBERO-PRO and 24 RoboCasa365 cases.
Many RoboCasa365 cases involve a target outside the current view that the agent never searches for, or an unobserved joint such as the scored hinge of a faucet.
The 27 waiting errors involve budget spent polling background jobs or episodes ending while a monitor remains armed.
Thus, the ability to write measurement and feedback programs does not ensure that the agent uses them correctly or verifies every required condition.

\begin{figure}[h]
\centering
\captionsetup[sub]{font=footnotesize,labelfont=footnotesize,skip=2pt}
\begin{subfigure}[t]{0.24\linewidth}\includegraphics[width=\linewidth]{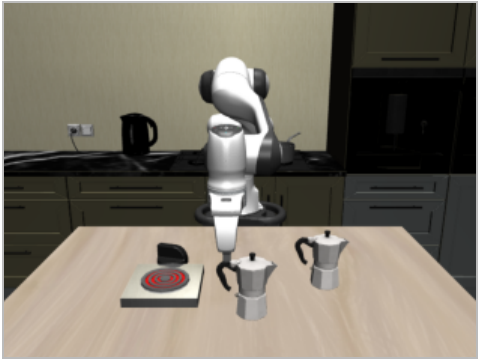}\caption{Start: two identical pots.}\end{subfigure}\hfill
\begin{subfigure}[t]{0.24\linewidth}\includegraphics[width=\linewidth]{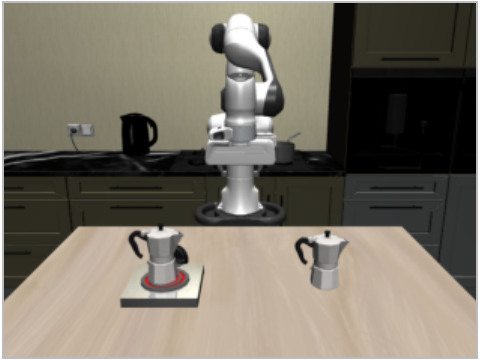}\caption{End: the pot the agent resolved as ``left'' is on the burner.}\end{subfigure}\hfill
\begin{subfigure}[t]{0.24\linewidth}\includegraphics[width=\linewidth]{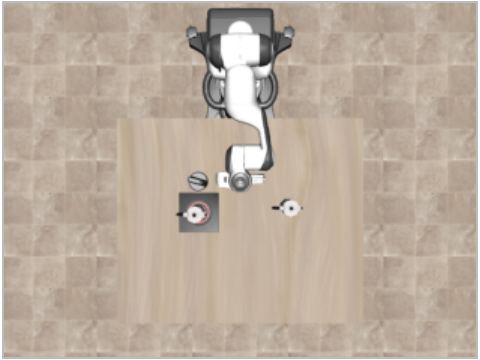}\caption{Overhead check, the other pot untouched.}\end{subfigure}\hfill
\begin{subfigure}[t]{0.24\linewidth}\includegraphics[width=\linewidth]{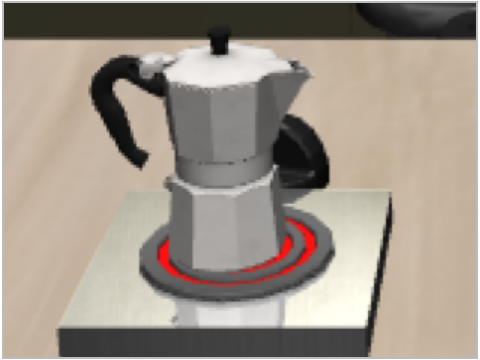}\caption{The agent's crop: axis 1.2\,mm from the burner centre.}\end{subfigure}
\caption{\textbf{The benchmark scores a different moka pot from the one specified by the instruction.}
This class-B case comes from a LIBERO-PRO variant of ``put the left moka pot on the stove'' (nine of ten seeds; seed 2 shown).
The BDDL file places \texttt{moka\_pot\_2} in \texttt{moka\_pot\_left\_init\_region} and \texttt{moka\_pot\_1} in \texttt{moka\_pot\_right\_init\_region}, but its goal is \texttt{(On moka\_pot\_1 flat\_stove\_1\_cook\_region)}.
After checking the image-right direction of all seven cameras, the agent moves the pot from the left region and places it 1.2\,mm from the burner center, with its full weight supported by the stove.
The predicate instead checks the other pot.}
\label{fig_fail_moka}
\end{figure}

\begin{figure}[h]
\centering
\captionsetup[sub]{font=footnotesize,labelfont=footnotesize,skip=2pt}
\begin{subfigure}[t]{0.24\linewidth}\includegraphics[width=\linewidth]{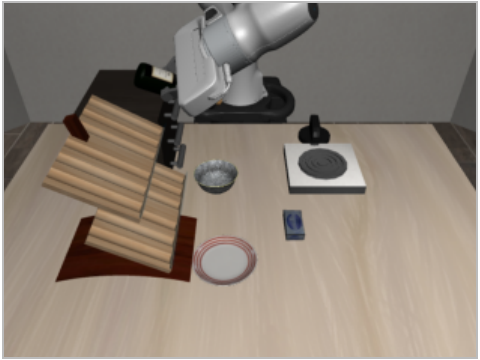}\caption{Pre-place pose over the rack.}\end{subfigure}\hfill
\begin{subfigure}[t]{0.24\linewidth}\includegraphics[width=\linewidth]{figures/fail_rack_b.pdf}\caption{Released: bottle along a groove.}\end{subfigure}\hfill
\begin{subfigure}[t]{0.24\linewidth}\includegraphics[width=\linewidth]{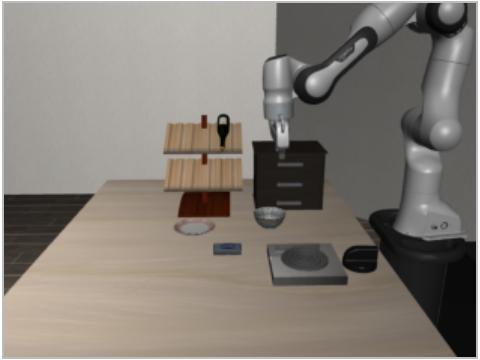}\caption{Side view after settling.}\end{subfigure}\hfill
\begin{subfigure}[t]{0.24\linewidth}\includegraphics[width=\linewidth]{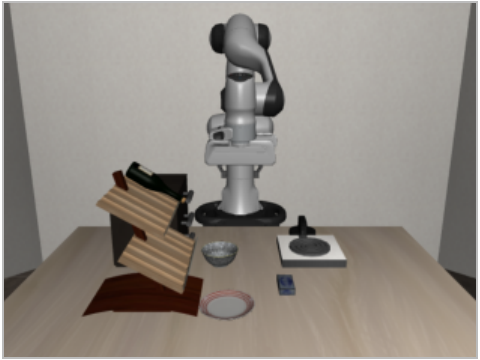}\caption{Front view after settling.}\end{subfigure}
\caption{\textbf{A stable bottle placement misses the benchmark's unstated target region.}
In this class-A LIBERO wine-rack case (seven of seven failed seeds; seed 5 shown), the agent models the shelf as a $28.6^\circ$ corrugated plane and lays the bottle along a groove to prevent rolling.
Its recorded reasoning is ``laying the bottle along a groove was the only stable option: cross-ways it rolls off''.
The predicate checks the bottle's base against \texttt{wine\_rack\_1\_top\_region}, a 22\,mm band at the foot of the slope; the base rests 40\,mm up the slope from that band and 10\,mm outside its width.
In the benchmark demonstrations, the bottle's base rests in the band.}
\label{fig_fail_rack}
\end{figure}

\begin{figure}[h]
\centering
\captionsetup[sub]{font=footnotesize,labelfont=footnotesize,skip=2pt}
\begin{subfigure}[t]{0.24\linewidth}\includegraphics[width=\linewidth]{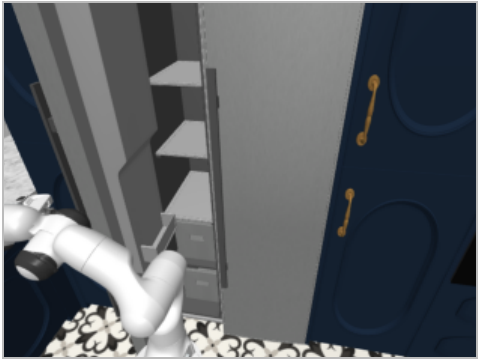}\caption{Freezer door opened.}\end{subfigure}\hfill
\begin{subfigure}[t]{0.24\linewidth}\includegraphics[width=\linewidth]{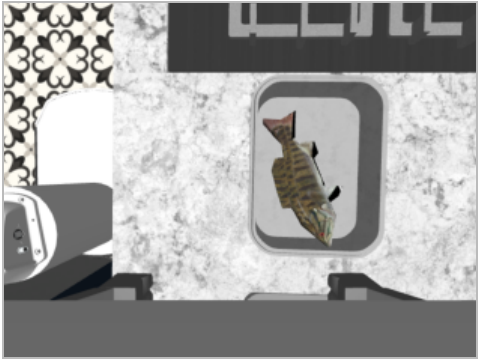}\caption{Tray grasped at the counter, 0.8\,m away.}\end{subfigure}\hfill
\begin{subfigure}[t]{0.24\linewidth}\includegraphics[width=\linewidth]{figures/fail_freezer_c.pdf}\caption{Back at the freezer: door shut again.}\end{subfigure}\hfill
\begin{subfigure}[t]{0.24\linewidth}\includegraphics[width=\linewidth]{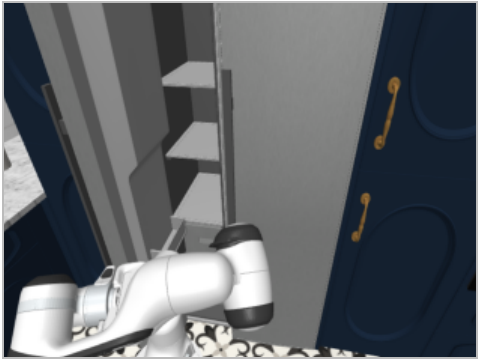}\caption{Door opened a third time.}\end{subfigure}
\caption{\textbf{Repeatedly reopening a self-closing freezer door exhausts the episode budget.}
This class-C RoboCasa365 case shows seed 4 of ``move the trays into the freezer''; all ten failed seeds reach the wall-clock limit.
The door closes while the robot fetches a tray from the counter, forcing it to reopen the door on each return.
The agent labels its move after (c) ``Retract arm from the closed door''.
The unbraked base also slides toward the fridge during pulling and can push the door shut.
No dropped object or misread observation was identified as the decisive error.}
\label{fig_fail_freezer}
\end{figure}

\begin{figure}[h]
\centering
\captionsetup[sub]{font=footnotesize,labelfont=footnotesize,skip=2pt}
\begin{subfigure}[t]{0.24\linewidth}\includegraphics[width=\linewidth]{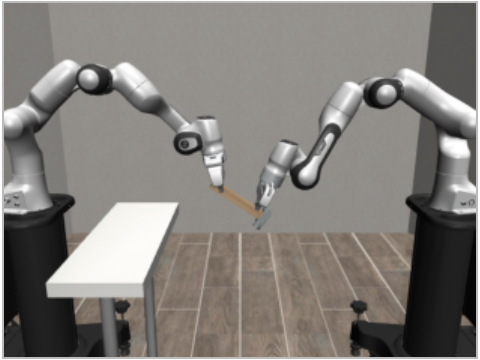}\caption{Both arms on the hammer.}\end{subfigure}\hfill
\begin{subfigure}[t]{0.24\linewidth}\includegraphics[width=\linewidth]{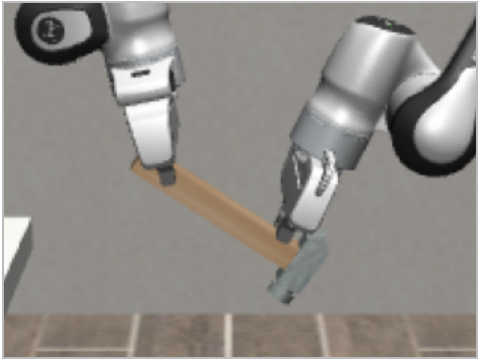}\caption{The agent's crop: handle at $33^\circ$.}\end{subfigure}\hfill
\begin{subfigure}[t]{0.24\linewidth}\includegraphics[width=\linewidth]{figures/fail_handover_c.pdf}\caption{After arm 0 opens: nothing in either gripper.}\end{subfigure}\hfill
\begin{subfigure}[t]{0.24\linewidth}\includegraphics[width=\linewidth]{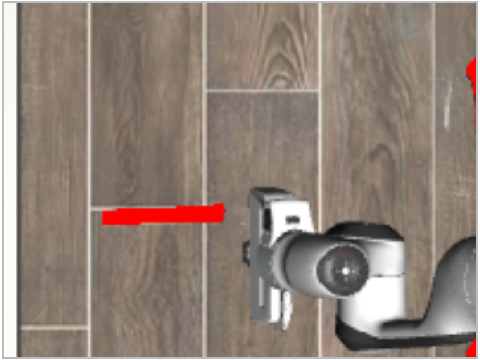}\caption{The agent's floor mask: hammer 0.5\,m out of reach.}\end{subfigure}
\caption{\textbf{An incorrect gripper command ejects the hammer during handover.}
This class-E CaP-Bench case shows seed 58; four of six failed seeds share this mechanism.
The agent measures a 0.030\,m handle but commands arm 1 to 0.008\,m per finger, despite the workspace warning: ``Command a target width, never 0 against an object; 0 keeps squeezing and can eject it''.
When arm 0 releases, arm 1's finger positions decrease from 0.0123 to 0.0010\,m and the hammer falls to the floor.
Earlier, arm 0 held the hammer 0.21\,m from the head, allowing its angle to increase from $3^\circ$ to $33^\circ$ during transport.}
\label{fig_fail_handover}
\end{figure}

\paragraph{By benchmark.}
On LIBERO-PRO, classes A and B hold 59 of 104 failures, and every one of these episodes ends with a completion message that states the agent's own criterion; the 17 class-B episodes come from two task variants whose predicate names a different object than the instruction or a region the object cannot fit. On CaP-Bench, six of the seven analyzed failures occur in Handover and are classified as control errors; four share the gripper-command mechanism illustrated in~\cref{fig_fail_handover}. On RoboCasa365, classes C and D hold 188 of 302 failures: the base is a velocity-servo joint with 250\,N of friction and no brake, so contact forces above that move the robot, and the arm cannot reach the floor, so a dropped object ends the episode; 76 episodes spent their budget on doors, handles and levers, and 49 on holding station. \cref{fig_fail_moka,fig_fail_rack,fig_fail_freezer,fig_fail_handover} show one episode for each of classes B, A, C and E in the images the agent saved.

\begin{table}[h]
\centering
\footnotesize
\setlength{\tabcolsep}{3pt}
\renewcommand{\arraystretch}{1.15}
\caption{\textbf{Detailed causes within the five failure categories.}
Counts are analyzed failed episodes with Claude Code and Opus~5, grouped as in~\cref{tab_failure}.
Thresholds come from the benchmark predicates: LIBERO joint ranges and BDDL region boxes, and RoboCasa365 task success checks.}
\label{tab_failure_detail}
\begin{tabular}{l>{\raggedright\arraybackslash}p{7.6cm}rrr}
\toprule
Class & Sub-cause & LIBERO & CaP & RoboCasa \\
\midrule
\multirow{5}{*}{A} & Joint threshold: ``open'' needs $>140$\,mm or $\geq 95\%$ of travel, a mixer head $\geq 99\%$; lift needs $>0.10$\,m & 14 & 1 & 6 \\ \addlinespace[4pt]
& Region box differs from the words: ``in front of the stove'', ``right of the plate'', ``on the rack'', ``in the microwave'' & 19 & 0 & 0 \\ \addlinespace[4pt]
& Tolerance of \texttt{On}: body origins within 30\,mm (a mug's origin sits 18\,mm toward its handle) & 2 & 0 & 0 \\ \addlinespace[4pt]
& Final check coarser than the predicate: drawer read as shut from a point cloud, door ``closed'' needs $<0.5\%$ of travel, lid seated within 4\,cm and $7^\circ$ & 7 & 0 & 10 \\ \addlinespace[4pt]
& Stand-off after release ($>0.15$ / $0.25$\,m), contact and timer clauses, sweep $\geq 0.10$\,m, ``do not touch the food'', navigation pose within 0.20\,m & 0 & 0 & 20 \\
\midrule
\multirow{3}{*}{B} & Predicate names the other object (``left moka pot'' scores the pot in the BDDL's right init region) & 9 & 0 & 0 \\ \addlinespace[4pt]
& Target region cannot hold the object (100\,mm mug, 55\,mm caddy slot) & 8 & 0 & 0 \\ \addlinespace[4pt]
& ``Right drawer'' resolved relative to the spawn pose & 0 & 0 & 2 \\
\midrule
\multirow{2}{*}{C} & Budget exhausted while handling fixtures: self-closing doors, round-bar handles, faucet levers, toaster lever & 0 & 0 & 76 \\ \addlinespace[4pt]
& Budget exhausted while holding station against base recoil and drift; no other decisive error & 15 & 0 & 49 \\
\midrule
\multirow{3}{*}{D} & Object at the edge of the gripper envelope; binary gripper extrudes soft, thin or round objects & 12 & 0 & 54 \\ \addlinespace[4pt]
& Base pushed or lifted by contact forces (velocity-servo base, 250\,N friction, no brake) & 0 & 0 & 7 \\ \addlinespace[4pt]
& Object out of the arm's reach after a drop (floor below $z \approx 0.3$\,m) & 0 & 0 & 2 \\
\midrule
\multirow{3}{*}{E} & Perception: second candidate of the same kind never resolved; target outside the parked view never searched; a joint the sensors do not expose (faucet hinge, knob modulo $180^\circ$, button geometry) & 3 & 0 & 24 \\ \addlinespace[4pt]
& Control: half the measured handle width commanded, IK branch jump, release before arrival, recovery sweep knocks the placed object & 12 & 6 & 28 \\ \addlinespace[4pt]
& Waiting: turns or wall clock spent polling background jobs; turn ended while a monitor was armed & 3 & 0 & 24 \\
\midrule
& Failed episodes & 104 & 7 & 302 \\
\bottomrule
\end{tabular}
\end{table}

\subsection{Failure Cases}

\begin{figure}[h]
\centering
\begin{subfigure}[t]{0.19\linewidth}\includegraphics[width=\linewidth]{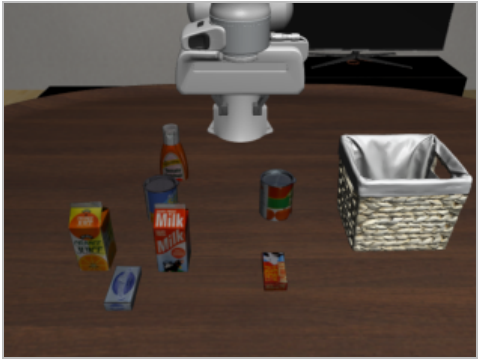}\caption{Scene: two cans.}\end{subfigure}\hfill
\begin{subfigure}[t]{0.19\linewidth}\includegraphics[width=\linewidth]{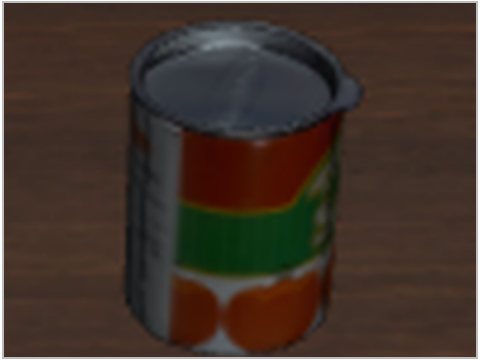}\caption{The only can it inspected.}\end{subfigure}\hfill
\begin{subfigure}[t]{0.19\linewidth}\includegraphics[width=\linewidth]{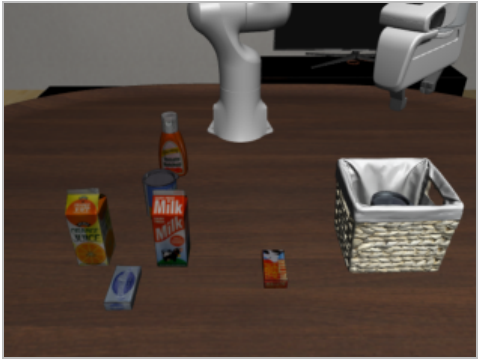}\caption{After the first pick.}\end{subfigure}\hfill
\begin{subfigure}[t]{0.19\linewidth}\includegraphics[width=\linewidth]{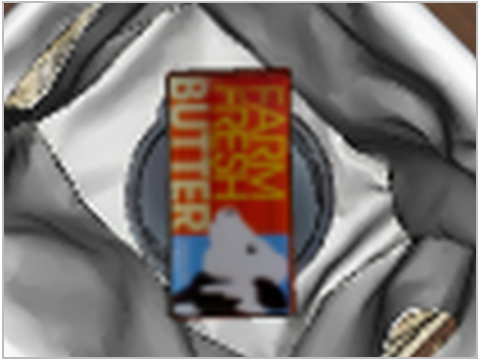}\caption{Its check of the basket.}\end{subfigure}\hfill
\begin{subfigure}[t]{0.19\linewidth}\includegraphics[width=\linewidth]{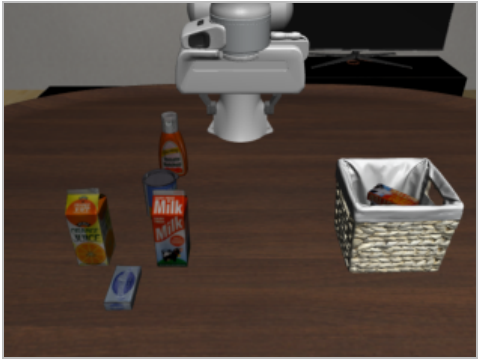}\caption{``Done'': the blue can remains.}\end{subfigure}
\caption{\textbf{Checking placement without checking object identity leads to a false completion.}
In this LIBERO-10 episode, the agent places the tomato can instead of the alphabet-soup can and reports success after inspecting the basket; (e) the requested blue can remains on the table.}
\label{fig_app_fail_moments}
\end{figure}

\parabf{Checking placement without checking object identity can produce a false completion.}
\Cref{fig_app_fail_moments} shows a LIBERO-10 episode in which the instruction asks for both the alphabet soup and the cream cheese in the basket.
The agent crops and identifies the tomato can in front but never inspects the blue alphabet-soup can behind the milk carton.
It moves the tomato can as if it were the alphabet soup, places it and the cream cheese in the basket, and verifies placement without checking identity.
Its final message reads ``Done --- both objects are in the basket'', although the requested blue can remains on the table.
The episode illustrates a missing perceptual check despite successful object transport; it is counted in~\cref{app_failure}.

\clearpage
\section{Initial Workspace Materials}
\label[appendix]{app_workspace_materials}

\subsection{Initial Materials and File Inventory}

We reproduce the complete workspace template below: six Markdown documents and six Python starter tools.
The listings retain the original file contents; long lines are wrapped for typesetting, and Python line numbers are added for reference.
These are the materials supplied before the agent begins the task; programs subsequently written by the agent are presented separately in~\cref{app_programs}.

\begin{center}
\small
\renewcommand{\arraystretch}{1.12}
\begin{tabular}{@{}p{0.40\linewidth}p{0.56\linewidth}@{}}
\toprule
File & Supplied content \\
\midrule
\texttt{README.md} & Entry point, file map, and example discovery commands. \\
\texttt{docs/10-machine.md} & Machine-manifest fields, ROS tools, installed software, and simulation timing. \\
\texttt{docs/20-perception.md} & Joint states, transforms, force/torque, odometry, cameras, and calibration. \\
\texttt{docs/30-action.md} & Trajectory, velocity, gripper, mobile-base, and planning interfaces. \\
\texttt{docs/40-patterns.md} & Generic verification, motion, grasping, troubleshooting, and session guidance. \\
\texttt{tools/README.md} & Tool index with usage and operation descriptions for all six utilities. \\
\texttt{tools/perception/cam\_snap.py} & Save an image topic as a PNG; also save a numerical array for depth. \\
\texttt{tools/perception/px2world.py} & Convert a selected depth pixel to world coordinates using intrinsics and TF. \\
\texttt{tools/action/fjt\_send.py} & Send a joint-trajectory action goal and wait for its result. \\
\texttt{tools/action/gripper\_cmd.py} & Send a gripper action goal and print the returned status. \\
\texttt{tools/action/ik\_move.py} & Request inverse kinematics and execute the resulting joint trajectory. \\
\texttt{tools/action/base\_goto.py} & Drive a mobile base toward a pose using odometry feedback. \\
\bottomrule
\end{tabular}
\end{center}

\paragraph{Generated machine description.}
The workspace root additionally receives \texttt{machine.yaml}, generated from the resolved robot configuration at startup rather than copied from the static template.
It contains \texttt{schema}, \texttt{robot}, \texttt{frames}, \texttt{actuators}, \texttt{sensors}, \texttt{planning}, and \texttt{hand} fields.
Actuator and sensor entries identify capability kinds, ROS ports and message types, with joint names, limits, coordinate frames, and gripper properties where applicable.
Multiple arms yield separate actuator and force-sensor entries; a mobile base adds velocity and odometry interfaces.
The manifest generator omits actuator entries without a port and sensor entries without a port or topic pattern, and adds explanatory comments for interface fields and units.
Consequently, the concrete manifest varies with the machine; the documents and tool source below constitute the shared template.
The opening prompt is reproduced in~\cref{fig_workspace_prompt}.

\begingroup
\lstdefinestyle{workspace-source}{
  basicstyle=\ttfamily\scriptsize,
  breaklines=true,
  breakatwhitespace=false,
  columns=fullflexible,
  keepspaces=true,
  showstringspaces=false,
  captionpos=t,
  literate={→}{{$\rightarrow$}}1 {–}{{{\normalfont\textendash}}}1 {…}{{\ldots}}1 {§}{{\S}}1 {≠}{{$\ne$}}1 {×}{{$\times$}}1
}
\lstdefinestyle{workspace-document}{
  style=workspace-source,
  language={},
  numbers=none,
  xleftmargin=0pt,
  keywordstyle={},
  commentstyle={},
  stringstyle={}
}
\lstdefinestyle{workspace-python}{
  style=workspace-source,
  language=Python,
  numbers=left
}

\subsection{Workspace Documentation}
\label[appendix]{app_workspace_documentation}
The following listings reproduce the Markdown documents as text, including their command examples and operating guidance.

\begin{lstlisting}[style=workspace-document,caption={\texttt{README.md}}]
# Workspace: start here

You are on the onboard computer of a robot arm workstation. This
workspace ships structured startup docs and a few generic tools.

## Map

| Read | For |
|---|---|
| [machine.yaml](machine.yaml) | THIS machine's facts: robot model, ports, joints, limits, frames |
| [docs/10-machine.md](docs/10-machine.md) | how to read machine.yaml; software stack, runtime environment, timing |
| [docs/20-perception.md](docs/20-perception.md) | how to sense: joints, poses, force, cameras |
| [docs/30-action.md](docs/30-action.md) | how to move: trajectories, servo, gripper, mobile base, planning |
| [docs/40-patterns.md](docs/40-patterns.md) | reliable working patterns and troubleshooting |
| [tools/README.md](tools/README.md) | ready-made utilities (perception/ and action/) |

## 30-second start

```bash
ros2 topic list                                    # what the machine exposes
ros2 topic echo /joint_states --once               # where the arm is
python3 tools/perception/cam_snap.py agentview     # what the scene looks like
```

Docs are numbered in reading order; each capability section follows the
same shape: what it is → port → `ros2` CLI → `rclpy` → tool shortcut →
practical notes.
\end{lstlisting}

\begin{lstlisting}[style=workspace-document,caption={\texttt{docs/10-machine.md}}]
# 10. The machine

Top-down: where the facts live → software stack → runtime environment →
timing.

## 1. Machine facts live in `machine.yaml`

This doc set is generic; everything specific to THIS machine; robot
model, joint names and limits, port names, frames, planning setup; is
in **`machine.yaml`** at the workspace root. Read it first. Structure:

- `robot`; model and one-line description
- `frames`; the TF names for `world`, the arm `base`, the `hand`
- `actuators`; a list of typed write-ports, each entry:
  `kind` (`joint_trajectory` / `cartesian_twist` / `gripper` / …),
  `port`, message `type`, plus kind-specific facts (joint names,
  limits, open/closed positions)
- `sensors`; a list of typed read-ports (`joint_state`, `wrench`,
  `camera_set`, …)

Every key in it is defined in a comment on its own line, units included;
`machine.yaml` explains itself and needs nothing from this doc.
- `planning`; MoveIt entry points, if present

Each `kind` has a how-to section in [20-perception.md](20-perception.md)
or [30-action.md](30-action.md). The manifest is an index of the ROS
graph; `ros2 topic list` / `ros2 action list` show the same ports live
(and anything beyond the manifest).

## 2. Software stack

- **ROS 2** is the control middleware: everything is a topic, service
  or action. **`ros2` CLI** for shell access, **`rclpy`** for Python.
- The `ros2` CLI verbs that cover daily work:

  | Verb | Does |
  |---|---|
  | `ros2 topic list` / `info <t>` / `echo <t> --once` | find, inspect, read topics |
  | `ros2 topic pub [--rate N] <t> <type> '{...}'` | publish (once or streaming) |
  | `ros2 action list` / `send_goal <a> <type> '{...}'` | find and call actions |
  | `ros2 service list` / `call <s> <type> '{...}'` | find and call services |
  | `ros2 node list` | who is running |
  | `ros2 interface show <type>` | field layout of any message type |
  | `ros2 run tf2_ros tf2_echo <from> <to>` | read a TF transform |

- Read `machine.yaml` in Python:
  ```python
  import yaml
  M = yaml.safe_load(open("machine.yaml"))
  fjt = next(a for a in M["actuators"] if a["kind"] == "joint_trajectory")
  ```

## 3. Runtime environment

- ROS is **already sourced** in every shell (bash and sh alike);
  no need to `source` manually.
- Python 3 has `rclpy`, `numpy`, `scipy`, `cv2`, `PIL`, `cv_bridge`,
  `tf2_ros`, `control_msgs`, `moveit_msgs`, `PyKDL`, `urdf_parser_py`.
- **This directory (`/workspace`) is yours**: write scripts, images,
  notes here.
- **No internet.** `pip install` and web access fail; everything needed
  is installed.

## 4. Timing

The system clock (`/clock`) advances with robot activity, not wall time.
Consequences:

- `sleep` does not let the physical world settle;
- confirm every motion by re-reading sensors after the command
  completes, not by waiting.

## Timing facts (paused-clock machine)

- At session start the scene may not have settled yet (objects can sit
  slightly above their supports); the world only advances while commands
  execute; waiting wall-clock time changes nothing.
- A few sim-seconds of trajectory can take minutes of wall time to
  return under load. A client-side timeout is NOT a motion failure;
  verify against `/joint_states` before re-sending.
\end{lstlisting}

\begin{lstlisting}[style=workspace-document,caption={\texttt{docs/20-perception.md}}]
# 20. Perception

Two families: **A. proprioception** (the robot's own state) and
**B. vision** (cameras). Concrete port names come from `machine.yaml`
(`sensors` list); sections below are keyed by sensor `kind`. Every
section: what → port → CLI → rclpy → tool → notes.

## A. Proprioception

### A1. Joint state (kind `joint_state`)

- **What**: positions/velocities/efforts for every joint.
- **Port**: `machine.yaml → sensors[kind=joint_state].port`
  (`sensor_msgs/msg/JointState`).
- **CLI**: `ros2 topic echo <port> --once`
- **rclpy**: subscribe once;
  ```python
  import rclpy, yaml
  from sensor_msgs.msg import JointState
  port = next(s for s in yaml.safe_load(open("machine.yaml"))["sensors"]
              if s["kind"] == "joint_state")["port"]
  rclpy.init(); node = rclpy.create_node("js")
  got = []
  node.create_subscription(JointState, port, got.append, 1)
  while not got: rclpy.spin_once(node, timeout_sec=0.5)
  print(dict(zip(got[0].name, got[0].position)))
  ```
- **Notes**: message order is not fixed; always match by joint `name`
  (arm joint names: `machine.yaml → actuators[kind=joint_trajectory].joints`).

### A2. Pose (kinematics via TF)

- **What**: where any frame is relative to any other (`frames` in
  `machine.yaml` names the important ones: `world`, `base`, `hand`).
- **CLI**: `ros2 run tf2_ros tf2_echo <world> <hand>` (frame names from
  `machine.yaml → frames`; prints once a second, Ctrl-C after the first).
- **rclpy**: `tf2_ros.Buffer` + `TransformListener`, then
  `buffer.lookup_transform(world, hand, Time())`.
- **Notes**: TF gives you where the ROBOT is; object locations you
  estimate from camera images (B) + geometry.

### A3. Wrist force/torque (kind `wrench`)

- **What**: external wrench at the end effector; contact evidence.
- **Port**: `machine.yaml → sensors[kind=wrench].port`
  (`geometry_msgs/msg/WrenchStamped`).
- **CLI**: `ros2 topic echo <port> --once`
- **Notes**: baseline is not zero (gravity); watch for *changes* on
  contact, not absolute values.

### A4. Odometry (kind `odometry` (only on mobile machines))

- **What**: the mobile base's pose; absent from the manifest = fixed
  base, skip.
- **Port**: `machine.yaml → sensors[kind=odometry].port`
  (`nav_msgs/msg/Odometry`; pose in the `world` frame, `child_frame_id`
  from the manifest entry's `frame`).
- **CLI**: `ros2 topic echo <port> --once`
- **Notes**: read before and after driving (30-action D); displacement
  is your evidence the base actually moved.

## B. Vision (kind `camera_set`)

### B1. Discover the cameras

- The manifest entry gives the topic pattern and the discovery command:
  `ros2 topic list | grep image_raw`. Each camera `<name>` publishes
  color, depth and intrinsics topics per that pattern.

### B2. Grab a color frame

- **Tool**: `python3 tools/perception/cam_snap.py <name>` → `<name>.png`,
  then open the PNG (read the file).
- **rclpy**: subscribe to the topic, convert with
  `cv_bridge.CvBridge().imgmsg_to_cv2(msg, "bgr8")`.

### B3. Depth and intrinsics

- Depth: same pattern on the depth topic (`cam_snap.py` accepts a full
  topic path; depth is 32-bit float meters; convert with
  `imgmsg_to_cv2(msg, "passthrough")`).
- Intrinsics: `ros2 topic echo /<name>/color/camera_info --once`
  (`k` = 3×3 camera matrix); pixels + depth → 3D points in the camera
  optical frame, then TF into `world`.

## C. Graph introspection (beyond the manifest)

- `ros2 topic list` / `ros2 service list` / `ros2 action list` /
  `ros2 node list`
- `ros2 topic info <topic>` and `ros2 interface show <type>` explain any
  port you find.

## Sensor-surface facts

- The TF arm chain may be DISCONNECTED from the world tree (lookups
  like world->hand fail with "unconnected trees"). For the hand pose
  use the FK service; TF stamps are sim-time and a cached buffer can
  serve stale poses after motion.
- `/robot_description` publishes with TRANSIENT_LOCAL durability; a
  plain subscription misses it; request that QoS.
- Camera topics may appear in the graph lazily; the manifest's
  camera_set entry is authoritative, `ros2 node info` shows the full set.
- Oblique (side-view) cameras systematically overestimate the height of
  low, flat objects near their edges; cross-view disagreement on such
  objects usually traces to this.
- The odometry topic publishes only as the sim advances.

**Tool**: `python3 tools/perception/px2world.py <camera> <u> <v>`;
depth + intrinsics + TF for one pixel, printed as world x y z.
\end{lstlisting}

\begin{lstlisting}[style=workspace-document,caption={\texttt{docs/30-action.md}}]
# 30. Action

Ways to move, coarse to fine, keyed by actuator `kind` in `machine.yaml`
(`actuators` list; a dual-arm machine simply has two `joint_trajectory`
entries; check what THIS machine has). Every section: what → port →
CLI → rclpy → tool → notes.

## A. Joint trajectory (kind `joint_trajectory` (the workhorse))

- **What**: track a joint-space trajectory; the primary way to move.
- **Port**: `machine.yaml → actuators[kind=joint_trajectory]` gives the
  action name, the joint names (in order) and their limits.
- **Tool**: `python3 tools/action/fjt_send.py <p1,...,pN> <seconds>`
- **rclpy**: `ActionClient` (`control_msgs/action/FollowJointTrajectory`)
  sending a trajectory whose points have `positions` (one per joint,
  manifest order) and `time_from_start`; several points = one smooth
  pass (see `fjt_send.py` for the skeleton).
- **CLI**: possible via `ros2 action send_goal`, but the YAML is
  unwieldy; prefer the tool/rclpy.
- **Notes**:
  - 2–4 s for large moves; too-short durations overshoot;
  - action result returning ≠ target reached exactly; re-read the
    joint state and compare;
  - nonzero `error_code` = tracking problem (see 40-patterns.md).

## B. Cartesian servo (kind `cartesian_twist`)

- **What**: stream end-effector velocities for small precise motion.
- **Port**: `machine.yaml → actuators[kind=cartesian_twist]`; topic,
  and the `frame` to stamp commands with (linear is m/s).
- **CLI**: `ros2 topic pub --rate 20 <port> geometry_msgs/msg/TwistStamped \
  "{header: {frame_id: <frame>}, twist: {linear: {z: -0.05}}}"`;
  Ctrl-C to stop the stream.
- **rclpy** (the world only advances while commands stream; publish in
  a loop, then re-check sensors):
  ```python
  import rclpy, yaml
  from geometry_msgs.msg import TwistStamped
  tw = next(a for a in yaml.safe_load(open("machine.yaml"))["actuators"]
            if a["kind"] == "cartesian_twist")
  rclpy.init(); node = rclpy.create_node("nudge")
  pub = node.create_publisher(TwistStamped, tw["port"], 10)
  msg = TwistStamped(); msg.header.frame_id = tw["frame"]
  msg.twist.linear.z = -0.05                 # 5 cm/s downward
  for _ in range(20):                        # ~1 s of motion
      pub.publish(msg); rclpy.spin_once(node, timeout_sec=0.05)
  ```
- **Notes**: burst → sense → correct beats one long blind stream.

## C. Gripper (kind `gripper`)

- **What**: open/close the fingers. The manifest entry carries this
  gripper's facts (`open_m`, `closed_m`, `stops_at`, `max_effort`),
  each with its definition written on the line.
- **Port**: `machine.yaml → actuators[kind=gripper]`
  (`control_msgs/action/GripperCommand`).
- **Tool**: `python3 tools/action/gripper_cmd.py <width_m>`
- **CLI** (fill `<port>`, `open_m`/`closed_m`, `max_effort` from the
  manifest entry):
  `ros2 action send_goal <port> control_msgs/action/GripperCommand \
  "{command: {position: <m>, max_effort: <n>}}"`
- **Notes**: after closing on something, finger positions in the joint
  state stop above `closed_m`; that gap is your grasp evidence.

## D. Mobile base (kind `base_twist` (only on mobile machines))

- **What**: drive the base with body-frame velocities. Absent from the
  manifest = this machine's base is fixed; skip this section.
- **Port**: `machine.yaml → actuators[kind=base_twist]`; plain
  `geometry_msgs/msg/Twist` (`/cmd_vel` convention): `linear.x` forward,
  `linear.y` left (holonomic bases only; check `type`), `angular.z`
  counter-clockwise yaw.
- **Tool**: `python3 tools/action/base_goto.py <x> <y> [yaw]`;
  closed-loop on odometry; exits with the last pose as fact on
  stall.
- **CLI**: `ros2 topic pub --rate 10 <port> geometry_msgs/msg/Twist \
  "{linear: {x: 0.3}}"`; Ctrl-C to stop.
- **Semantics**: **one message = one control tick** of motion at that
  velocity. Messages queue up to a small depth (4); anything beyond is
  dropped, so flooding buys nothing. Queued ticks still execute after
  you stop publishing; pace yourself by odometry feedback, not by
  message count or wall time.
- **Notes**:
  - closed-loop is the only reliable pattern: publish a few ticks →
    read odometry (20-perception A4) → correct → repeat;
  - drive and arm motion are separate commands; position the base
    first, then work the arm;
  - operational norm for go-to/docking jobs: park the base directly in
    front of the target, within arm's reach (close enough to work on it
    without re-driving), squarely facing it. "Near" is not parked.

## E. Planning (MoveIt): `machine.yaml → planning`

- **What**: collision-aware planning and IK when raw trajectories are
  not enough (present when `planning.moveit` is true).
- **Ports**: `planning.move_action` (plan+execute,
  `moveit_msgs/action/MoveGroup`, group `planning.group`);
  `planning.ik_service` (`moveit_msgs/srv/GetPositionIK`); Cartesian
  pose → joint positions.
- **Tool**: `python3 tools/action/ik_move.py <x> <y> <z> <qx> <qy> <qz>
  <qw> [seconds] [--at tcp]`; IK then trajectory, loud on failure.
- **Pattern**: IK for the target pose, then A (trajectory) to go there;
  or `move_action` end-to-end when obstacles matter.
- **Notes**: check the returned `error_code` (1 = success); frame
  behavior is NOT what headers suggest; see "Planning facts" below.

## Actuation facts

- Gripper commands are PER-FINGER positions: the opening between the
  fingers is twice the commanded value (nominals in machine.yaml).
  What this machine does with the commanded value is its own fact, not
  a general rule: read `stops_at` on the manifest's gripper entry. Gap
  readings need a few settle ticks; the measured maximum opening can sit
  slightly under nominal.
- A trajectory tolerance-violation result on a long goal is often
  controller lag, not an obstruction: resending the same goal typically
  converges.
- Arm contact and reaction forces can displace the mobile base by
  centimeters; re-read odometry after arm work before trusting stored
  base poses.

## Planning facts (MoveIt)

- The planner's model frame is the ARM BASE even where message headers
  say "world" (machine.yaml planning.planning_frame). Leave `frame_id`
  EMPTY in IK requests; a base-frame label is rejected (-21).
- The planning scene starts EMPTY; no world collision objects exist
  until you publish them; collision meshes carry roughly a centimeter
  of padding, so trust measurements over nominals.
- IK/FK requests HANG if their robot_state carries non-arm joints
  (finger joints included): seed with the arm joints only.
- IK can fail silently (no error, no motion) and non-monotonically
  (nearby poses flip feasibility). Seed requests with the current
  configuration to keep branch continuity, and verify motion actually
  happened from `/joint_states`.
\end{lstlisting}

\begin{lstlisting}[style=workspace-document,caption={\texttt{docs/40-patterns.md}}]
# 40. Working patterns

Cross-cutting habits that save turns and time.

## P1. Act → verify, every time

The clock only moves with activity (10-machine.md §4), so verification
is sensor re-reads, never waits:

1. command (30-action) → 2. re-read the relevant sensor (20-perception)
→ 3. compare with what you intended → 4. correct or continue.

Motion: compare the joint state against your target. Manipulation:
finger gap + a fresh camera frame. World changes: before/after frames
from the same camera.

## P2. Coarse-to-fine motion

Large repositioning with one trajectory move (fast, accurate over big
spans), final approach with short servo bursts (small, sensed,
correctable). Avoid many small trajectory hops and long blind servo
streams alike.

## P3. Approach–grasp–verify (generic pick pattern)

1. Snapshot cameras; locate the target object in the image (pixels +
   depth + intrinsics → world position, 20-perception B3).
2. Move above it (IK → trajectory), open the gripper.
3. Descend with short servo bursts, watching a wrist camera if present.
4. Close the gripper; verify the grasp: finger gap above `closed_m`,
   wrench change on contact.
5. Lift and confirm visually (fresh frame) before moving on.

Placing is the mirror: position above the destination, descend, open,
retreat, verify with a fresh frame.

## P4. Troubleshooting table

| Symptom | Likely cause | Fix |
|---|---|---|
| trajectory result `error_code != 0` | duration too short / target beyond a joint limit | slow down; check limits (machine.yaml) |
| arm lags the servo stream | per-message motion is tiny | keep streaming in a loop; verify between bursts |
| gripper "closed" but nothing held | closed on air | reopen, re-approach lower/centered, use the wrist camera |
| image looks stale | pulled an old frame | snapshot again after the motion completed |
| a port seems missing | wrong name assumed | `ros2 topic list` / `ros2 action list`, then `ros2 topic info` |

## Session-runtime facts

- NEVER end your turn while a motion or background task is pending: an
  ended turn ends the session; nothing resumes it, and the paused
  world means what you were "waiting for" will never arrive. Poll to
  completion first.
- Long commands: run them detached writing to a file with `python3 -u`
  (pipes buffer output invisibly), then poll the file with short
  commands. Build service/action clients once and reuse them; every
  rebuild costs seconds.
\end{lstlisting}

\begin{lstlisting}[style=workspace-document,caption={\texttt{tools/README.md}}]
# tools/: minimal utilities

Six small Python tools for camera capture, geometric measurement, and
motion requests. Plain source files: read them, run them, adapt them,
or write your own programs using the native ROS interfaces (see docs).

| Tool | Usage | Operation |
|---|---|---|
| `perception/cam_snap.py` | `python3 tools/perception/cam_snap.py <camera_or_topic> [out.png]` | save a camera frame (color → PNG; depth → visualization PNG + numerical `.npy` array) |
| `perception/px2world.py` | `python3 tools/perception/px2world.py <camera> <u> <v>` | convert a selected depth pixel to world coordinates using camera intrinsics and TF |
| `action/fjt_send.py` | `python3 tools/action/fjt_send.py <p1,...,pN> <seconds>` | send a joint trajectory and wait for its result; joint names and port come from `machine.yaml` |
| `action/gripper_cmd.py` | `python3 tools/action/gripper_cmd.py <width_m>` | command a gripper position and print the result; `width_m` is the per-finger position |
| `action/ik_move.py` | `python3 tools/action/ik_move.py <x> <y> <z> <qx> <qy> <qz> <qw> [seconds]` | request inverse kinematics, then execute the joint trajectory; see `docs/30-action.md` for planning-frame conventions |
| `action/base_goto.py` | `python3 tools/action/base_goto.py <x> <y> [yaw_rad]` | drive a mobile base toward a world pose using odometry feedback; requires base and odometry ports in `machine.yaml` |
\end{lstlisting}

\subsection{Starter Tool Source Code}
\label[appendix]{app_workspace_tools}
The six tools below are provided as ordinary Python source files.
Their arguments select sensor inputs or motion targets; the agent can inspect or modify the implementations and write additional programs using the same native interfaces.

\begin{lstlisting}[style=workspace-python,caption={\texttt{tools/perception/cam\_snap.py}}]
#!/usr/bin/env python3
"""Save one frame from an image topic to a PNG.

Usage: python3 tools/cam_snap.py <camera_name> [out.png]
Reads /<camera_name>/color/image_raw (pass a full topic to use it as-is).
"""
import sys

import rclpy
from cv_bridge import CvBridge
from sensor_msgs.msg import Image

import cv2


def main() -> None:
    if len(sys.argv) < 2:
        raise SystemExit(__doc__)
    cam = sys.argv[1]
    topic = cam if cam.startswith("/") else f"/{cam}/color/image_raw"
    out = sys.argv[2] if len(sys.argv) > 2 else f"{cam.strip('/').split('/')[0]}.png"
    rclpy.init()
    node = rclpy.create_node("cam_snap")
    got: list = []
    node.create_subscription(Image, topic, lambda m: got.append(m), 1)
    while not got:
        rclpy.spin_once(node, timeout_sec=1.0)
    msg = got[0]
    if "FC" in msg.encoding or "16UC" in msg.encoding:  # depth image
        import numpy as np
        depth = CvBridge().imgmsg_to_cv2(msg, desired_encoding="passthrough")
        np.save(out.rsplit(".", 1)[0] + ".npy", depth)  # raw meters
        finite = depth[np.isfinite(depth)]
        lo, hi = (finite.min(), finite.max()) if finite.size else (0.0, 1.0)
        cv2.imwrite(out, ((depth - lo) / max(hi - lo, 1e-6) * 255).clip(0, 255)
                    .astype("uint8"))
        print(f"{out} (visualization; raw meters in .npy)")
    else:
        cv2.imwrite(out, CvBridge().imgmsg_to_cv2(msg, desired_encoding="bgr8"))
        print(out)
    node.destroy_node()
    rclpy.shutdown()


if __name__ == "__main__":
    main()
\end{lstlisting}

\begin{lstlisting}[style=workspace-python,caption={\texttt{tools/perception/px2world.py}}]
#!/usr/bin/env python3
"""Convert one image pixel to world coordinates (depth + intrinsics + TF).

Usage: python3 tools/perception/px2world.py <camera> <u> <v>
Prints "x y z" in the world frame for pixel column u, row v of the
camera's CURRENT depth frame. Exits nonzero if the pixel has no valid
depth. Pure geometry; which pixel is worth asking about is your call.
"""
import struct
import sys

import numpy as np
import rclpy
from sensor_msgs.msg import CameraInfo, Image
from tf2_ros import Buffer, TransformListener


def _grab(node, topic, msg_type, timeout=15.0):
    got = {}
    sub = node.create_subscription(msg_type, topic,
                                   lambda m: got.setdefault("m", m), 1)
    end = node.get_clock().now().nanoseconds / 1e9 + timeout
    while "m" not in got and node.get_clock().now().nanoseconds / 1e9 < end:
        rclpy.spin_once(node, timeout_sec=0.2)
    node.destroy_subscription(sub)
    if "m" not in got:
        raise SystemExit(f"no message on {topic} within {timeout}s")
    return got["m"]


def main() -> None:
    if len(sys.argv) != 4:
        raise SystemExit(__doc__)
    cam, u, v = sys.argv[1], int(sys.argv[2]), int(sys.argv[3])
    rclpy.init()
    node = rclpy.create_node("px2world")
    tfbuf = Buffer()
    TransformListener(tfbuf, node)

    depth = _grab(node, f"/{cam}/depth/image_raw", Image)
    info = _grab(node, f"/{cam}/color/camera_info", CameraInfo)
    if not (0 <= v < depth.height and 0 <= u < depth.width):
        raise SystemExit(f"pixel ({u},{v}) outside {depth.width}x{depth.height}")
    z = struct.unpack_from("<f", depth.data, (v * depth.width + u) * 4)[0]
    if not np.isfinite(z) or z <= 0:
        raise SystemExit(f"no valid depth at ({u},{v}): {z}")

    fx, fy = info.k[0], info.k[4]
    cx, cy = info.k[2], info.k[5]
    p_cam = np.array([(u - cx) * z / fx, (v - cy) * z / fy, z, 1.0])

    frame = f"{cam}_optical_frame"
    end = node.get_clock().now().nanoseconds / 1e9 + 10.0
    while node.get_clock().now().nanoseconds / 1e9 < end:
        rclpy.spin_once(node, timeout_sec=0.2)
        if tfbuf.can_transform("world", frame, rclpy.time.Time()):
            break
    t = tfbuf.lookup_transform("world", frame, rclpy.time.Time())
    q = t.transform.rotation
    x, y, zz, w = q.x, q.y, q.z, q.w
    R = np.array([
        [1 - 2 * (y * y + zz * zz), 2 * (x * y - zz * w), 2 * (x * zz + y * w)],
        [2 * (x * y + zz * w), 1 - 2 * (x * x + zz * zz), 2 * (y * zz - x * w)],
        [2 * (x * zz - y * w), 2 * (y * zz + x * w), 1 - 2 * (x * x + y * y)],
    ])
    T = np.eye(4)
    T[:3, :3] = R
    T[:3, 3] = [t.transform.translation.x, t.transform.translation.y,
                t.transform.translation.z]
    p = T @ p_cam
    print(f"{p[0]:.4f} {p[1]:.4f} {p[2]:.4f}")
    rclpy.shutdown()


if __name__ == "__main__":
    main()
\end{lstlisting}

\begin{lstlisting}[style=workspace-python,caption={\texttt{tools/action/fjt\_send.py}}]
#!/usr/bin/env python3
"""Send one joint-space trajectory goal and wait for the result.

Usage: python3 tools/action/fjt_send.py <p1,p2,...,pN> <seconds>
Joint names and action port come from machine.yaml (kind
joint_trajectory); N must match that entry's joint count.
"""
import sys
from pathlib import Path

import rclpy
import yaml
from builtin_interfaces.msg import Duration
from control_msgs.action import FollowJointTrajectory
from rclpy.action import ActionClient
from trajectory_msgs.msg import JointTrajectoryPoint


def manifest_entry() -> dict:
    root = Path(__file__).resolve().parents[2]
    m = yaml.safe_load((root / "machine.yaml").read_text())
    return next(a for a in m["actuators"] if a["kind"] == "joint_trajectory")


def main() -> None:
    if len(sys.argv) != 3:
        raise SystemExit(__doc__)
    entry = manifest_entry()
    joints, action = entry["joints"], entry["port"]
    positions = [float(x) for x in sys.argv[1].split(",")]
    seconds = float(sys.argv[2])
    if len(positions) != len(joints):
        raise SystemExit(f"need {len(joints)} positions ({joints}), "
                         f"got {len(positions)}")
    rclpy.init()
    node = rclpy.create_node("fjt_send")
    client = ActionClient(node, FollowJointTrajectory, action)
    if not client.wait_for_server(timeout_sec=10.0):
        raise SystemExit(f"no action server at {action}")
    goal = FollowJointTrajectory.Goal()
    goal.trajectory.joint_names = joints
    pt = JointTrajectoryPoint(positions=positions)
    pt.time_from_start = Duration(sec=int(seconds),
                                  nanosec=int((seconds % 1) * 1e9))
    goal.trajectory.points = [pt]
    send = client.send_goal_async(goal)
    rclpy.spin_until_future_complete(node, send)
    result = send.result().get_result_async()
    rclpy.spin_until_future_complete(node, result)
    code = result.result().result.error_code
    print(f"done error_code={code}" + ("" if code == 0 else " (nonzero = tracking problem)"))
    node.destroy_node()
    rclpy.shutdown()


if __name__ == "__main__":
    main()
\end{lstlisting}

\begin{lstlisting}[style=workspace-python,caption={\texttt{tools/action/gripper\_cmd.py}}]
#!/usr/bin/env python3
"""Open/close the gripper to a target width and wait for the result.

Usage: python3 tools/action/gripper_cmd.py <width_m>
width_m is the desired PER-FINGER position (machine.yaml gripper:
open_m..closed_m). Command a target width, not 0: driving to 0 against
an object keeps squeezing it. Absent from the manifest = this machine
has no gripper.
"""
import sys
from pathlib import Path

import rclpy
import yaml
from control_msgs.action import GripperCommand
from rclpy.action import ActionClient


def main() -> None:
    if len(sys.argv) != 2:
        raise SystemExit(__doc__)
    width = float(sys.argv[1])
    root = Path(__file__).resolve().parents[2]
    m = yaml.safe_load((root / "machine.yaml").read_text())
    entry = next((a for a in m["actuators"] if a["kind"] == "gripper"), None)
    if entry is None:
        raise SystemExit("this machine has no gripper (see machine.yaml)")
    rclpy.init()
    node = rclpy.create_node("gripper_cmd")
    client = ActionClient(node, GripperCommand, entry["port"])
    if not client.wait_for_server(timeout_sec=10.0):
        raise SystemExit(f"no action server at {entry['port']}")
    goal = GripperCommand.Goal()
    goal.command.position = width
    goal.command.max_effort = float(entry.get("max_effort", 30.0))
    fut = client.send_goal_async(goal)
    rclpy.spin_until_future_complete(node, fut, timeout_sec=30)
    handle = fut.result()
    res_fut = handle.get_result_async()
    rclpy.spin_until_future_complete(node, res_fut, timeout_sec=120)
    res = res_fut.result().result
    print(f"reached_goal={res.reached_goal} stalled={res.stalled}")
    rclpy.shutdown()


if __name__ == "__main__":
    main()
\end{lstlisting}

\begin{lstlisting}[style=workspace-python,caption={\texttt{tools/action/ik\_move.py}}]
#!/usr/bin/env python3
"""Move the hand to a world-frame pose: IK -> one trajectory -> done.

Usage: python3 tools/action/ik_move.py <x> <y> <z> <qx> <qy> <qz> <qw> \
           [seconds=4] [--at tcp|hand]
Pose is where the HAND frame goes; --at tcp aims the fingertip point
instead (offset from machine.yaml gripper.tcp_offset_m along hand -Z...
+Z; the offset is applied along the hand's approach axis). IK failures
exit loudly with the MoveIt error code; no solution means NO MOTION
happened. Approach direction, grasp logic, verification stay yours.
"""
import sys
from pathlib import Path

import numpy as np
import rclpy
import yaml
from moveit_msgs.srv import GetPositionIK
from sensor_msgs.msg import JointState


def manifest() -> dict:
    root = Path(__file__).resolve().parents[2]
    return yaml.safe_load((root / "machine.yaml").read_text())


def main() -> None:
    args = [a for a in sys.argv[1:] if not a.startswith("--")]
    if len(args) not in (7, 8):
        raise SystemExit(__doc__)
    x, y, z, qx, qy, qz, qw = map(float, args[:7])
    seconds = float(args[7]) if len(args) == 8 else 4.0
    at_tcp = "--at" in sys.argv and "tcp" in sys.argv

    m = manifest()
    traj_entry = next(a for a in m["actuators"]
                      if a["kind"] == "joint_trajectory")
    planning = m.get("planning", {})
    if at_tcp:
        off = float(m.get("hand", {}).get("tcp_offset_m", 0.0))
        # shift the target back along the hand's approach (+Z of the
        # hand frame, expressed in world via the quaternion)
        R = _quat_to_R(qx, qy, qz, qw)
        x, y, z = np.array([x, y, z]) - off * R[:, 2]

    rclpy.init()
    node = rclpy.create_node("ik_move")
    js = {}
    node.create_subscription(JointState, "/joint_states",
                             lambda msg: js.setdefault("m", msg), 1)
    end = node.get_clock().now().nanoseconds / 1e9 + 10
    while "m" not in js and node.get_clock().now().nanoseconds / 1e9 < end:
        rclpy.spin_once(node, timeout_sec=0.2)
    if "m" not in js:
        raise SystemExit("no /joint_states")

    cli = node.create_client(GetPositionIK, planning.get("ik_service",
                                                         "/compute_ik"))
    if not cli.wait_for_service(timeout_sec=10):
        raise SystemExit("IK service unavailable")
    req = GetPositionIK.Request()
    req.ik_request.group_name = planning.get("group", "panda_arm")
    # machine fact: leave frame_id empty; poses are interpreted in the
    # planner's model frame (the arm base), see machine.yaml planning
    req.ik_request.pose_stamped.header.frame_id = ""
    p = req.ik_request.pose_stamped.pose
    p.position.x, p.position.y, p.position.z = x, y, z
    p.orientation.x, p.orientation.y = qx, qy
    p.orientation.z, p.orientation.w = qz, qw
    # seed with ARM joints only: a full /joint_states (finger joints
    # included) can hang move_group's IK service indefinitely
    seed = JointState()
    arm = set(traj_entry["joints"])
    for n_, p_ in zip(js["m"].name, js["m"].position):
        if n_ in arm:
            seed.name.append(n_)
            seed.position.append(p_)
    req.ik_request.robot_state.joint_state = seed
    fut = cli.call_async(req)
    rclpy.spin_until_future_complete(node, fut, timeout_sec=60)
    res = fut.result()
    if res is None:
        raise SystemExit("IK service did not answer within 60s; no motion "
                         "sent (service load, not necessarily unreachable)")
    if res.error_code.val != 1:
        raise SystemExit(f"IK FAILED (error_code={res.error_code.val}) "
                         "; no motion sent")

    sol = dict(zip(res.solution.joint_state.name,
                   res.solution.joint_state.position))
    joints = traj_entry["joints"]
    positions = ",".join(f"{sol[j]:.6f}" for j in joints)
    rclpy.shutdown()
    # execute through the standard trajectory tool (same act path you
    # would use by hand)
    import subprocess
    r = subprocess.run([sys.executable,
                        str(Path(__file__).parent / "fjt_send.py"),
                        positions, str(seconds)])
    sys.exit(r.returncode)


def _quat_to_R(x, y, z, w):
    return np.array([
        [1 - 2 * (y * y + z * z), 2 * (x * y - z * w), 2 * (x * z + y * w)],
        [2 * (x * y + z * w), 1 - 2 * (x * x + z * z), 2 * (y * z - x * w)],
        [2 * (x * z - y * w), 2 * (y * z + x * w), 1 - 2 * (x * x + y * y)],
    ])


if __name__ == "__main__":
    main()
\end{lstlisting}

\begin{lstlisting}[style=workspace-python,caption={\texttt{tools/action/base\_goto.py}}]
#!/usr/bin/env python3
"""Drive the mobile base to a world pose, closed-loop on odometry.

Usage: python3 tools/action/base_goto.py <x> <y> [yaw_rad] \
           [--tol 0.05] [--yaw-tol 0.1]
Streams velocity bursts, re-reads /odom between bursts, corrects.
Exits 0 at the target; exits 3 on stall, printing the LAST POSE as a
fact ("stalled at x y yaw"); why it stalled is yours to find out.
Absent base_twist in the manifest = this machine's base is fixed.
"""
import math
import sys
from pathlib import Path

import rclpy
import yaml
from geometry_msgs.msg import Twist
from nav_msgs.msg import Odometry


def _pose(msg):
    p, q = msg.pose.pose.position, msg.pose.pose.orientation
    yaw = math.atan2(2 * (q.w * q.z + q.x * q.y),
                     1 - 2 * (q.y * q.y + q.z * q.z))
    return p.x, p.y, yaw


def _read_odom(node, odom, timeout=10.0):
    got = {}
    sub = node.create_subscription(Odometry, odom,
                                   lambda m: got.setdefault("m", m), 1)
    end = node.get_clock().now().nanoseconds / 1e9 + timeout
    while "m" not in got and node.get_clock().now().nanoseconds / 1e9 < end:
        rclpy.spin_once(node, timeout_sec=0.2)
    node.destroy_subscription(sub)
    if "m" not in got:
        raise SystemExit(f"no odometry on {odom}")
    return _pose(got["m"])


def main() -> None:
    args = [a for a in sys.argv[1:] if not a.startswith("--")]
    if len(args) not in (2, 3):
        raise SystemExit(__doc__)
    gx, gy = float(args[0]), float(args[1])
    gyaw = float(args[2]) if len(args) == 3 else None
    tol = float(sys.argv[sys.argv.index("--tol") + 1]) \
        if "--tol" in sys.argv else 0.05
    ytol = float(sys.argv[sys.argv.index("--yaw-tol") + 1]) \
        if "--yaw-tol" in sys.argv else 0.10

    root = Path(__file__).resolve().parents[2]
    m = yaml.safe_load((root / "machine.yaml").read_text())
    base = next((a for a in m["actuators"] if a["kind"] == "base_twist"), None)
    odom_entry = next((s for s in m["sensors"] if s["kind"] == "odometry"), None)
    if base is None or odom_entry is None:
        raise SystemExit("this machine has no mobile base (see machine.yaml)")

    rclpy.init()
    node = rclpy.create_node("base_goto")
    pub = node.create_publisher(Twist, base["port"], 10)
    latest = {}
    node.create_subscription(Odometry, odom_entry["port"],
                             lambda m: latest.update(p=_pose(m)), 10)
    end = node.get_clock().now().nanoseconds / 1e9 + 15
    while "p" not in latest and node.get_clock().now().nanoseconds / 1e9 < end:
        rclpy.spin_once(node, timeout_sec=0.2)
    if "p" not in latest:
        raise SystemExit(f"no odometry on {odom_entry['port']}")

    # Continuous stream, passive odometry: pauses reset the base's
    # velocity ramp, so publish steadily and let the same spin loop
    # deliver odometry updates on the side.
    no_progress = 0
    last_err = None
    for i in range(20000):
        x, y, yaw = latest["p"]
        dx, dy = gx - x, gy - y
        dist = math.hypot(dx, dy)
        dyaw = 0.0
        if gyaw is not None:
            dyaw = math.atan2(math.sin(gyaw - yaw), math.cos(gyaw - yaw))
        if dist < tol and abs(dyaw) < ytol:
            print(f"arrived at {x:.3f} {y:.3f} {yaw:.3f}")
            rclpy.shutdown()
            return
        bvx = math.cos(-yaw) * dx - math.sin(-yaw) * dy
        bvy = math.sin(-yaw) * dx + math.cos(-yaw) * dy
        n = max(dist, 1e-6)
        t = Twist()
        t.linear.x = bvx / n
        t.linear.y = bvy / n
        t.angular.z = max(-1.0, min(1.0, dyaw * 2))
        pub.publish(t)
        rclpy.spin_once(node, timeout_sec=0.03)
        if i % 100 == 99:  # progress check every ~100 ticks
            err = dist + abs(dyaw)
            if last_err is not None and err > last_err - 5e-3:
                no_progress += 1
                if no_progress >= 5:
                    print(f"stalled at {x:.3f} {y:.3f} {yaw:.3f} "
                          f"(remaining {dist:.3f} m, {dyaw:.2f} rad)")
                    rclpy.shutdown()
                    sys.exit(3)
            else:
                no_progress = 0
            last_err = err
    print("cycle budget exhausted before arrival")
    rclpy.shutdown()
    sys.exit(4)


if __name__ == "__main__":
    main()
\end{lstlisting}

\endgroup

\section{Agent-Written Programs}
\label[appendix]{app_programs}

\begingroup
\lstset{aboveskip=2pt,belowskip=3pt}
The programs below are taken verbatim from mainline episodes (Claude Code with Opus 5); only imports and repeated blocks are elided, marked \texttt{...}. They are the kind of code behind the numbers in \cref{tab_perception,tab_manipulation}.

\subsection{CaP-Bench Nut}

\paragraph{Perception: segment the depth array by height and back-project (CaP-Bench Nut).}
The agent saved the overhead depth topic as an array, thresholded height above the table, and turned each connected component into world coordinates with the camera intrinsics it had read from the \texttt{camera\_info} topic.
\begin{lstlisting}
d = np.load("bird_depth.npy")          # depth in meters, from cam_snap.py
f = 579.4112549695428                  # focal length, from /birdview/color/camera_info
zw = 2.0 - d                           # height above the floor (camera at z = 2.0)
mask = ((zw > 0.825) & (zw < 1.3)).astype(np.uint8)
n, lab, stats, cent = cv2.connectedComponentsWithStats(mask, 8)
for i in range(1, n):
    if stats[i, 4] < 50: continue
    ys, xs = np.where(lab == i)
    zz = zw[ys, xs]
    wx = (ys - 240) * d[ys, xs] / f; wy = (xs - 320) * d[ys, xs] / f
    print(f"comp{i} area={stats[i,4]} zmax={zz.max():.4f} zmin={zz.min():.4f}")
    print(f"   world x[{wx.min():.4f},{wx.max():.4f}] y[{wy.min():.4f},{wy.max():.4f}]"
          f" centroid=({wx.mean():.4f},{wy.mean():.4f})")
\end{lstlisting}

\paragraph{Perception: cross-check from a second camera (same episode).}
Before moving, the agent repeated the measurement from an oblique camera, transforming its depth image into the world frame with the camera pose it had read from the transform tree.
\begin{lstlisting}
d = np.load("ag_depth.npy"); f = 579.4112549695428
# agentview optical frame: world pos (0.5,0,1.35) quat (0.6531,0.6531,-0.2710,-0.2710)
q = np.array([0.6531, 0.6531, -0.2710, -0.2710]); q = q / np.linalg.norm(q)
x, y, z, w = q
R = np.array([[1-2*(y*y+z*z), 2*(x*y-z*w),   2*(x*z+y*w)],
              [2*(x*y+z*w),   1-2*(x*x+z*z), 2*(y*z-x*w)],
              [2*(x*z-y*w),   2*(y*z+x*w),   1-2*(x*x+y*y)]])
t = np.array([0.5, 0.0, 1.35])
H, W = d.shape
u, v = np.meshgrid(np.arange(W), np.arange(H))
Z = d.astype(np.float64)
P = np.stack([(u-320)*Z/f, (v-240)*Z/f, Z], -1)   # camera frame
Wd = P @ R.T + t                                  # world frame
...
blob((zz > 0.94) & (zz < 0.96), "PEGTOP")
blob((zz > 0.834) & (zz < 0.846), "NUTTOP")
\end{lstlisting}

\paragraph{Perception: cropping and annotating a camera image for inspection (same episode).}
To read the nut's orientation and handle width, the agent cropped the overhead image, enlarged it, and drew a pixel grid with coordinate labels, then read the result with its own vision.
\begin{lstlisting}
img = cv2.imread("snap_birdview.png")
c = img[130:235, 335:445]
c = cv2.resize(c, None, fx=6, fy=6, interpolation=cv2.INTER_NEAREST)
# draw grid every 10 original px
for i in range(0, 111, 10):
    cv2.line(c, (i*6, 0), (i*6, c.shape[0]), (0, 0, 255), 1)
    cv2.putText(c, str(335 + i), (i*6 + 2, 12), 0, 0.35, (0, 0, 255), 1)
for j in range(0, 106, 10):
    cv2.line(c, (0, j*6), (c.shape[1], j*6), (0, 255, 0), 1)
    cv2.putText(c, str(130 + j), (2, j*6 - 2), 0, 0.35, (0, 255, 0), 1)
cv2.imwrite("nut_zoom.png", c)
\end{lstlisting}

\paragraph{Manipulation: insert in monitored segments (CaP-Bench Nut).}
The insertion is a sequence of short joint-trajectory segments; after each, the program reads the wrench and the finger gap, and it aborts if the planned path dips below a floor it computed from the arm's link heights.
\begin{lstlisting}
def go(name, zt, secs, prev, phi=PHI):
    tw = (XY[0], XY[1], zt)
    qs = plan.solve(tw, phi, seeds(prev, tw))       # own IK search
    if qs is None:
        raise SystemExit("IK FAIL " + name)
    mz, t = plan.path_min_z(prev, qs)
    floor = min(plan.link_zs(prev) + plan.link_zs(qs)) - 0.004
    if mz < floor:
        raise SystemExit("PATH minz %.3f floor %.3f %s" % (mz, floor, name))
    code = r.goto_q(qs, secs)                        # FollowJointTrajectory
    act = np.array(r.arm_q())
    pa, _ = kin.fk_tcp(act)
    w = r.wrench()                                   # external wrench topic
    print(f"{name} z={zt:.3f} code={code} tcp={np.round(pa+rc.BASE_IN_WORLD,4)} "
          f"gap={r.finger_gap():.5f} F={np.round(w[:3],2)} T={np.round(w[3:],3)}")
    return act

for z, s in [(0.99, 1.5), (0.965, 1.0), (0.945, 1.0), (0.92, 1.0),
             (0.89, 1.0), (0.86, 1.0), (0.833, 1.0)]:
    q = go("descend", z, s, q)
\end{lstlisting}

\subsection{LIBERO-PRO Drawer}

\paragraph{Manipulation: a client library (LIBERO-PRO, open the drawer and put the cream cheese inside).}
The agent wrapped the robot's native interfaces once, in a module it then imported from every step program: the inverse-kinematics service, the trajectory and gripper actions, the joint-state subscription, and the velocity stream.
\begin{lstlisting}
class Rig:
    def __init__(self):
        rclpy.init()
        self.node = rclpy.create_node("rig")
        self._js = {}
        self.node.create_subscription(JointState, "/joint_states", self._on_js, 10)
        self.ik = self.node.create_client(GetPositionIK, "/compute_ik")
        self.fk = self.node.create_client(GetPositionFK, "/compute_fk")
        self.traj = ActionClient(self.node, FollowJointTrajectory, TRAJ["port"])
        self.grip = ActionClient(self.node, GripperCommand, GRIP["port"])
        self.twist_pub = self.node.create_publisher(TwistStamped, TWIST["port"], 10)
        ...
    def servo(self, lin, ticks=20, rate_hz=20.0):
        """Stream a world/base-frame twist (m/s) for `ticks` messages."""
        msg = TwistStamped()
        msg.header.frame_id = TWIST["frame"]
        msg.twist.linear.x, msg.twist.linear.y, msg.twist.linear.z = (
            float(lin[0]), float(lin[1]), float(lin[2]))
        for _ in range(ticks):
            msg.header.stamp = self.node.get_clock().now().to_msg()
            self.twist_pub.publish(msg)
            rclpy.spin_once(self.node, timeout_sec=1.0 / rate_hz)
\end{lstlisting}

\paragraph{Manipulation: approach the handle under feedback (same episode).}
After a first grasp had closed on air, the agent rewrote the approach as a proportional servo on the measured offset and used the finger gap as evidence of a grasp.
\begin{lstlisting}
"""Reopen, drive the fingers the rest of the way onto the bar, close.

First attempt stopped with the fingertips at y=-0.1354, still 4 mm in
front of the bar's front surface (-0.1392), so the fingers closed on air
(gap 0.0019). Push to y=-0.1545 this time and keep servoing until the
error is actually gone.
"""
GRASP = np.array([0.0365, -0.1545, 1.0880])
r = Rig()
print("open ->", round(r.gripper(0.04), 4))
for i in range(25):
    p, _ = r.tcp_pose()
    dy = GRASP[1] - p[1]
    dz = GRASP[2] - p[2]
    print(f"  {i} tcp {np.round(p, 4)} dy {dy:+.4f}")
    if abs(dy) < 0.0025:
        break
    r.servo([0.0, np.clip(dy * 5, -0.05, 0.05), np.clip(dz * 5, -0.02, 0.02)], ticks=20)
gap = r.gripper(0.0)
print("closed -> gap", round(gap, 5))
print("grasp evidence: gap > closed_m(0.0) means something is held")
\end{lstlisting}

\paragraph{Manipulation: pull the drawer with guards (same episode).}
\leavevmode\par
\begin{lstlisting}
"""Pull the grasped top drawer open along +y, watching the grip."""
PULL_TO = -0.020  # world y for the fingertips; ~0.14 m of travel
r = Rig()
p0, _ = r.tcp_pose()
for i in range(20):
    p, _ = r.tcp_pose()
    gap = r.finger_gap()
    if gap < 0.008:
        print("  GRIP LOST (gap collapsed)"); break
    if p[1] >= PULL_TO:
        print("  reached pull target"); break
    r.servo([0.0, 0.05, 0.0], ticks=20)
    q, _ = r.tcp_pose()
    print(f"  {i} tcp {np.round(q, 4)} gap {r.finger_gap():.5f} moved {q[1]-p[1]:+.4f}")
    if abs(q[1] - p[1]) < 0.001:
        print("  stalled -- drawer at its travel limit or arm blocked"); break
p, _ = r.tcp_pose()
print("end", np.round(p, 4), "total pull", round(p[1] - p0[1], 4))
\end{lstlisting}

\subsection{CaP-Bench Wipe}

\paragraph{Perception: ray-cast a colour mask onto the table plane (CaP-Bench Wipe).}
Each connected blob of the brown mask is cast from the camera centre onto the table plane; a blob counts as spill only if nine tenths of it lands on the table, and the hits are rasterized into a 5\,mm grid that is intersected across three cameras.
\begin{lstlisting}
grid = np.zeros((len(GX), len(GY)), dtype=bool)
n_lab, lab = cv2.connectedComponents(mask)
for i in range(1, n_lab):
    vs, us = np.nonzero(lab == i)
    if len(us) < 30: continue
    dirs = (R @ np.vstack([(us - cx) / fx, (vs - cy) / fy, np.ones(len(us))])).T
    s = (Z0 - C[2]) / dirs[:, 2]          # ray length to the table plane
    good = s > 0
    P = C + dirs[good] * s[good, None]
    on = ((P[:, 0] > TAB[0]) & (P[:, 0] < TAB[1]) & (P[:, 1] > TAB[2]) & (P[:, 1] < TAB[3]))
    # a blob is spill only if essentially ALL of it lands on the table
    if on.mean() < 0.9: continue
    P = P[on]
    ix = np.clip(((P[:, 0] - TAB[0]) / CELL).astype(int), 0, len(GX) - 1)
    iy = np.clip(((P[:, 1] - TAB[2]) / CELL).astype(int), 0, len(GY) - 1)
    grid[ix, iy] = True
\end{lstlisting}

\subsection{CaP-Bench Restack}

\paragraph{Manipulation: a force-guarded descent (CaP-Bench Restack).}
\leavevmode\par
\begin{lstlisting}
def guarded_down(self, xy, z_start, z_floor, step=0.002, fthresh=1.5, yaw=YAW, label="guarded"):
    """Creep down in `step` increments, stopping at the first real change
    in the wrist wrench.  Returns (contact_z, contacted)."""
    f0 = self.force()
    seed = self.a.arm_joints()
    z = z_start
    while z > z_floor + 1e-9:
        z = max(z - step, z_floor)
        seed = self.ik_tcp(xy, z, yaw, seed)
        self.a.send_traj([(seed, 0.35)], "")
        t = self.tcp()
        df = self.force() - f0
        n = np.linalg.norm(df)
        print(f"    z={t[2]:.4f} dF={np.round(df, 2)} |dF|={n:.2f}", flush=True)
        if n > fthresh:
            print(f"  {label}: CONTACT at z={t[2]:.4f} (|dF|={n:.2f})", flush=True)
            return t[2], True
\end{lstlisting}

\subsection{CaP-Bench Two-Arm Lift}

\paragraph{Manipulation: move both arms at once (CaP-Bench two-arm lift).}
Both trajectory goals are sent before either result is awaited.
\begin{lstlisting}
def move_both(self, traj, label=""):
    """traj: {arm: (list_of_q, list_of_times)}; both arms execute concurrently."""
    handles = {}
    for arm, (qs, ts) in traj.items():
        f = self.fjt[arm].send_goal_async(self._goal(arm, qs, ts))
        rclpy.spin_until_future_complete(self.node, f)
        gh = f.result()
        if not gh.accepted:
            raise SystemExit(f"{arm}: goal rejected")
        handles[arm] = gh.get_result_async()
    codes = {}
    for arm, fut in handles.items():
        rclpy.spin_until_future_complete(self.node, fut)
        codes[arm] = fut.result().result.error_code
    print(f"[move {label}] error_codes={codes}", flush=True)
    self.refresh()
    return codes
\end{lstlisting}

\subsection{CaP-Bench Handover}

\paragraph{Manipulation: level a drooping handle (CaP-Bench handover).}
After the handover the hammer drooped; the agent measures the handle's pitch from its top face, counter-rotates the wrist, and re-solves so the handle centre lands at the target height.
\begin{lstlisting}
pitch, mid, R = measure(r)
T = r.hand_world("right")
p_h = T[:3, :3].T @ (mid - T[:3, 3])          # handle midpoint in the hand frame
dR = Rx(-pitch)
R_new = dR @ T[:3, :3]
off = R_new @ p_h
origin = np.array([mid[0], mid[1], Z_TARGET]) - off
q = r.solve("right", origin, R_new, seed=r.q("right"), other_q=r.q("left"),
            table_z=0.845, grasp_contact=True)
dq = np.abs(q - r.q("right")).max()
print(f"== levelling, dq_max={dq:.3f}")
r.send_q("right", q, float(np.clip(2.5 + dq * 2.5, 2.5, 7.0)))
\end{lstlisting}
\endgroup

\end{document}